\documentclass{article}
\usepackage{microtype}
\usepackage{graphicx}
\usepackage{booktabs}
\usepackage{multirow}
\usepackage{amsmath,amssymb,amsthm,mathtools}
\usepackage{enumitem}
\usepackage{longtable}
\usepackage{tikz}
\usetikzlibrary{calc}
\usepackage{hyperref}

\newif\ifanon
\anonfalse   
\ifanon\usepackage{icml2027}\else\usepackage[preprint]{icml2027}\fi
\usepackage[capitalize,noabbrev]{cleveref}

\newcommand{\KL}{\mathrm{KL}}
\newcommand{\fl}{\mathrm{fl}}
\newcommand{\lgp}{\operatorname{log1p}}
\newcommand{\emo}{\operatorname{expm1}}
\newcommand{\kld}{\texttt{kl\_divergence}}
\newcommand{\fkl}{\texttt{F.kl\_div}}

\newcommand{\A}{\textsc{Official}}
\setlist{nosep,leftmargin=*}
\icmltitlerunning{The KL Divergence Is Never Negative, Except in PyTorch}

\begin{document}
\twocolumn[
\icmltitle{The KL Divergence Is Never Negative, Except in PyTorch}
\begin{icmlauthorlist}
\icmlauthor{Zhen Yang}{yale}
\icmlauthor{Wei Tang}{oxford}
\icmlauthor{Chu Chen}{cambridge}
\icmlauthor{Rui Qian}{fudan}
\icmlauthor{Yixuan Chen}{oxford}
\icmlauthor{Fan Yang}{wfu}
\icmlauthor{Kangning Cui}{cityudg}
\end{icmlauthorlist}

\icmlaffiliation{yale}{Yale University, New Haven, Connecticut, USA}
\icmlaffiliation{oxford}{University of Oxford, Oxford, United Kingdom}
\icmlaffiliation{cambridge}{University of Cambridge, Cambridge, United Kingdom}
\icmlaffiliation{fudan}{Fudan University, Shanghai, China}
\icmlaffiliation{wfu}{Wake Forest University, Winston-Salem, North Carolina, USA}
\icmlaffiliation{cityudg}{City University of Hong Kong (Dongguan), Dongguan, China}
\icmlcorrespondingauthor{Fan Yang}{yangfan@wfu.edu}
\icmlcorrespondingauthor{Kangning Cui}{kangning.cui@cityu-dg.edu.cn}
\icmlkeywords{KL divergence, floating point, numerical stability, PyTorch, distillation, quantization}
\vskip 0.2in
]
\printAffiliationsAndNotice{}

\begin{abstract}
A Kullback--Leibler (KL) divergence is never negative, yet PyTorch users have reported negative values for years, and the standard advice is to clamp them at zero.
The sign is only the visible part of a silent failure. Many machine-learning applications compare nearby distributions, where conventional KL formulas lose accuracy.
On 8.5 million inputs with certified references across 18 closed-form KL families and \texttt{F.kl\_div}, we find widespread errors in values and gradients on CPU and GPU. Among scalar families in ordinary parameter ranges, over 85\% of FP32 values have at least 1\% relative error near KL $10^{-7}$ (FP16: $10^{-3}$), and about 30\% become negative at smaller divergences.
More precision is not the cure: evaluated one precision higher, the same formulas still leave 7--51\% of FP32 values significantly wrong for 22 of 23 configurations.
Stable closed-form decompositions, compensated arithmetic, and analytic gradients remove over 97\% of significant errors in the working format. Strict repairs eliminate significant errors on 7.2 million scored closed-form samples and correctly round all 567 million scored \texttt{F.kl\_div} components.
The errors affect parameter fitting, Bayesian decisions, identical-start distillation (up to 97\% top-1 disagreement), KL-based layer rankings in a quantized LLM, and 19--59\% of GPT-2 circuit edge scores. With the repaired code they all but vanish.
\ifanon
Code and data will be released upon publication.
\else
Code and data: \url{https://github.com/xi-xiaoran/stablekl}.
\fi
\end{abstract}
\section{Introduction}
\label{sec:intro}

The Kullback--Leibler (KL) divergence \citep{kullback1951} between two distributions is non-negative, and zero only if they coincide, but code does not always agree.
Since 2019, users of PyTorch \citep{paszke2019} have repeatedly reported negative values from both of its KL functions, and the answer has been the same each time: this is floating-point error, clamp the result at zero (\cref{app:reports}).
Clamping hides the symptom and leaves the error. Clamping negative KL estimates sets their values and backpropagated gradients to zero. Many inaccurate KL estimates are positive and finite.

The KL divergence is the regulariser of variational autoencoders \citep{kingma2014}, the trust-region constraint of policy optimisation \citep{schulman2015,schulman2017,shao2024}, the loss of knowledge distillation \citep{hinton2015,agarwal2024}, and a standard metric for how far a compressed model has moved \citep{nagel2021,xin2026}.
These applications often compare closely related distributions: consecutive policies, a student and its teacher, or a model and its quantized copy. Among the 116 PyTorch-based packages we inspected, 39 call at least one of the two KL interfaces (\cref{app:usage}).

\begin{figure*}[t]
\centering
\includegraphics[width=0.86\textwidth]{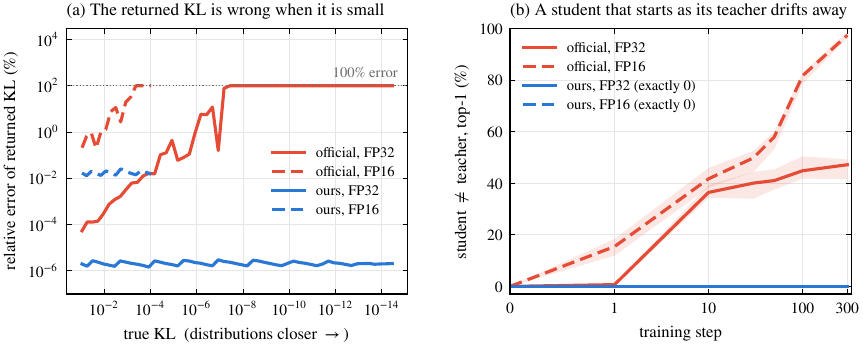}
\caption{The problem and one consequence. The official PyTorch code is shown in red and ours in blue, with solid lines for FP32 and dashed lines for FP16.
\textbf{(a)} Relative error of the returned KL between two Exponential distributions as they approach each other (median over 3000 parameter pairs per point, with FP16 shown while the true value is a normal number). The official closed form is implemented correctly, yet its error reaches 100\% once the true KL falls below about $4\times10^{-8}$ in FP32 ($4\times10^{-4}$ in FP16), where it returns zero or a negative number. The repaired divergence stays at the accuracy of the format.
\textbf{(b)} A student initialised as an exact copy of its teacher and trained with the KL loss alone (vocabulary 151{,}936, Adam, learning rate $10^{-2}$, median and range over 10 seeds, other settings in \cref{app:downstream}). With the official loss the student's top-1 predictions leave the teacher's, whereas with our logits path they do not move.}
\label{fig:overview}
\end{figure*}

Closeness is exactly where the textbook formulas break down.
If $P$ and $Q$ differ by a relative amount $\delta$, then typically $\KL(P\,\|\,Q)=O(\delta^2)$, while almost every closed form computes it as a sum of $O(1)$ or $O(\delta)$ terms of mixed sign.
The rounding error of those terms is of the order of the unit roundoff $u$ (about $6\times10^{-8}$ in FP32, $5\times10^{-4}$ in FP16), so the computed divergence is noise once the true KL is of the order of $u$ or smaller (\cref{fig:overview}). It may be negative, zero for distinct distributions, or non-zero for identical ones, and its gradient need not point downhill.
Low-precision formats \citep{micikevicius2018,kalamkar2019} move this threshold into the range of practical values.

Numerical failures of \texttt{softmax} and cross-entropy explain stalled grokking \citep{prieto2025} and distillation drift \citep{barron2026}, but the KL divergence itself, across distribution families and interfaces, has not been measured.
We measure it in both KL interfaces of PyTorch, \nolinkurl{torch.distributions.kl_divergence} (interface~A, closed forms between two distribution objects) and \nolinkurl{torch.nn.functional.kl_div} (interface~B, the loss behind \texttt{nn.KLDivLoss}), and show how to remove the error:

\begin{itemize}
\item \textbf{The failure is systematic, and silent} (\cref{sec:measure}). On 7.5 million parameter pairs for the 18 same-family closed forms of interface~A and 0.94 million batches for interface~B, stratified by the true KL from $10^{-15}$ to $1$ and certified by interval arithmetic, the official results degrade steadily as the distributions approach each other, for every family, on CPU and GPU.
\item \textbf{More precision is not the fix, but a different formula is} (\cref{sec:method}). Across 18 closed forms, we combine non-negative decompositions, compensated arithmetic, special-function evaluation, and analytic gradients. For \fkl{}, we repair signed elements, reductions, and gradients.
The rewrites remove over 97\% of significant errors in the working precision, and strict repairs yield no significant errors on 7.2 million scored closed-form samples and correctly round all 567 million interface-B components.
\item \textbf{Two error sources must be told apart} (\cref{sec:bg}): the \emph{operator error} of \fkl{} on the inputs it receives, and the \emph{input-representation error} already present in rounded (log-)probabilities, which no implementation of \fkl{} can repair and which requires a divergence computed from logits.
\item \textbf{The errors change outcomes} (\cref{sec:exp}), from fitting a single distribution and a sequential Bayesian experiment to distillation, KL-based decisions for a quantized LLM and circuit analysis of GPT-2 models.
\end{itemize}

\section{Background and Error Model}
\label{sec:bg}

\paragraph{Floating point.} We use the standard model $\fl(a\circ b)=(a\circ b)(1+\epsilon)$, $|\epsilon|\le u$, for $\circ\in\{+,-,\times,/\}$ \citep{higham2002}, for binary16 (FP16), binary32 (FP32), binary64 (FP64) \citep{ieee754} and bfloat16 (BF16).
A result is \emph{correctly rounded} if it equals the exact value rounded to nearest in the output format, and errors in \emph{ULP} are in units of the spacing of the format at the exact value.
Subtracting two nearby floating-point numbers is exact \citep{sterbenz1974}, but exposes the rounding errors already present in them (\emph{cancellation}) \citep{goldberg1991}.

\paragraph{Cancellation near equality.} Take two Exponential distributions with rates $\lambda_p,\lambda_q$, for which $\KL=r-1-\log r$ with $r=\lambda_q/\lambda_p=1+x$.
The exact value is $x^2/2+O(x^3)$, but $r-1-\log r$ is evaluated near $1$, where the absolute rounding error is about $u$. The relative error is therefore about $2u/x^2$, which exceeds 100\% once $|x|<\sqrt{2u}\approx3.5\times10^{-4}$ in FP32.
For example, for $\lambda_p=1$ and $\lambda_q=\fl(1.0001)$ in FP32 the true value is $5.0013\times10^{-9}$, but $r-\log r=1.000000005\ldots$ rounds to $1$ and the official formula returns exactly $0$, so two different distributions are reported as identical (\cref{fig:overview}a).
The same structure, a second-order quantity obtained from first-order or zeroth-order terms, appears in the closed forms we examined (\cref{app:ifaceA}). When only the support changes (Uniform, Pareto), the KL is first order.
It is not removed by more precision alone, because FP32 parameters can differ by $x=2^{-24}$, for which the same formula in FP64, whose unit roundoff is $u_{64}=2^{-53}$, can have a relative error of $2u_{64}/x^2\approx0.06$.

{\emergencystretch=2em\paragraph{Interfaces.} Interface~A, \kld$(P,Q)$, dispatches on the types of the two distribution objects to a closed form in their parameters. Gradients come from automatic differentiation, and for families with two parameterisations (\texttt{probs} or \texttt{logits}) the constructors convert and clamp the parameters first.
Interface~B, \fkl$(x,t)$, takes a tensor $x$ of log-probabilities and a target $t$ of probabilities (or of log-probabilities when \texttt{log\_target=True}), evaluates $t(\log t-x)$ or $e^{t}(t-x)$ element-wise, and reduces.\par}

\paragraph{Operator error versus input-representation error.} For interface~B the inputs are themselves outputs of a rounded computation, typically \texttt{log\_softmax} of model logits.
Let $T_{\mathrm{op}}$ be the exact operator result on its stored inputs and $T_{\mathrm{e2e}}$ the corresponding result from the original logits, with the same reduction.
We call the difference between the computed output and $T_{\mathrm{op}}$ the \emph{operator error}, and $T_{\mathrm{op}}-T_{\mathrm{e2e}}$ the \emph{input-representation error}.
Only the first belongs to \fkl.
A related backward error is Softmax Cancellation \citep{barron2026}. The gradient of the forward KL with respect to $x$ involves no cancellation. The cancellation happens afterwards, in the backward pass of \texttt{log\_softmax}, where $q-p$, the difference between the input distribution $q=e^{x}$ and the target $p=t$, is formed from two rounded probability vectors, and the result is non-zero even when the two distributions are identical.
\section{Numerical Errors in PyTorch}
\label{sec:measure}

We first measure the failure, with a reference that is exact for the bit patterns the library receives, so that every reported error belongs to the implementation and not to the test.

\paragraph{Data and ground truth.} For interface~A we test the 18 distribution-specific same-family closed forms in 23 parameter and dimension configurations (five generic or delegated registrations are out of scope, and the two multivariate families are scored in FP32 only).
For each we draw $P$ from an ordinary parameter region R0 and from stress regions (very large or small parameters, probabilities near 0 or 1), and construct $Q$ so that the \emph{true} $\KL(P\|Q)$ falls in a prescribed decade: one control bin $[10^{-1},1)$ and 14 bins down to $[10^{-15},10^{-14})$.
Two further groups contain pairs whose parameters differ by 1--8 ULP, and identical pairs.
This gives 7,542,469 samples, of which 7,234,454 are scored. Cells that the floating-point grid cannot populate remain empty.
For interface~B we generate 942,200 batches over \texttt{log\_target}, FP16/FP32, $K\in\{2,4,16,256,1024\}$ classes, two batch sizes and all four reductions, from logits stratified in the same way.
Parameters are stored in the target format first. The exact KL and every gradient component are then enclosed with Arb ball arithmetic \citep{johansson2017} tightly enough to determine a unique correctly rounded value in that format (\cref{app:protocol,app:details}).

\paragraph{Methods and metrics.} \A{} is PyTorch 2.15.0 on CPU, and we repeated the evaluation on an NVIDIA A100 GPU (\cref{app:gpu}).
\textsc{Official+}, the natural ``just use more precision'' baseline, evaluates the unchanged official code after promoting the parameters exactly to the next format (FP32 for FP16 inputs, FP64 for FP32 inputs) and rounds the result once.
An output is a \emph{significant error} if it is off by at least 8 ULP or is qualitatively wrong (negative, NaN or infinite KL where the truth is finite and positive; spurious zero or non-zero; sign-flipped gradient component).

\begin{figure*}[t]
\centering
\includegraphics[width=0.86\textwidth]{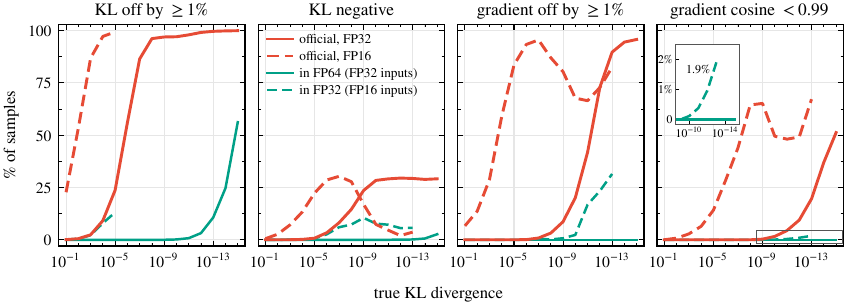}
\caption{Errors of the official interface-A implementation (red) and of the same formula evaluated in the next higher precision (green) versus the true KL, for the 16 scalar-family configurations in the ordinary parameter region R0. Solid lines are FP32 inputs and dashed lines FP16 inputs. The inset enlarges the lower right corner of the last panel. FP16 relative error is shown only while the true KL is a normal number (above about $6\times10^{-5}$).}
\label{fig:errkl}
\end{figure*}

\paragraph{Errors near equality.} The error grows as the true KL approaches the unit roundoff $u$ (\cref{fig:errkl}, which pools the 16 scalar-family configurations in R0, with details in \cref{app:bins}).
In the control bin the official values are accurate (0.1\% of the FP32 values are off by 1\% or more). As the KL approaches $u\approx6\times10^{-8}$ most are wrong (86\% at $10^{-7}$), and below that nearly all are, with about three in ten negative from $10^{-10}$ down.
FP16 follows the same curve four decades earlier, as its roundoff predicts.
Gradients, being first-order quantities, survive longer but fail in a worse way. At $10^{-13}$ a fifth of the gradient vectors have a cosine below 0.99 with the true one, and at $10^{-15}$, 6.8\% point into the opposite half-space.
No family is exempt (\cref{tab:main}, columns \A).

\paragraph{Precision promotion.} Evaluating the same formula one format higher moves the failure; it does not remove it, because two inputs can still differ by one unit in the last place of the \emph{input} format (\cref{sec:bg}).
For FP32 inputs \textsc{Official+} shifts the curve of \cref{fig:errkl} about nine decades down (56\% off by $\ge1\%$ at $10^{-15}$), and for FP16 inputs it leaves a residue (13\% at $10^{-5}$).
Over all bins and regions it leaves 7--51\% of the FP32 KL values significantly wrong in 22 of 23 configurations (\cref{tab:main}), Uniform being the only exception.

\paragraph{Loss-operator errors.} \fkl{} fails in a different way.
Its element $t(\log t-x)$ cancels whenever $t\approx e^{x}$. For probability targets 35--70\% of the FP32 \texttt{sum} outputs are off by at least 1\%, and scaled reductions can produce large target-gradient errors (FP16, \texttt{mean}, $K=256$: 95.4\% of gradient rows contain a component of the wrong sign).
Unlike the closed forms, these \emph{operator} errors essentially disappear in FP64, but the error inherited from the inputs does not, since $T_{\mathrm{op}}$ differs from $T_{\mathrm{e2e}}$ by $\ge1\%$ on 40--99\% of the batches, in any precision of the operator (\cref{app:ifaceB}).

\paragraph{CPU and GPU.} The failure persists on the GPU. On an A100 the aggregate significant-error rates for interface~A differ from the CPU rates by at most 1.5 percentage points. For \fkl{}, 50.0\% of forward and 71.1\% of non-boundary backward components are correctly rounded (CPU: 52.5\% and 76.4\%).
The two devices nevertheless do not return the same bits, and 27\% of the FP32 KL values of \A{} differ.
Our strict repair \textsc{Ours} (\cref{sec:method}) has no significant error on either device (\cref{app:gpu}).
\section{Stable KL Computation}
\label{sec:method}

The repairs of the 18 closed forms differ in detail (\cref{app:ifaceA}) but share four ingredients, which we illustrate with the smallest example that needs each. \fkl{} additionally requires accurate signed elements and reductions (\cref{app:ifaceB}).

\paragraph{Stable decompositions.}
The function
\begin{equation}
\varphi(x)=x-\lgp(x)\;\ge 0,\qquad x>-1,
\label{eq:phi}
\end{equation}
is the Exponential KL with $x=(\lambda_q-\lambda_p)/\lambda_p$, and it is the building block of almost every other family.
For example, for two normal distributions $\mathcal N_p=\mathcal N(\mu_p,\sigma_p^2)$ and $\mathcal N_q=\mathcal N(\mu_q,\sigma_q^2)$, with $e=(\sigma_p-\sigma_q)/\sigma_q$ and $w=(\mu_p-\mu_q)/\sigma_q$,
\begin{equation}
\KL\big(\mathcal N_p\,\|\,\mathcal N_q\big)=\varphi(e)+\tfrac12e^2+\tfrac12w^2 ,
\label{eq:normal}
\end{equation}
whereas the official code evaluates $\tfrac12(r^2+w^2-1-\log r^2)$ with $r=\sigma_p/\sigma_q$, in which $w^2$ is absorbed by $r^2\approx1$.
Analogous identities (\cref{app:ifaceA}, which defines each argument) give $\KL=\lgp(r)$ with $r\ge0$ for Cauchy and Uniform, $p\,\varphi(x_1)+(1-p)\,\varphi(x_2)$ for Bernoulli, $\sum_kp_k\varphi(\rho_k)$ for Categorical, $\sum_i\varphi(N_{ii})+\frac12\|N\|_F^2+\frac12\|v\|^2$ for the multivariate normal, and sums of $\varphi$ terms and Bregman divergences of convex functions for Gamma, Beta and Dirichlet.
Every term is non-negative, so the sum is as accurate as its terms, cannot be negative, and is exactly zero when all terms are.
$\varphi$ itself still cancels as $x\to0$, so on $-\tfrac34\le x\le3$ we use the exact identities $\lgp(x)=2\operatorname{atanh}(s)$, $s=x/(2+x)$, to write
\begin{equation}
\varphi(x)=\frac{x^2}{2+x}-2s^3\sum_{k\ge1}\frac{s^{2(k-1)}}{2k+1},
\label{eq:series}
\end{equation}
a convergent series whose truncation after 9/17/36 terms (FP16/FP32/FP64) is bounded rigorously by $u/16$ relative to $\varphi$.

\paragraph{Accurate differences.}
A non-negative decomposition is only useful if its arguments ($x$, $e$, $\rho_k$, $N$) are accurate relative to their own size.
We therefore compute them from differences of the inputs, which are exact for nearby floating-point numbers \citep{sterbenz1974}, and, where a difference of products is needed (e.g.\ $N_k=y_kS_x-x_kS_y$ for Categorical parameters $x$, $y$ with sums $S_x$, $S_y$, \cref{app:cat}), from error-free transformations \citep{knuth1997,dekker1971,shewchuk1997,ogita2005}.
As a consequence, structurally exact zeros are returned as $+0$, including pairs whose parameters differ but whose distributions coincide (proportional \texttt{probs}, shifted \texttt{logits}, different low-rank factors of one covariance).

\paragraph{Analytic gradients.}
Automatic differentiation of a cancelling formula inherits the cancellation and adds failure modes of its own.
We implement the derivative of the rewritten form directly, e.g.\ $\partial\KL/\partial\sigma_p=e(e+2)/\sigma_p$ for \cref{eq:normal}, evaluating the resulting products and differences with stable arithmetic.
Where a gradient has a genuine zero crossing, the residual cancellation is intrinsic in the working precision and is resolved by the strict mode below.

\paragraph{Strict rounding.}
\textsc{Rewrite} runs the rewritten algorithm entirely in the target precision, with compensated arithmetic. The strict variant, \textsc{Ours}, evaluates the same algorithm in FP64 and rounds once to the target format.
Rounding once matters, because FP64$\to$FP32$\to$FP16 is a double rounding (\cref{app:blocks}).
For \fkl{} we additionally guard the last rounding. If the FP64 value is closer to a rounding boundary than its a-priori error bound, the sum is re-evaluated with interval arithmetic at increasing precision, in the spirit of \citet{ziv1991}.

\begin{table*}[t]
\caption{Significant KL error rates (\%) for interface~A, across all KL bins and parameter regions. Entries are percentages of scored inputs with $\ge8$ ULP error or a qualitative failure. $N$ counts scored inputs across supported formats. \A{} is PyTorch, \textsc{Off.+} the promoted official formula, \textsc{Rewrite} our same-format formula and \textsc{Ours} the strict repair. Ranges span dimension settings. The two multivariate families lack official FP16 support. Gradient results are in \cref{app:ifaceA}.}
\label{tab:main}
\centering\small
\setlength{\tabcolsep}{2.6pt}
\begin{tabular}{lrcccccccc}
\toprule
 & & \multicolumn{4}{c}{FP16 inputs} & \multicolumn{4}{c}{FP32 inputs}\\
\cmidrule(lr){3-6}\cmidrule(lr){7-10}
Distribution & $N$ & \A & \textsc{Off.+} & \textsc{Rewrite} & \textsc{Ours} & \A & \textsc{Off.+} & \textsc{Rewrite} & \textsc{Ours}\\
\midrule
Cauchy & 345,028 & 64.05 & 10.60 & 0.003 & \textbf{0} & 88.88 & 34.84 & 0 & \textbf{0} \\
Uniform & 181,798 & 51.62 & 0 & 0 & \textbf{0} & 75.08 & 0.012 & 0 & \textbf{0} \\
Exponential & 308,000 & 72.57 & 0 & 0 & \textbf{0} & 85.64 & 33.49 & 0 & \textbf{0} \\
Poisson & 323,498 & 75.64 & 9.75 & 0.002 & \textbf{0} & 85.82 & 34.44 & 0 & \textbf{0} \\
Normal & 339,499 & 42.40 & 0 & 0 & \textbf{0} & 82.18 & 13.84 & 0 & \textbf{0} \\
Laplace & 339,692 & 58.32 & 2.24 & 1.46 & \textbf{0} & 85.87 & 31.95 & 0 & \textbf{0} \\
Pareto & 206,950 & 73.52 & 0 & 0 & \textbf{0} & 83.76 & 6.99 & 0 & \textbf{0} \\
Bernoulli (probs) & 327,099 & 61.43 & 0.29 & 0.41 & \textbf{0} & 81.06 & 27.15 & 0 & \textbf{0} \\
Bernoulli (logits) & 353,392 & 49.92 & 1.29 & 0.012 & \textbf{0} & 75.95 & 25.48 & 0.012 & \textbf{0} \\
Binomial (probs) & 329,885 & 83.86 & 10.43 & 0.32 & \textbf{0} & 85.60 & 37.51 & 0 & \textbf{0} \\
Binomial (logits) & 345,715 & 69.69 & 34.93 & 0.020 & \textbf{0} & 86.95 & 43.79 & 0.031 & \textbf{0} \\
Geometric (probs) & 304,861 & 91.17 & 36.53 & 0 & \textbf{0} & 94.19 & 48.35 & 0 & \textbf{0} \\
Geometric (logits) & 352,178 & 94.35 & 42.77 & 0.17 & \textbf{0} & 96.62 & 51.47 & $<$0.001 & \textbf{0} \\
Gumbel & 339,249 & 66.18 & 13.78 & 0.007 & \textbf{0} & 90.02 & 38.42 & 0 & \textbf{0} \\
Gamma & 331,067 & 78.47 & 9.09 & 0.003 & \textbf{0} & 88.67 & 36.92 & 0.002 & \textbf{0} \\
Beta & 329,156 & 85.38 & 40.32 & 2.11 & \textbf{0} & 92.62 & 43.77 & 0.65 & \textbf{0} \\
Dirichlet ($K{=}2,4,16$) & 314,451 & 84.3--89.0 & 29.6--55.9 & 0.009--2.45 & \textbf{0} & 92.1--93.8 & 41.6--49.8 & 0.003--0.69 & \textbf{0} \\
Categorical (probs; $K{=}2,4,16$) & 372,429 & 46.6--47.6 & 3.4--4.6 & $\le$0.01 & \textbf{0} & 86.0--86.9 & 32.2--32.5 & $\le$0.01 & \textbf{0} \\
Categorical (logits; $K{=}2,4,16$) & 367,746 & 59.4--62.7 & 12.4--18.3 & $<$0.01 & \textbf{0} & 88.3--90.5 & 35.8--38.4 & 0 & \textbf{0} \\
Cont.\ Bernoulli (probs) & 342,070 & 78.37 & 6.27 & 0.09 & \textbf{0} & 89.75 & 38.92 & 0.04 & \textbf{0} \\
Cont.\ Bernoulli (logits) & 376,691 & 86.17 & 27.81 & $<$0.01 & \textbf{0} & 90.89 & 42.56 & $<$0.01 & \textbf{0} \\
Multivariate Normal ($d{=}2,4$) & 202,000 & \multicolumn{4}{c}{FP16 unsupported} & 88.6--88.9 & 35.8 & 0 & \textbf{0} \\
Low-rank MVN ($d{=}2,4$; $r{=}1,2$) & 202,000 & \multicolumn{4}{c}{FP16 unsupported} & 91.1--93.4 & 39.5--41.9 & 0 & \textbf{0} \\
\bottomrule\relax  
\end{tabular}
\end{table*}

\paragraph{Results on the frozen data.} \Cref{tab:main} compares the four methods on exactly the data of \cref{sec:measure}.
The formula does most of the repair.
\textsc{Rewrite}, without promoting the input or output format, removes more than 97\% of the significant KL errors in every configuration and all of them in seven.
What remains (at most 0.7\% for FP32 and 2.5\% for FP16 inputs, 1.6\% in any gradient component, \cref{app:ifaceA}) sits mostly in gradients near their zero crossings and in the Beta and Dirichlet families.
The price is arithmetic rather than precision, at 2.4--5.9$\times$ the official call on a GPU for five of seven timed families (\cref{app:cost}).
Conversely, \textsc{Official+} shows that precision without the formula leaves up to half of the FP32 values wrong.
The strict mode removes the residue. \textsc{Ours} has no significant error in the KL value or in any gradient component on any of the 7,234,454 scored samples, returns exactly zero value and (where it exists) gradient at every identity sample, and never returns a negative KL or a NaN.

For interface~B (942,200 batches), \textsc{Ours} is correctly rounded on all 83,899,496 forward and 483,279,156 backward scored components, against 52.5\% and 74.9\% for \A, with no regression.

\paragraph{From logits.} For teacher logits $a$ and student logits $c$, we compute $\KL(\operatorname{softmax}(a)\,\|\,\operatorname{softmax}(c))$ directly from logits.
Let $q=\operatorname{softmax}(c)$ be computed in FP64, choose $r=\arg\max_j c_j$, and form the compensated, centred differences $\delta_j=(a_j-c_j)-(a_r-c_r)$ (\cref{app:cat}),
\begin{equation}
\begin{aligned}
\Delta&=\lgp\Big(\sum\nolimits_jq_j\,\emo(\delta_j)\Big),\\
w_j&=\delta_j-\Delta,\\
\KL&=\sum\nolimits_jq_j\big(w_je^{w_j}-\emo(w_j)\big),\\
\partial\KL/\partial c_j&=-q_j\,\emo(w_j).
\end{aligned}
\label{eq:logits}
\end{equation}
Here $p_j=q_je^{w_j}$ is the teacher probability and $h(w)=we^w-\emo(w)\ge0$ for every real $w$. Near zero $h$ is evaluated by its series $\sum_{n\ge2}(n-1)w^n/n!$, and at $\delta=0$ the KL and its logit gradients are exactly zero.
This \emph{logits path}, a row-vectorised form of the \texttt{Categorical(logits)} repair of \cref{tab:main}, is what the distillation and language-model experiments use. It is tested against high-precision references and is closely related to the fused soft-target loss of \citet{barron2026}.
\section{Downstream Experiments}
\label{sec:exp}

Do the wrong numbers matter? Four experiments, ordered by scale, compare the official implementation with the repaired one in tasks where nothing else changes.

\subsection{Parameter Fitting}
\label{sec:exp1}

\paragraph{Setup.} Fitting one distribution to another ends where loss and gradient must both vanish, the regime of \cref{sec:measure}.
We fit $\mathrm{Geo}(p)$ to a fixed $\mathrm{Geo}(p^\star)$ by optimising the logit $\theta$, with $p=\sigma(\theta)$, using Adam (learning rate $10^{-2}$, 3000 steps) from 12 shared starts, for $p^\star\in\{0.9,0.3,0.01,0.001\}$, FP16 and FP32, both parameterisations and both KL directions (32 settings), with $\theta$ kept in FP32 and the true KL computed with 60-digit arithmetic.

\paragraph{Result.} The official loss can miss a target that it is given exactly.
In FP16 with $p^\star=0.001$ and $\KL(\mathrm{Geo}(p)\|\mathrm{Geo}(p^\star))$ it reaches the target from none of the 12 starts (median relative deviation of the fitted $p$ 95\%, true KL up to 1.7), while the repaired loss converges to the target bit pattern from all 12.
In FP32, for $p^\star\le0.01$, it stalls at relative deviations of up to $4.3\times10^{-4}$, almost a hundred times further out than the repaired one.
The reason is visible at the optimum itself, where the official code returns a non-zero loss or gradient in 28 of 32 settings (e.g.\ loss 0.098 and gradient $-100$ for FP16, $p^\star=0.001$). The repair returns exact zeros in all 32.
The official runs do not converge to a biased point but wander (\cref{fig:exp1full}).

\subsection{Sequential Bayesian Design}
\label{sec:exp2beta}

\paragraph{Setup.} Bayesian experimental design chooses the next observation by its expected information gain (EIG), the expected KL between the posterior after the observation and the current one \citep{lindley1956,chaloner1995}.
We take $K=10$ Bernoulli arms with Beta posteriors. In every round the arm with the largest EIG receives the next $m$ observations, 100{,}000 in total, for $m\in\{1,100,1000,10^4\}$, from a cold and from a warm start, with 5 seeds.
The KL is the official FP32 \texttt{kl\_divergence(Beta, Beta)} or our rewrite, also in FP32. The reference policy uses the exact EIG. A choice is counted as wrong if its exact EIG is more than 0.1\% below the maximum.
The experiment drives itself into the failing range, because the gain of one more observation shrinks with the number already made, and for $\mathrm{Beta}(n{+}1,n)$ against $\mathrm{Beta}(n,n)$ with $n=10^5$ the true KL is $2.5\times10^{-6}$ and the official code returns $+0.31$.

\paragraph{Result.} The official KL misdirects the experiment: in seven of the eight settings 40--91\% of the rounds choose a wrong arm, and 35--60\% of the EIG values it computes there are zero or negative (\cref{tab:exp2beta_full}).
With the rewrite no round is wrong in any of the 40 runs.
The cost depends on the feedback.
When the posterior is updated after every observation ($m=1$), a wrong choice costs one observation, and the final posterior variance is only 1--10\% higher than with the exact policy.
When a choice commits a batch, the loss can be much larger. For $m=10^4$ from a warm start the total variance is 1.2 to 5.2 times and that of the worst arm up to 11 times the exact policy's, because an arm with few observations is passed over in most rounds.

\subsection{Identical-Start Distillation}
\label{sec:exp2}

\paragraph{Setup.} A student MLP ($32\to256\to V$) is initialised as an exact copy of its teacher and trained for 300 steps with the forward KL as the only loss.
The exact loss and gradient are zero, so any movement is numerical.
We run the full grid of vocabulary $V\in\{1024,32768,151936\}$, precision $\in\{$FP32, FP16, BF16$\}$, optimiser and learning rate (Adam at five rates, SGD at three), four losses and 10 seeds, 2,880 runs in total.
The official losses are \fkl{} on \texttt{softmax}/\texttt{log\_softmax} outputs (with and without \texttt{log\_target}) and the official \texttt{Categorical(logits)} KL. Ours is the logits path.

\begin{table*}[t]
\caption{Top-1 disagreement (\%) between student and teacher after 300 steps from an identical initialisation (median over 10 seeds). Each cell gives official \fkl{} / \textbf{ours}. The other two official variants agree with the first within 2.1 points in every cell. ``\textbf{0}'' means that no weight changed in any seed.}
\label{tab:exp2}
\centering\small
\setlength{\tabcolsep}{5pt}
\begin{tabular}{rlcccccccc}
\toprule
 & & \multicolumn{5}{c}{Adam, learning rate} & \multicolumn{3}{c}{SGD, learning rate}\\
\cmidrule(lr){3-7}\cmidrule(lr){8-10}
$V$ & dtype & $10^{-4}$ & $3{\times}10^{-4}$ & $10^{-3}$ & $3{\times}10^{-3}$ & $10^{-2}$ & $0.01$ & $0.1$ & $1$\\
\midrule
1,024 & FP32 & 0.3 / \textbf{0} & 1.0 / \textbf{0} & 4.1 / \textbf{0} & 11.9 / \textbf{0} & 31.4 / \textbf{0} & 0.0 / \textbf{0} & 0.0 / \textbf{0} & 0.0 / \textbf{0}\\
1,024 & FP16 & 0.5 / \textbf{0} & 1.4 / \textbf{0} & 3.8 / \textbf{0} & 11.5 / \textbf{0} & 29.9 / \textbf{0} & 0.0 / \textbf{0} & 0.0 / \textbf{0} & 0.0 / \textbf{0}\\
1,024 & BF16 & 0.7 / \textbf{0} & 1.4 / \textbf{0} & 4.1 / \textbf{0} & 12.1 / \textbf{0} & 30.0 / \textbf{0} & 0.0 / \textbf{0} & 0.0 / \textbf{0} & 0.2 / \textbf{0}\\
32,768 & FP32 & 0.7 / \textbf{0} & 1.7 / \textbf{0} & 6.9 / \textbf{0} & 16.7 / \textbf{0} & 40.5 / \textbf{0} & 0.0 / \textbf{0} & 0.0 / \textbf{0} & 0.0 / \textbf{0}\\
32,768 & FP16 & 2.1 / \textbf{0} & 4.3 / \textbf{0} & 9.8 / \textbf{0} & 19.3 / \textbf{0} & 41.0 / \textbf{0} & 0.0 / \textbf{0} & 0.0 / \textbf{0} & 0.4 / \textbf{0}\\
32,768 & BF16 & 1.7 / \textbf{0} & 2.5 / \textbf{0} & 6.7 / \textbf{0} & 17.0 / \textbf{0} & 39.7 / \textbf{0} & 0.0 / \textbf{0} & 0.0 / \textbf{0} & 0.2 / \textbf{0}\\
151,936 & FP32 & 0.8 / \textbf{0} & 1.6 / \textbf{0} & 6.4 / \textbf{0} & 19.0 / \textbf{0} & 47.2 / \textbf{0} & 0.0 / \textbf{0} & 0.0 / \textbf{0} & 0.0 / \textbf{0}\\
151,936 & FP16 & 6.2 / \textbf{0} & 15.1 / \textbf{0} & 36.7 / \textbf{0} & 76.3 / \textbf{0} & 97.4 / \textbf{0} & 0.2 / \textbf{0} & 1.5 / \textbf{0} & 2.1 / \textbf{0}\\
151,936 & BF16 & 1.5 / \textbf{0} & 2.7 / \textbf{0} & 7.5 / \textbf{0} & 19.1 / \textbf{0} & 46.1 / \textbf{0} & 0.0 / \textbf{0} & 0.0 / \textbf{0} & 0.1 / \textbf{0}\\
\bottomrule
\end{tabular}
\end{table*}

\paragraph{Result.} With the official losses the student moves in all 2,160 runs. With ours it does not move in any of the 720 runs. Every recorded weight change, the initial loss and the initial gradient norm are exactly zero (\cref{tab:exp2}).
The optimiser determines how this small numerical error is translated into a parameter update.
In FP32 the spurious initial gradient has a norm of about $5\times10^{-8}$. SGD scales its step with this norm and barely moves, whereas Adam normalises it to a full-size step \citep{kingma2015}, so that the drift grows with the learning rate, from below 1\% to 31--47\% top-1 disagreement in FP32.
The largest vocabulary is worse than the smallest in every Adam comparison, most strongly in FP16, where the disagreement reaches 97.4\% and the final true KL 3.6 nats.
A visible symptom is the one users report. A quarter of the official runs show a negative loss at step~0.

\paragraph{Attribution.} The three official variants exhibit similar drift. In the paired FP16 and FP32 tests, replacing \fkl{} alone yields student weights bit-identical to the official runs. The spurious parameter gradient arises in \texttt{log\_softmax} backward, which the logits path avoids \citep[see also][]{barron2026}.

\subsection{KL in Language Models}
\label{sec:exp3}

\paragraph{Setup.} We use Qwen2.5-1.5B-Instruct \citep{qwen25} (vocabulary 151,936, FP16 forward) on 16,384 WikiText-2 tokens \citep{merity2017} and compute $\KL(\text{original}\,\|\,\text{quantized})$ per token.
Part~A evaluates six quantized versions (round-to-nearest, RTN, at 8, 4 and 3 bits, GPTQ \citep{frantar2023} at 8 and 4, AWQ \citep{lin2024} at 4).
Part~B ranks the 28 decoder layers by the KL caused by quantizing one layer alone, and uses the ranking to keep the $k\in\{4,8\}$ most sensitive layers in FP16, a standard mixed-precision heuristic \citep{dong2019,cai2020}.
A decision counts as harmful if it raises the true KL of the mixed model. References are computed in FP64.

\begin{table}[t]
\caption{KL errors and layer rankings at 8-bit quantization (FP16). Part~A reports signed relative error (\%) in mean KL and the percentage of tokens with negative or infinite KL (GPTQ / RTN). Part~B reports inverted layer pairs out of 378 and Kendall's $\tau_b$ (1 indicates perfect agreement). ``B fix'' repairs \fkl{} on the same rounded probability inputs.}
\label{tab:exp3}
\centering\footnotesize
\setlength{\tabcolsep}{2.5pt}
\resizebox{\columnwidth}{!}{%
\begin{tabular}{lcccc}
\toprule
 & \multicolumn{2}{c}{Part A} & \multicolumn{2}{c}{Part B}\\
\cmidrule(lr){2-3}\cmidrule(lr){4-5}
Method & mean err.\ (\%) & tokens (\%) & inv. & $\tau_b$\\
\midrule
\fkl & $-19.0$ / $-4.4$ & 30.2 / 12.3 neg. & 19 & 0.898\\
\fkl, \texttt{log\_target} & $<1$ & -- & 35 & 0.815\\
\fkl, \texttt{mean} & reports 0 & -- & \multicolumn{2}{c}{378 ties}\\
\texttt{Categorical} & $\infty$ & 98.0 / 97.1 inf & \multicolumn{2}{c}{not rankable}\\
B fix & $+16.5$ / $+4.4$ & 16.9 / 7.5 neg. & 20 & 0.894\\
\textbf{Logits path} & \textbf{0.0 / 0.0} & \textbf{0 neg., 0 inf} & \textbf{0} & \textbf{1.000}\\
\bottomrule
\end{tabular}}
\end{table}

\begin{table}[t]
\caption{KL between cached decoding and full forward evaluation of the same model. The first numeric column is the FP64 reference KL, and all remaining values are signed relative errors (\%) at model precision.}
\label{tab:exp3c}
\centering\footnotesize
\setlength{\tabcolsep}{3pt}
\resizebox{\columnwidth}{!}{%
\begin{tabular}{lccccc}
\toprule
 & true KL & \fkl & \texttt{log\_target} & \texttt{Categorical} & \textbf{logits path}\\
\midrule
FP32 & $8.1\times10^{-11}$ & $-2304$ & $+1967$ & $-1413$ & \textbf{0.0}\\
FP16 & $1.8\times10^{-5}$ & $-684$ & $-17$ & non-finite & \textbf{0.0}\\
BF16 & $1.1\times10^{-3}$ & $-19$ & $-2.1$ & $+79$ & \textbf{0.0}\\
\bottomrule
\end{tabular}}
\end{table}

\begin{table*}[t]
\caption{Circuit edge-score accuracy against the FP64 KL of the same stored logits. ``Wrong'' is the percentage of positive-reference edges with negative scores or relative error above 10\%, and ``negative'' gives their count. Spearman measures ranking agreement, and top-1,000 overlap is the fraction of reference top edges recovered. For GPT-2 medium a stratified 13\% sample of the edges is scored. Detailed values appear in \cref{tab:exp3d}.}
\label{tab:exp3d_main}
\centering\footnotesize
\setlength{\tabcolsep}{5pt}
\begin{tabular}{llrlrrccrcc}
\toprule
 & & & & \multicolumn{4}{c}{library} & \multicolumn{3}{c}{\textbf{logits path}}\\
\cmidrule(lr){5-8}\cmidrule(lr){9-11}
task & model & edges & precision & wrong & negative & Spearman & top 1{,}000 & wrong & Spearman & top 1{,}000\\
\midrule
IOI & GPT-2 small & 32{,}477 & FP32 & 19.3\% & 1{,}449 & 1.00 & 1.000 & \textbf{0} & \textbf{1.00} & \textbf{1.000}\\
 & &  & FP16 & 81.4\% & 3 & 0.74 & 0.963 & \textbf{0} & \textbf{1.00} & \textbf{1.000}\\
 & &  & BF16 & 2.5\% & 3 & 0.99 & 0.849 & \textbf{0} & \textbf{1.00} & \textbf{0.976}\\
\addlinespace
Greater-than & GPT-2 small & 32{,}477 & FP32 & 45.8\% & 3{,}271 & 0.97 & 1.000 & \textbf{0} & \textbf{1.00} & \textbf{1.000}\\
 & & 32{,}476 & FP16 & 6.8\% & 2 & 0.95 & 0.698 & \textbf{0.02\%} & \textbf{1.00} & \textbf{0.999}\\
\addlinespace
Docstring & attn-only-4l & 1{,}272 & FP32 & 2.4\% & 0 & 1.00 & 1.000 & \textbf{0} & \textbf{1.00} & \textbf{1.000}\\
 & &  & FP16 & 54.2\% & 10 & 0.98 & 0.959 & \textbf{0} & \textbf{1.00} & \textbf{1.000}\\
\addlinespace
IOI & GPT-2 medium & 29{,}999 & FP32 & 58.8\% & 4{,}401 & 0.94 & 1.000 & \textbf{0} & \textbf{1.00} & \textbf{1.000}\\
 & &  & FP16 & 81.1\% & 338 & 0.47 & 0.375 & \textbf{0.01\%} & \textbf{1.00} & \textbf{0.999}\\
\bottomrule\relax  
\end{tabular}
\end{table*}

\paragraph{Part A: KL Accuracy.} The official KL is least reliable for the quantized models closest to the original.
For the two near-lossless 8-bit models the official \fkl{} misreports the mean KL by $-19.0\%$ and $-4.4\%$, with 30.2\% and 12.3\% of the tokens negative (\cref{tab:exp3}), and with \texttt{reduction="mean"} it reports exactly 0.
\texttt{Categorical(logits)} returns $+\infty$ for almost every token, because about 38\% of the FP16 \texttt{softmax} probabilities underflow to zero.
Repairing the operator alone is not enough. The repaired \fkl{} is correctly rounded for its inputs on every token, yet 98.8\% and 96.8\% of the tokens remain off by $\ge1\%$, because the error is already in the FP16 \texttt{softmax} outputs.
The logits path equals the rounded FP64 reference on every token of every model in FP16 and FP32.

\paragraph{Part B: Layer Ranking.} A ranking depends on differences between small numbers, and here it breaks.
Single-layer divergences at 8 bits are only $3.6\times10^{-5}$ to $2.6\times10^{-4}$. The official FP16 \fkl{} reports a negative mean KL for 24 of 28 layers and inverts 19 of the 378 layer pairs (up to 158 for the BF16 variants), while \texttt{reduction="mean"} reports 0 for all layers and \texttt{Categorical} is infinite, so neither can rank at all.
The logits path has no inversion in FP16 or FP32.
The decision itself changes at the boundary. For $k=8$, several configurations protect one wrong layer, and the BF16 \texttt{Categorical} KL protects four, which raises the true KL of the resulting mixed model by 9.8\% on the disjoint evaluation set.

\paragraph{Part C: Execution Gaps.} The KL also measures how far two executions of one model are apart, such as rollout and training in RL fine-tuning \citep{qi2025}.
Each precision produces gaps of the size that its own KL cannot resolve (\cref{tab:exp3c}).
For cached decoding against a full forward pass the official \fkl{} is off by more than 600\% in FP32 and FP16 and by 19\% in BF16, the logits path by less than 0.05\% on all 11 pairs.
Nor should a reported KL depend on the batch size, yet in FP16 with the default reduction (\texttt{mean}) the KL to an 8-bit copy is reported as anything between 0 and 58\% of the truth, depending on the number of rows per call, because the mean underflows, while the logits path varies by less than 0.5\% (\cref{app:downstream}).

\paragraph{Part D: Circuit Scores.} Circuit discovery scores each edge of a model by the KL between the clean output and the output with that edge ablated \citep{conmy2023}. Most edges matter little, so most of these scores are tiny, and many of them fall in the failing range.
We score edges one at a time with the ablation loop of the \texttt{auto-circuit} library \citep{miller2024}: IOI \citep{wang2023ioi} and Greater-than \citep{hanna2023} on GPT-2 small, Docstring \citep{heimersheim2023} on a 4-layer attention-only model, and IOI on GPT-2 medium.

In FP32, the default precision, the library's score is wrong for 19\%, 46\% and 59\% of the edges in the three GPT-2 settings, and thousands of scores are negative (\cref{tab:exp3d_main}), while the logits path is wrong for none.
The errors sit in the scores below about $10^{-6}$, so they change which small edges pass a threshold while the ranking of the largest edges changes little, and the larger model is affected more because its scores are smaller (median $3\times10^{-8}$ against $9\times10^{-7}$).
In FP16 the ranking suffers as well. The score is wrong for 54--81\% of the edges in three of the four settings, and on GPT-2 medium only 38\% of the 1{,}000 highest-scoring edges are correct, against 99.9\% for the logits path.
Wrong-score rates alone do not determine ranking stability. In Docstring FP32, 2.4\% wrong scores leave the top-1,000 set unchanged, whereas in GPT-2 small IOI with BF16, 2.5\% wrong scores yield a top-1,000 overlap of only 0.849.
\section{Related Work}
\label{sec:related}

The work closest to ours is that of \citet{barron2026}, who identify two bugs in standard \texttt{softmax}/cross-entropy code, \emph{Softmax Collapse} (first diagnosed by \citet{prieto2025}) and \emph{Softmax Cancellation}, and show that a fused implementation stops distillation drift.
Our distillation experiment (\cref{sec:exp2}) observes the same drift independently, and our logits path is closely related to their soft-target loss.
They treat the categorical soft-target loss and its training dynamics. We measure and repair the KL interfaces themselves, across 18 families and with certified ground truth, separate operator from input-representation error, and study evaluation and decisions as well as training.
Like other work that traces a training failure to one floating-point mechanism \citep{prieto2025,mishne2023,wortsman2023}, we isolate a mechanism, but a mostly silent one. Studies and detectors of numerical defects in deep-learning software \citep{zhang2018,humbatova2020,wang2022,kloberdanz2022,zhang2020,yan2021} focus on exceptional values.
Our tools are those of classical numerical analysis \citep{higham2002,sterbenz1974,dekker1971,shewchuk1997,ogita2005,ziv1991,johansson2017}. Automatic rewriting \citep{panchekha2015} targets local rewrites, whereas those needed here are global and must also deliver gradients.
Low-precision training and inference \citep{micikevicius2018,kalamkar2019,qi2025,yuan2025} make the regime we study routine (see \cref{app:related}).
\section{Limitations and Conclusion}
\label{sec:conclusion}

\paragraph{Limitations.}
Error rates are conditional on a sampling design that oversamples small divergences and stress regions on purpose.
We evaluate PyTorch 2.15.0 on CPU and 2.4.0 on an A100 GPU.
The repairs are unoptimised reference implementations. On a GPU the rewrite costs 2.4--5.9$\times$ the official call for five closed forms, and the repairs of Gamma, Beta and \fkl{} far more (\cref{app:cost}). The BF16 results are exploratory.
The first three experiments are synthetic or simulated, the training effect is shown for an identical start only, and in the layer-ranking experiment the most common official configuration still made the correct selection.
\Cref{app:null} lists settings with no measured effect.

\paragraph{Conclusion.}
Negative KL outputs reveal a broader numerical failure: many inaccurate KL values remain non-negative, finite and unnoticed.
The cause is structural: the divergence between nearby distributions is typically a second-order quantity assembled from larger terms.
For the closed forms, higher precision alone does not remove the resulting errors. Rewriting does.
The practical advice is to compute divergences from parameters or logits rather than rounded probabilities, sum non-negative terms, and add precision only where correct rounding is required.

\newpage
\section*{Impact Statement}
This paper aims to make a basic numerical primitive of machine learning more reliable. We do not foresee specific negative societal consequences of this work.

\section*{AI Disclosure Statement}
ChatGPT and Claude were used to assist with coding, debugging, technical discussions, and grammar checking.
All mathematical derivations, implementations, and experimental results were verified by the authors, who take full responsibility for the work.

\bibliography{refs}

@article{kullback1951,
  author = {Kullback, Solomon and Leibler, Richard A.},
  title = {On Information and Sufficiency},
  journal = {The Annals of Mathematical Statistics},
  volume = {22}, number = {1}, pages = {79--86}, year = {1951}
}

@article{goldberg1991,
  author = {Goldberg, David},
  title = {What Every Computer Scientist Should Know About Floating-Point Arithmetic},
  journal = {ACM Computing Surveys},
  volume = {23}, number = {1}, pages = {5--48}, year = {1991}
}

@book{higham2002,
  author = {Higham, Nicholas J.},
  title = {Accuracy and Stability of Numerical Algorithms},
  edition = {2nd}, publisher = {SIAM}, year = {2002}
}

@book{muller2018,
  author = {Muller, Jean-Michel and Brunie, Nicolas and de Dinechin, Florent and Jeannerod, Claude-Pierre and Joldes, Mioara and Lef{\`e}vre, Vincent and Melquiond, Guillaume and Revol, Nathalie and Torres, Serge},
  title = {Handbook of Floating-Point Arithmetic},
  edition = {2nd}, publisher = {Birkh{\"a}user}, year = {2018}
}

@book{sterbenz1974,
  author = {Sterbenz, Pat H.},
  title = {Floating-Point Computation},
  publisher = {Prentice-Hall}, year = {1974}
}

@article{dekker1971,
  author = {Dekker, Theodorus J.},
  title = {A Floating-Point Technique for Extending the Available Precision},
  journal = {Numerische Mathematik},
  volume = {18}, number = {3}, pages = {224--242}, year = {1971}
}

@book{knuth1997,
  author = {Knuth, Donald E.},
  title = {The Art of Computer Programming, Volume 2: Seminumerical Algorithms},
  edition = {3rd}, publisher = {Addison-Wesley}, year = {1997}
}

@article{shewchuk1997,
  author = {Shewchuk, Jonathan Richard},
  title = {Adaptive Precision Floating-Point Arithmetic and Fast Robust Geometric Predicates},
  journal = {Discrete \& Computational Geometry},
  volume = {18}, number = {3}, pages = {305--363}, year = {1997}
}

@article{ogita2005,
  author = {Ogita, Takeshi and Rump, Siegfried M. and Oishi, Shin'ichi},
  title = {Accurate Sum and Dot Product},
  journal = {SIAM Journal on Scientific Computing},
  volume = {26}, number = {6}, pages = {1955--1988}, year = {2005}
}

@article{kahan1965,
  author = {Kahan, William},
  title = {Pracniques: Further Remarks on Reducing Truncation Errors},
  journal = {Communications of the ACM},
  volume = {8}, number = {1}, pages = {40}, year = {1965}
}

@article{ziv1991,
  author = {Ziv, Abraham},
  title = {Fast Evaluation of Elementary Mathematical Functions with Correctly Rounded Last Bit},
  journal = {ACM Transactions on Mathematical Software},
  volume = {17}, number = {3}, pages = {410--423}, year = {1991}
}

@inproceedings{sibidanov2022,
  author = {Sibidanov, Alexei and Zimmermann, Paul and Glondu, St{\'e}phane},
  title = {The {CORE-MATH} Project},
  booktitle = {IEEE Symposium on Computer Arithmetic (ARITH)}, year = {2022}
}

@article{blanchard2021,
  author = {Blanchard, Pierre and Higham, Desmond J. and Higham, Nicholas J.},
  title = {Accurately Computing the Log-Sum-Exp and Softmax Functions},
  journal = {IMA Journal of Numerical Analysis},
  volume = {41}, number = {4}, pages = {2311--2330}, year = {2021}
}

@inproceedings{panchekha2015,
  author = {Panchekha, Pavel and Sanchez-Stern, Alex and Wilcox, James R. and Tatlock, Zachary},
  title = {Automatically Improving Accuracy for Floating Point Expressions},
  booktitle = {ACM SIGPLAN Conference on Programming Language Design and Implementation (PLDI)}, year = {2015}
}

@article{johansson2017,
  author = {Johansson, Fredrik},
  title = {Arb: Efficient Arbitrary-Precision Midpoint-Radius Interval Arithmetic},
  journal = {IEEE Transactions on Computers},
  volume = {66}, number = {8}, pages = {1281--1292}, year = {2017}
}

@misc{mpmath,
  author = {{The mpmath development team}},
  title = {mpmath: a {P}ython library for arbitrary-precision floating-point arithmetic},
  howpublished = {\url{https://mpmath.org/}}, year = {2023}
}

@misc{ieee754,
  author = {{IEEE}},
  title = {{IEEE} Standard for Floating-Point Arithmetic},
  howpublished = {IEEE Std 754-2019}, year = {2019}
}

@inproceedings{micikevicius2018,
  author = {Micikevicius, Paulius and Narang, Sharan and Alben, Jonah and Diamos, Gregory and Elsen, Erich and Garcia, David and Ginsburg, Boris and Houston, Michael and Kuchaiev, Oleksii and Venkatesh, Ganesh and Wu, Hao},
  title = {Mixed Precision Training},
  booktitle = {International Conference on Learning Representations (ICLR)}, year = {2018}
}

@article{kalamkar2019,
  author = {Kalamkar, Dhiraj and Mudigere, Dheevatsa and Mellempudi, Naveen and Das, Dipankar and Banerjee, Kunal and Avancha, Sasikanth and Vooturi, Dharma Teja and Jammalamadaka, Nataraj and Huang, Jianyu and Yuen, Hector and others},
  title = {A Study of {BFLOAT16} for Deep Learning Training},
  journal = {arXiv preprint arXiv:1905.12322}, year = {2019}
}

@article{zamirai2020,
  author = {Zamirai, Pedram and Zhang, Jian and Aberger, Christopher R. and De Sa, Christopher},
  title = {Revisiting {BFloat16} Training},
  journal = {arXiv preprint arXiv:2010.06192}, year = {2020}
}

@inproceedings{paszke2019,
  author = {Paszke, Adam and Gross, Sam and Massa, Francisco and Lerer, Adam and Bradbury, James and Chanan, Gregory and Killeen, Trevor and Lin, Zeming and Gimelshein, Natalia and Antiga, Luca and others},
  title = {{PyTorch}: An Imperative Style, High-Performance Deep Learning Library},
  booktitle = {Advances in Neural Information Processing Systems (NeurIPS)}, year = {2019}
}

@article{dillon2017,
  author = {Dillon, Joshua V. and Langmore, Ian and Tran, Dustin and Brevdo, Eugene and Vasudevan, Srinivas and Moore, Dave and Patton, Brian and Alemi, Alex and Hoffman, Matt and Saurous, Rif A.},
  title = {{TensorFlow} Distributions},
  journal = {arXiv preprint arXiv:1711.10604}, year = {2017}
}

@article{bingham2019,
  author = {Bingham, Eli and Chen, Jonathan P. and Jankowiak, Martin and Obermeyer, Fritz and Pradhan, Neeraj and Karaletsos, Theofanis and Singh, Rohit and Szerlip, Paul and Horsfall, Paul and Goodman, Noah D.},
  title = {Pyro: Deep Universal Probabilistic Programming},
  journal = {Journal of Machine Learning Research},
  volume = {20}, number = {28}, pages = {1--6}, year = {2019}
}

@article{carpenter2017,
  author = {Carpenter, Bob and Gelman, Andrew and Hoffman, Matthew D. and Lee, Daniel and Goodrich, Ben and Betancourt, Michael and Brubaker, Marcus and Guo, Jiqiang and Li, Peter and Riddell, Allen},
  title = {Stan: A Probabilistic Programming Language},
  journal = {Journal of Statistical Software},
  volume = {76}, number = {1}, year = {2017}
}

@misc{bradbury2018,
  author = {Bradbury, James and Frostig, Roy and Hawkins, Peter and Johnson, Matthew James and Leary, Chris and Maclaurin, Dougal and Necula, George and Paszke, Adam and VanderPlas, Jake and Wanderman-Milne, Skye and Zhang, Qiao},
  title = {{JAX}: Composable Transformations of {P}ython+{N}um{P}y Programs},
  howpublished = {\url{http://github.com/jax-ml/jax}}, year = {2018}
}

@article{hinton2015,
  author = {Hinton, Geoffrey and Vinyals, Oriol and Dean, Jeff},
  title = {Distilling the Knowledge in a Neural Network},
  journal = {arXiv preprint arXiv:1503.02531}, year = {2015}
}

@inproceedings{kim2021,
  author = {Kim, Taehyeon and Oh, Jaehoon and Kim, Nak Yil and Cho, Sangwook and Yun, Se-Young},
  title = {Comparing {K}ullback-{L}eibler Divergence and Mean Squared Error Loss in Knowledge Distillation},
  booktitle = {International Joint Conference on Artificial Intelligence (IJCAI)}, year = {2021}
}

@inproceedings{agarwal2024,
  author = {Agarwal, Rishabh and Vieillard, Nino and Zhou, Yongchao and Stanczyk, Piotr and Ramos, Sabela and Geist, Matthieu and Bachem, Olivier},
  title = {On-Policy Distillation of Language Models: Learning from Self-Generated Mistakes},
  booktitle = {International Conference on Learning Representations (ICLR)}, year = {2024}
}

@inproceedings{gu2024,
  author = {Gu, Yuxian and Dong, Li and Wei, Furu and Huang, Minlie},
  title = {{MiniLLM}: On-Policy Distillation of Large Language Models},
  booktitle = {International Conference on Learning Representations (ICLR)}, year = {2024}
}

@inproceedings{kingma2014,
  author = {Kingma, Diederik P. and Welling, Max},
  title = {Auto-Encoding Variational {B}ayes},
  booktitle = {International Conference on Learning Representations (ICLR)}, year = {2014}
}

@inproceedings{higgins2017,
  author = {Higgins, Irina and Matthey, Loic and Pal, Arka and Burgess, Christopher and Glorot, Xavier and Botvinick, Matthew and Mohamed, Shakir and Lerchner, Alexander},
  title = {beta-{VAE}: Learning Basic Visual Concepts with a Constrained Variational Framework},
  booktitle = {International Conference on Learning Representations (ICLR)}, year = {2017}
}

@inproceedings{loaiza2019,
  author = {Loaiza-Ganem, Gabriel and Cunningham, John P.},
  title = {The Continuous {B}ernoulli: Fixing a Pervasive Error in Variational Autoencoders},
  booktitle = {Advances in Neural Information Processing Systems (NeurIPS)}, year = {2019}
}

@inproceedings{schulman2015,
  author = {Schulman, John and Levine, Sergey and Abbeel, Pieter and Jordan, Michael and Moritz, Philipp},
  title = {Trust Region Policy Optimization},
  booktitle = {International Conference on Machine Learning (ICML)}, year = {2015}
}

@article{schulman2017,
  author = {Schulman, John and Wolski, Filip and Dhariwal, Prafulla and Radford, Alec and Klimov, Oleg},
  title = {Proximal Policy Optimization Algorithms},
  journal = {arXiv preprint arXiv:1707.06347}, year = {2017}
}

@misc{schulman2020,
  author = {Schulman, John},
  title = {Approximating {KL} Divergence},
  howpublished = {Blog post, \url{http://joschu.net/blog/kl-approx.html}}, year = {2020}
}

@article{shao2024,
  author = {Shao, Zhihong and Wang, Peiyi and Zhu, Qihao and Xu, Runxin and Song, Junxiao and Bi, Xiao and Zhang, Haowei and Zhang, Mingchuan and Li, Y. K. and Wu, Y. and Guo, Daya},
  title = {{DeepSeekMath}: Pushing the Limits of Mathematical Reasoning in Open Language Models},
  journal = {arXiv preprint arXiv:2402.03300}, year = {2024}
}

@inproceedings{frantar2023,
  author = {Frantar, Elias and Ashkboos, Saleh and Hoefler, Torsten and Alistarh, Dan},
  title = {{GPTQ}: Accurate Post-Training Quantization for Generative Pre-trained Transformers},
  booktitle = {International Conference on Learning Representations (ICLR)}, year = {2023}
}

@inproceedings{lin2024,
  author = {Lin, Ji and Tang, Jiaming and Tang, Haotian and Yang, Shang and Chen, Wei-Ming and Wang, Wei-Chen and Xiao, Guangxuan and Dang, Xingyu and Gan, Chuang and Han, Song},
  title = {{AWQ}: Activation-aware Weight Quantization for {LLM} Compression and Acceleration},
  booktitle = {Proceedings of Machine Learning and Systems (MLSys)}, year = {2024}
}

@inproceedings{hu2022,
  author = {Hu, Edward J. and Shen, Yelong and Wallis, Phillip and Allen-Zhu, Zeyuan and Li, Yuanzhi and Wang, Shean and Wang, Lu and Chen, Weizhu},
  title = {{LoRA}: Low-Rank Adaptation of Large Language Models},
  booktitle = {International Conference on Learning Representations (ICLR)}, year = {2022}
}

@article{nagel2021,
  author = {Nagel, Markus and Fournarakis, Marios and Amjad, Rana Ali and Bondarenko, Yelysei and van Baalen, Mart and Blankevoort, Tijmen},
  title = {A White Paper on Neural Network Quantization},
  journal = {arXiv preprint arXiv:2106.08295}, year = {2021}
}

@inproceedings{dong2019,
  author = {Dong, Zhen and Yao, Zhewei and Gholami, Amir and Mahoney, Michael and Keutzer, Kurt},
  title = {{HAWQ}: Hessian AWare Quantization of Neural Networks with Mixed-Precision},
  booktitle = {IEEE/CVF International Conference on Computer Vision (ICCV)}, year = {2019}
}

@inproceedings{cai2020,
  author = {Cai, Yaohui and Yao, Zhewei and Dong, Zhen and Gholami, Amir and Mahoney, Michael W. and Keutzer, Kurt},
  title = {{ZeroQ}: A Novel Zero Shot Quantization Framework},
  booktitle = {IEEE/CVF Conference on Computer Vision and Pattern Recognition (CVPR)}, year = {2020}
}

@inproceedings{liu2024,
  author = {Liu, Zechun and Oguz, Barlas and Zhao, Changsheng and Chang, Ernie and Stock, Pierre and Mehdad, Yashar and Shi, Yangyang and Krishnamoorthi, Raghuraman and Chandra, Vikas},
  title = {{LLM-QAT}: Data-Free Quantization Aware Training for Large Language Models},
  booktitle = {Findings of the Association for Computational Linguistics (ACL)}, year = {2024}
}

@article{xin2026,
  author = {Xin, Meng and Priyadarshi, Sweta and Xin, Jingyu and Kartal, Bilal and Vavre, Aditya and others},
  title = {Quantization-Aware Distillation for {NVFP4} Inference Accuracy Recovery},
  journal = {arXiv preprint arXiv:2601.20088}, year = {2026}
}

@inproceedings{kingma2015,
  author = {Kingma, Diederik P. and Ba, Jimmy},
  title = {Adam: A Method for Stochastic Optimization},
  booktitle = {International Conference on Learning Representations (ICLR)}, year = {2015}
}

@inproceedings{prieto2025,
  author = {Prieto, Lucas and Barsbey, Melih and Mediano, Pedro A. M. and Birdal, Tolga},
  title = {Grokking at the Edge of Numerical Stability},
  booktitle = {International Conference on Learning Representations (ICLR)}, year = {2025}
}

@article{barron2026,
  author = {Barron, Jonathan T. and Gnanapragasam, Dan and Tschannen, Michael},
  title = {A Fix for Softmax and Cross-Entropy},
  journal = {arXiv preprint arXiv:2610.31642}, year = {2026}
}

@article{yuan2025,
  author = {Yuan, Jiayi and Li, Hao and Ding, Xinheng and Xie, Wenya and Li, Yu-Jhe and Zhao, Wentian and Wan, Kun and Shi, Jing and Hu, Xia and Liu, Zirui},
  title = {Give Me {FP32} or Give Me Death? {C}hallenges and Solutions for Reproducible Reasoning},
  journal = {arXiv preprint arXiv:2506.09501}, year = {2025}
}

@article{qi2025,
  author = {Qi, Penghui and Liu, Zichen and Zhou, Xiangxin and Pang, Tianyu and Du, Chao and Lee, Wee Sun and Lin, Min},
  title = {Defeating the Training-Inference Mismatch via {FP16}},
  journal = {arXiv preprint arXiv:2510.26788}, year = {2025}
}

@inproceedings{zhang2018,
  author = {Zhang, Yuhao and Chen, Yifan and Cheung, Shing-Chi and Xiong, Yingfei and Zhang, Lu},
  title = {An Empirical Study on {TensorFlow} Program Bugs},
  booktitle = {ACM SIGSOFT International Symposium on Software Testing and Analysis (ISSTA)}, year = {2018}
}

@inproceedings{humbatova2020,
  author = {Humbatova, Nargiz and Jahangirova, Gunel and Bavota, Gabriele and Riccio, Vincenzo and Stocco, Andrea and Tonella, Paolo},
  title = {Taxonomy of Real Faults in Deep Learning Systems},
  booktitle = {International Conference on Software Engineering (ICSE)}, year = {2020}
}

@inproceedings{wang2022,
  author = {Wang, Gan and Wang, Zan and Chen, Junjie and Chen, Xiang and Yan, Ming},
  title = {An Empirical Study on Numerical Bugs in Deep Learning Programs},
  booktitle = {IEEE/ACM International Conference on Automated Software Engineering (ASE)}, year = {2022}
}

@inproceedings{kloberdanz2022,
  author = {Kloberdanz, Eliska and Kloberdanz, Kyle G. and Le, Wei},
  title = {{DeepStability}: A Study of Unstable Numerical Methods and Their Solutions in Deep Learning},
  booktitle = {International Conference on Software Engineering (ICSE)}, year = {2022}
}

@inproceedings{yan2021,
  author = {Yan, Ming and Chen, Junjie and Zhang, Xiangyu and Tan, Lin and Wang, Gan and Wang, Zan},
  title = {Exposing Numerical Bugs in Deep Learning via Gradient Back-Propagation},
  booktitle = {ACM Joint European Software Engineering Conference and Symposium on the Foundations of Software Engineering (ESEC/FSE)}, year = {2021}
}

@inproceedings{zhang2020,
  author = {Zhang, Yuhao and Ren, Luyao and Chen, Liqian and Xiong, Yingfei and Cheung, Shing-Chi and Xie, Tao},
  title = {Detecting Numerical Bugs in Neural Network Architectures},
  booktitle = {ACM Joint European Software Engineering Conference and Symposium on the Foundations of Software Engineering (ESEC/FSE)}, year = {2020}
}

@article{qwen25,
  author = {Yang, An and Yang, Baosong and Zhang, Beichen and Hui, Binyuan and Zheng, Bo and Yu, Bowen and Li, Chengyuan and Liu, Dayiheng and Huang, Fei and others},
  title = {{Qwen2.5} Technical Report},
  journal = {arXiv preprint arXiv:2412.15115}, year = {2024}
}

@inproceedings{merity2017,
  author = {Merity, Stephen and Xiong, Caiming and Bradbury, James and Socher, Richard},
  title = {Pointer Sentinel Mixture Models},
  booktitle = {International Conference on Learning Representations (ICLR)}, year = {2017}
}

@misc{trl,
  author = {von Werra, Leandro and Belkada, Younes and Tunstall, Lewis and others},
  title = {{TRL}: Transformer Reinforcement Learning},
  howpublished = {\url{https://github.com/huggingface/trl}}, year = {2020}
}

@inproceedings{conmy2023,
  author = {Conmy, Arthur and Mavor-Parker, Augustine N. and Lynch, Aengus and Heimersheim, Stefan and Garriga-Alonso, Adri{\`a}},
  title = {Towards Automated Circuit Discovery for Mechanistic Interpretability},
  booktitle = {Advances in Neural Information Processing Systems}, year = {2023}
}

@inproceedings{miller2024,
  author = {Miller, Joseph and Chughtai, Bilal and Saunders, William},
  title = {Transformer Circuit Faithfulness Metrics Are Not Robust},
  booktitle = {Conference on Language Modeling (COLM)}, year = {2024}
}

@inproceedings{wang2023ioi,
  author = {Wang, Kevin and Variengien, Alexandre and Conmy, Arthur and Shlegeris, Buck and Steinhardt, Jacob},
  title = {Interpretability in the Wild: a Circuit for Indirect Object Identification in {GPT}-2 Small},
  booktitle = {International Conference on Learning Representations}, year = {2023}
}

@article{lindley1956,
  author = {Lindley, D. V.},
  title = {On a Measure of the Information Provided by an Experiment},
  journal = {The Annals of Mathematical Statistics}, volume = {27}, number = {4}, pages = {986--1005}, year = {1956}
}

@article{chaloner1995,
  author = {Chaloner, Kathryn and Verdinelli, Isabella},
  title = {Bayesian Experimental Design: A Review},
  journal = {Statistical Science}, volume = {10}, number = {3}, pages = {273--304}, year = {1995}
}

@inproceedings{hanna2023,
  author = {Hanna, Michael and Liu, Ollie and Variengien, Alexandre},
  title = {How Does {GPT}-2 Compute Greater-Than?: Interpreting Mathematical Abilities in a Pre-Trained Language Model},
  booktitle = {Advances in Neural Information Processing Systems}, year = {2023}
}

@misc{heimersheim2023,
  author = {Heimersheim, Stefan and Janiak, Jett},
  title = {A Circuit for {P}ython Docstrings in a 4-Layer Attention-Only Transformer},
  howpublished = {AI Alignment Forum, \url{https://www.alignmentforum.org/posts/u6KXXmKFbXfWzoAXn}}, year = {2023}
}

@inproceedings{mishne2023,
  author = {Mishne, Gal and Wan, Zhengchao and Wang, Yusu and Yang, Sheng},
  title = {The Numerical Stability of Hyperbolic Representation Learning},
  booktitle = {International Conference on Machine Learning (ICML)}, year = {2023}
}

@inproceedings{wortsman2023,
  author = {Wortsman, Mitchell and Dettmers, Tim and Zettlemoyer, Luke and Morcos, Ari and Farhadi, Ali and Schmidt, Ludwig},
  title = {Stable and Low-Precision Training for Large-Scale Vision-Language Models},
  booktitle = {Advances in Neural Information Processing Systems (NeurIPS)}, year = {2023}
}
\bibliographystyle{icml2027}

\newpage
\appendix
\crefalias{section}{appendix}
\crefalias{subsection}{appendix}
\crefalias{subsubsection}{appendix}
\onecolumn
\section{Measurement and Validation Protocol}
\label{app:protocol}

\paragraph{Sampling (interface A).} For each configuration and precision, samples are organised in \emph{cells} indexed by a parameter region and a KL group.
Regions: R0 (ordinary parameters), and stress regions R1, R2 (and R3 for some families) with very large or very small parameters or probabilities close to 0 or 1; for example, for Dirichlet the concentration components are log-uniform in $[10^{-2},10^{2}]$ (R0), $[10^{2},10^{4}]$ (R1) and $[10^{-4},10^{-2}]$ (R2).
KL groups: a control bin $[10^{-1},1)$, fourteen bins $[10^{-k-1},10^{-k})$, $k=1,\dots,14$, a ULP group ($Q$ obtained by moving parameters of $P$ by 1--8 ULP) and an identity group ($Q=P$).
The quota is 10,000 samples per R0 cell and per ULP cell and 1,000 per stress cell.
A cell is recorded as \emph{empty} after 2,000 consecutive failed attempts and as \emph{below quota} when the quota is not reached within 50 times the quota in attempts; the dominant reason is that $Q$ rounds to $P$ in the target format.
For the 16 scalar-family configurations (5,057,067 samples), 1,194 of 1,568 cells are full, 49 below quota and 325 empty (285 of them in FP16); the other 7 configurations (Categorical, ContinuousBernoulli, Dirichlet and the two multivariate normal families, with their sub-configurations; 2,485,402 samples) follow identical rules. Of the 7,542,469 samples, 7,234,454 are scored: the 308,015 FP16 samples of the two multivariate normal families are not scored for any method, because the official implementation does not run on them (\cref{tab:main}). A call of the official implementation that raises an error on a valid input counts as a significant error.

\paragraph{Truth and scoring.} For interface~A, the reference is the exact KL of the distributions defined by the stored raw constructor arguments, with gradients taken with respect to those arguments. Arb certifies the rounding of each reference value to the target format. Six uncertifiable candidates were rejected.
Platform: PyTorch 2.15.0+cpu, Python 3.11.8, python-flint 0.9.0, mpmath 1.4.1, batch size 4,096, single thread.
Each output is classified as correctly rounded, off by a stated number of ULP, or qualitatively wrong (negative KL, NaN, Inf, spurious zero, spurious non-zero at a structural zero, sign flip, vanished gradient). A \emph{spurious zero} is a zero output where the correctly rounded reference is non-zero; a positive KL whose correctly rounded value in the target format is zero is scored as correct when the output is zero.
\emph{Significant error} means $\ge8$ ULP or any qualitative error; gradient vectors are additionally scored by relative $L_2$ error and cosine.
Components that are not differentiable at the sample (e.g.\ Uniform at shared end points, the strict upper triangle of \texttt{scale\_tril}) are not scored.

\paragraph{Methods.} A: official. B32/B64: official code on exactly promoted parameters, rounded once (\textsc{Official+}, abbreviated \textsc{Off.+} in tables, is B32 for FP16 inputs and B64 for FP32 inputs). C: rewritten algorithm in the target precision (\textsc{Rewrite}). D32/D64: rewritten algorithm in FP32/FP64, rounded once; D64 is the strict repair (\textsc{Ours}).

\paragraph{Validation.} All methods are evaluated on the same frozen inputs for correct rounding, significant errors, structural zeros, and regressions against the official baselines.
The diagnostic error estimate $E$ uses first-order rounding-error propagation with an empirical safety factor \citep{higham2002}; series and quadrature truncation bounds are rigorous.
A series may be used on a segment only if it comes from an exact identity, converges on the whole segment, and has a rigorous truncation bound of at most $u/16$ relative to the final result (or, for gradients that can cancel, relative to the approximated term).

\paragraph{Interface B.} 20 configurations: \texttt{log\_target}$\,\in\{$False, True$\}\times\{$FP16, FP32$\}\times K\in\{2,4,16,256\}$ plus four $K=1024$ stress configurations; batch sizes 1 and 8; reductions \texttt{none}, \texttt{sum}, \texttt{mean}, \texttt{batchmean}; gradients with respect to both arguments.
Inputs are produced from logits in regions R0 (components uniform in $[-5,5]$), R1 (range in $[10,30]$) and R2 (range in $[30,10^{3}]$) and from boundary strata (zeros, $-\infty$, extreme magnitudes).
942,200 batches; $5.69\times10^{8}$ certified truths, of which $5.67\times10^{8}$ are scored (inputs outside the domain of the operator and target gradients that do not exist are not); 118 of the 1,920 cells were completed by additional sampling, kept as separate strata.
Three truths are stored per batch: $T_{\mathrm{op}}$, $T_{\mathrm{e2e}}$ and their difference.

\subsection{Error by KL interval}
\label{app:bins}

\begin{figure}[h]
\centering
\includegraphics[width=\textwidth]{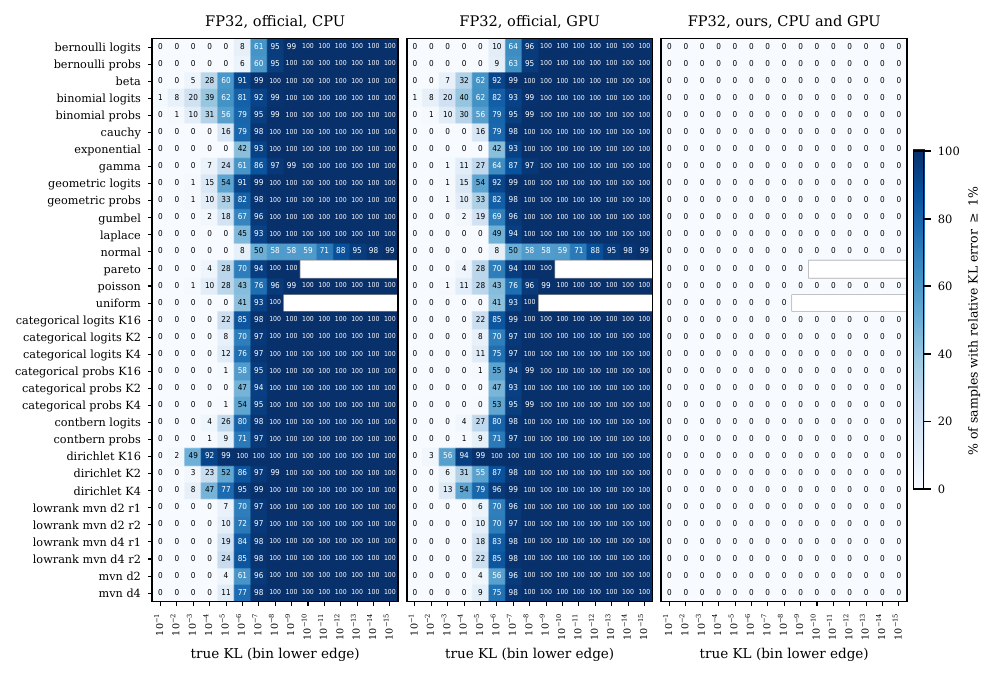}
\caption{Interface A, FP32 inputs, region R0. Share of samples whose KL has relative error $\ge1\%$, per configuration and KL interval. The left panel shows the official implementation on CPU (PyTorch 2.15.0), the middle panel the official implementation on an A100 GPU (PyTorch 2.4.0), and the right panel \textsc{Ours}, whose CPU and GPU results are identical (all zero). Blank cells have fewer than 100 samples (not constructible on the FP32 grid).}
\label{fig:heat32}
\end{figure}

\begin{figure}[h]
\centering
\includegraphics[width=0.7\textwidth]{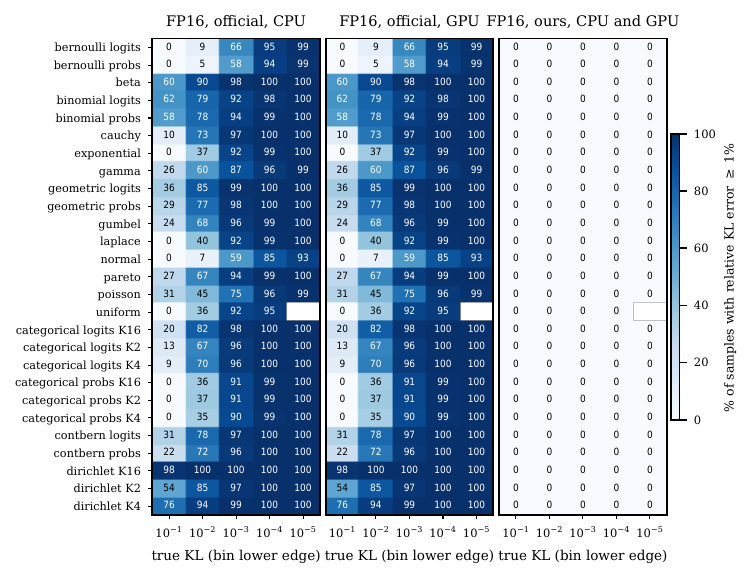}
\caption{Interface A, FP16 inputs, region R0, as \cref{fig:heat32}. The relative error is defined only while the true KL is a normal FP16 number (above about $6\times10^{-5}$). MultivariateNormal and LowRankMultivariateNormal have no FP16 results because the official implementation does not run.}
\label{fig:heat16}
\end{figure}

\Cref{tab:bins} gives the numbers behind \cref{fig:errkl}: the 16 scalar-family configurations pooled, region R0, per decade of the true KL.
\Cref{fig:heat32,fig:heat16} resolve the first metric by configuration for all 33 configurations and sub-configurations (the 23 configurations of \cref{tab:main} with their dimension variants listed separately), for the official implementation on CPU and on GPU and for \textsc{Ours}.
In FP32 every configuration is above 90\% by $\KL\approx10^{-8}$, with Normal as the only partial exception (50--99\%); the families that fail earliest are Dirichlet, Binomial and Beta.
The rows ``in FP64'' and ``in FP32'' of \cref{tab:bins} show that evaluating the official formula in the next precision moves the same pattern nine decades down (FP32 inputs) or leaves a visible residue already at $10^{-4}$ (FP16 inputs); it does not remove it.

\begin{table}[h]
\caption{Interface A, region R0, 16 scalar-family configurations pooled: share of samples (\%) per KL interval. Column heads are the lower edge of the interval (e.g.\ $10^{-9}$ means $[10^{-9},10^{-8})$). ``--'': fewer than 300 samples, or (FP16 relative error) true value subnormal. Gradient metrics refer to the full gradient vector.}
\label{tab:bins}
\centering\scriptsize
\setlength{\tabcolsep}{2.6pt}
\begin{tabular}{lll*{15}{r}}
\toprule
input & evaluated & metric & $10^{-1}$ & $10^{-2}$ & $10^{-3}$ & $10^{-4}$ & $10^{-5}$ & $10^{-6}$ & $10^{-7}$ & $10^{-8}$ & $10^{-9}$ & $10^{-10}$ & $10^{-11}$ & $10^{-12}$ & $10^{-13}$ & $10^{-14}$ & $10^{-15}$\\
\midrule
FP32 & official & KL rel.\ err.\ $\ge1\%$ & 0.1 & 0.5 & 2.4 & 9.2 & 23.7 & 56.0 & 86.4 & 96.2 & 96.9 & 97.0 & 98.0 & 99.1 & 99.6 & 99.8 & 99.9 \\
FP32 & official & KL negative & 0.0 & 0.0 & 0.0 & 0.2 & 0.8 & 3.4 & 8.2 & 14.6 & 23.6 & 28.4 & 29.1 & 29.5 & 29.3 & 28.9 & 29.1 \\
FP32 & official & grad.\ rel.\ err.\ $\ge1\%$ & 0.0 & 0.0 & 0.0 & 0.0 & 0.0 & 0.1 & 0.8 & 3.2 & 8.6 & 20.1 & 41.8 & 72.8 & 89.7 & 94.5 & 95.8 \\
FP32 & official & grad.\ cosine $<0.99$ & 0.0 & 0.0 & 0.0 & 0.0 & 0.0 & 0.0 & 0.0 & 0.0 & 0.3 & 1.6 & 4.4 & 9.4 & 19.6 & 37.0 & 51.2 \\
FP32 & in FP64 & KL rel.\ err.\ $\ge1\%$ & 0.0 & 0.0 & 0.0 & 0.0 & 0.0 & 0.0 & 0.0 & 0.0 & 0.0 & 0.2 & 0.8 & 3.2 & 10.7 & 24.7 & 56.2 \\
FP32 & in FP64 & KL negative & 0.0 & 0.0 & 0.0 & 0.0 & 0.0 & 0.0 & 0.0 & 0.0 & 0.0 & 0.0 & 0.0 & 0.1 & 0.1 & 0.6 & 2.7 \\
FP16 & official & KL rel.\ err.\ $\ge1\%$ & 22.7 & 53.4 & 86.8 & 97.2 & 99.3 & -- & -- & -- & -- & -- & -- & -- & -- & -- & -- \\
FP16 & official & KL negative & 0.5 & 2.7 & 7.1 & 13.3 & 22.3 & 28.6 & 30.3 & 27.8 & 16.9 & 7.4 & 4.6 & 2.0 & 3.7 & -- & -- \\
FP16 & official & grad.\ rel.\ err.\ $\ge1\%$ & 6.7 & 13.7 & 28.8 & 59.0 & 83.6 & 93.3 & 95.5 & 87.0 & 80.3 & 67.7 & 66.5 & 72.8 & 82.5 & -- & -- \\
FP16 & official & grad.\ cosine $<0.99$ & 0.0 & 0.8 & 2.8 & 6.4 & 14.2 & 28.5 & 43.9 & 64.5 & 65.2 & 49.5 & 48.1 & 49.1 & 67.2 & -- & -- \\
FP16 & in FP32 & KL rel.\ err.\ $\ge1\%$ & 0.1 & 0.5 & 2.2 & 8.0 & 13.1 & -- & -- & -- & -- & -- & -- & -- & -- & -- & -- \\
FP16 & in FP32 & KL negative & 0.0 & 0.0 & 0.0 & 0.2 & 0.6 & 2.6 & 6.0 & 7.3 & 10.5 & 7.7 & 7.2 & 5.6 & 5.7 & -- & -- \\
\bottomrule\relax  
\end{tabular}
\end{table}

\section{Interface A: Analysis and Repair of Each Closed Form}
\label{app:ifaceA}

For every distribution we give the official formula (PyTorch 2.15.0), the origin of its error, the repair, the reason the repair works, and the result on the frozen data.
Rates are significant-error rates in the order FP16 / FP32 unless stated otherwise; ``\textsc{Off.+}'' is the official code in the next precision, ``\textsc{Rewrite}'' the repaired formula in the target precision, and ``\textsc{Ours}'' the repaired formula in FP64 rounded once.
In all 18 distributions \textsc{Ours} has zero significant errors in the KL and in every gradient component and zero new errors; we do not repeat this in each subsection.

\subsection{Common building blocks}
\label{app:blocks}

\paragraph{The function $\varphi$.} $\varphi(x)=x-\lgp(x)\ge0$. On $-\tfrac34\le x\le3$ we evaluate \cref{eq:series}; with $z=s^2\le0.36$ and $Q_K(z)=\sum_{k=1}^{K}z^{k-1}/(2k+1)$ (written $Q$ below) evaluated by Horner's rule, the remainder $R_K$ of the series after $K$ terms satisfies $|R_K|/\varphi\le4\,(0.6)^{2K+1}/(2K+3)$, i.e.\ $1.2\times10^{-5}$, $1.9\times10^{-9}$, $3.4\times10^{-18}$ for $K=9,17,36$ (FP16, FP32, FP64), each below $u/16$.
The two terms have ratio at most about 0.12 for $x>0$ and add for $x<0$, so there is no cancellation.
Outside the segment, $\varphi=x-\lgp(x)$ ($x>3$) and $\varphi=x-\log(b/a)$ ($x<-\tfrac34$, with $1+x=b/a$) are well conditioned (the subtracted term is at most 0.86 of the result, resp.\ condition number $\le3.4$).
When $x=d/a$ is formed from $d=b-a$, $d$ is exact for $b\in[a/2,2a]$ \citep{sterbenz1974}, and the series is evaluated with $s=d/(a+b)$ computed directly from the inputs.
We also use the exponential counterpart $\xi(t)=e^{t}-1-t\ge0$.

\paragraph{Error-free transformations.} TwoSum returns $\fl(a+b)$ together with the exact error $a+b-\fl(a+b)$; Dekker's TwoProduct returns $\fl(ab)$ together with $ab-\fl(ab)$ (exact unless the product of the last significant bits underflows, which is checked per sample); Shewchuk's Grow-Expansion sums such pairs exactly.
\emph{Double-length arithmetic} represents a number as an unevaluated pair (hi, lo) of working-precision numbers and implements $+,\times,\div,\sqrt{\ }$ on pairs with these primitives.

\paragraph{Mantissa/exponent products.} Products and quotients whose intermediate results could overflow or become subnormal although the final result is representable are evaluated on \texttt{frexp} mantissas, with the exponents added separately and applied by one exact scaling by $2^k$ followed by a single rounding.

\paragraph{Special functions.} For Gumbel, Gamma, Beta and Dirichlet we do not call \texttt{torch.lgamma}, \texttt{digamma} or \texttt{polygamma}.
We use the Binet function $w(a)=\log\Gamma(a)-(a-\tfrac12)\log a+a-\tfrac12\log2\pi$, which is convex, its derivatives through Stirling-type series with rigorous tail bounds, exact-coefficient zeta series near small arguments, exact shifts by integers, and Gauss--Legendre quadrature (5/7/13 nodes for FP16/FP32/FP64) of same-sign integrands for differences at nearby arguments.
The quadrature error bound (maximum modulus on a Bernstein ellipse) is rigorous; the node counts were verified on a parameter grid and independently measured as $8.3\times10^{-10}$, $3.9\times10^{-14}$, $6.3\times10^{-27}$ maximum relative error.

\paragraph{Notation.} $\partial_\theta$ denotes $\partial\KL/\partial\theta$; $\psi=(\log\Gamma)'$ is the digamma function; $B_f(x,y)=f(x)-f(y)-f'(y)(x-y)\ge0$ is the Bregman divergence of a convex function $f$; $\sigma(\cdot)$ is the logistic sigmoid and $\mathrm{logit}$ its inverse. $c_{\log}u$ denotes a bound on the relative error of the library's \texttt{log} and \texttt{log1p}; $c_{\log}$ is a small constant measured on the platform. $e_{\max}$ and $x_{\max}$ are the largest exponent and the largest finite number of the format, and \texttt{eps} is its machine epsilon ($2^{-10}$, $2^{-23}$, $2^{-52}$), the constant with which PyTorch's constructors clamp probabilities to $[\texttt{eps},1-\texttt{eps}]$.

\paragraph{Range and boundary semantics.} We use power-of-two scaling and mantissa/exponent arithmetic for products and quotients near the format limits. The repairs evaluate distributions from raw parameters, bypassing constructor clamps and \texttt{softmax}.

\paragraph{Midpoint-aware final rounding.} Computing in FP64 and rounding once to FP16 or FP32 fails when the leading term of the result is exactly a rounding midpoint of the target format and the correction is below FP64 resolution relative to it. This occurs for logit-space gradients near zero (leading term $(w-z)/4$ for Bernoulli and $(w-z)/2$ for Geometric, where $z$ and $w$ are the logits of $P$ and $Q$).
There we write the result as an exactly representable leading term plus a small series, recover the rounded-away remainder with Fast2Sum, and move one ULP towards the truth when the FP64 result sits on a midpoint.

\subsection{Cauchy}
\textbf{Official.} With locations $\mu_p,\mu_q$, scales $\gamma_p,\gamma_q$ and $N=(\gamma_p+\gamma_q)^2+(\mu_p-\mu_q)^2$: $\KL=\log N-\log(4\gamma_p\gamma_q)$. We write $\Sigma\gamma=\gamma_p+\gamma_q$, $\Delta\gamma=\gamma_p-\gamma_q$ and $\Delta\mu=\mu_p-\mu_q$.

\textbf{Error.} For $P\approx Q$ both logarithms are $\approx\log(4\gamma^2)$ with $\gamma\approx\gamma_p\approx\gamma_q$; each carries an absolute error of order $(c+|\log N|)u$ that survives the subtraction, so the relative error is $\approx(c_1+2|\log4\gamma^2|)u/\KL$, with small constants $c,c_1$, and the sign is random once KL is below that error. Autograd gives $\partial\KL/\partial\gamma_p=2\Sigma\gamma/N-1/\gamma_p$, a difference of two $O(1/\gamma)$ terms. In FP16, for small $\gamma$ both arguments of the logarithms become subnormal and round to the same number, the derivatives $1/N$ and $1/(4\gamma_p\gamma_q)$ overflow, and the gradient is $\infty-\infty=$ NaN, also at $P=Q$. The $\mu$ gradient has no cancellation.

\textbf{Repair.} From $(\gamma_p+\gamma_q)^2-4\gamma_p\gamma_q=(\gamma_p-\gamma_q)^2$,
\[\KL=\lgp(r),\qquad r=\frac{(\gamma_p-\gamma_q)^2+(\mu_p-\mu_q)^2}{4\gamma_p\gamma_q}\ge0.\]
$r$ is homogeneous of degree zero, so all four quantities are scaled by a power of two $2^k$ with $\max(\gamma_p,\gamma_q,|\Delta\mu|)\,2^k\in[8,16)$ (applied as two representable factors); if $r$ overflows, a logarithmic fallback is used, which cannot cancel because then $\KL>\log x_{\max}$.
Gradients are analytic: $\partial_{\mu_p}=2\Delta\mu/N=-\partial_{\mu_q}$, $\partial_{\gamma_p}=(\Delta\gamma\,\Sigma\gamma-\Delta\mu^2)/(\gamma_pN)$, $\partial_{\gamma_q}=(-\Delta\gamma\,\Sigma\gamma-\Delta\mu^2)/(\gamma_qN)$, evaluated on the scaled quantities.

\textbf{Why it works.} The squared differences enter $r$ as small numbers instead of being recovered from large ones, and $\lgp$ is well conditioned for $r\ge0$: the first-order relative error is about $(6+c_{\log})u$ independent of KL. The condition number of the $\gamma$ gradient drops from $\kappa_A=(2\gamma_p\Sigma\gamma+N)/|\Delta\gamma\Sigma\gamma-\Delta\mu^2|$ to $\kappa_C=(|\Delta\gamma|\Sigma\gamma+\Delta\mu^2)/|\Delta\gamma\Sigma\gamma-\Delta\mu^2|\le\kappa_A$; the remaining cancellation at $\gamma_p^2\approx\gamma_q^2+\Delta\mu^2$ is intrinsic and is the reason \textsc{Rewrite} keeps a few errors and \textsc{Ours} does not. After scaling all denominators are positive and bounded away from zero, so no NaN can occur, and $P=Q$ gives $r=0$, $\KL=+0$ and zero gradients.

\textbf{Frozen data} (345,028 samples). KL: \A{} 64.05 / 88.88\%; \textsc{Off.+} 10.60 / 34.84\%; \textsc{Rewrite} 0.003 / 0\%. $P$-scale gradient: \A{} 53.27 / 82.93\%; \textsc{Off.+} 0.34 / 6.37\%; \textsc{Rewrite} 0.18 / 0.09\%. \textsc{Rewrite} adds new errors on 41 samples (0.012\%), all within the error estimate.

\subsection{Uniform}
\textbf{Official.} With supports $[p_l,p_h]$ and $[q_l,q_h]$: $\KL=\log\dfrac{q_h-q_l}{p_h-p_l}$ for nested supports, $+\infty$ otherwise. We write $w_p=p_h-p_l$ and $w_q=q_h-q_l$ for the widths.

\textbf{Error.} The two widths are rounded separately and their ratio is $\approx1+\KL$, so the relative error is about $2u/\KL$ (12\% at $\KL=10^{-6}$ in FP32), and the ratio rounds to 1 for $\KL<u/2$ (spurious zero). The gradients $\pm1/w_p$ and $\pm1/w_q$ involve no cancellation and are correct in the official code.

\textbf{Repair.} On nested supports $w_q-w_p=(q_h-p_h)+(p_l-q_l)$ with both terms non-negative, so
\[\KL=\lgp(r),\qquad r=\frac{d_1+d_2}{w_p},\quad d_1=q_h-p_h\ge0,\quad d_2=p_l-q_l\ge0 .\]
All inputs are halved if one of them is $\ge2^{e_{\max}-1}$; if the ratio overflows, $\KL=\log w_q-\log w_p$. Gradients are $\pm1/w_p$, $\mp1/w_q$.

\textbf{Why it works.} $d_1,d_2$ are direct differences of the inputs (exact for nearby end points), the ratio has relative error $\le4u$ and $\lgp$ has condition number $\le1$ on $r\ge0$, hence the relative error of the KL is at most $(4+c_{\log})u$. $P=Q$ gives $+0$.

\textbf{Frozen data} (181,798). KL: \A{} 51.62 / 75.08\%; \textsc{Off.+} 0 / 0.012\%; \textsc{Rewrite} 0 / 0\% (maximum error 2 ULP). Gradients have no significant error in any method.

\subsection{Exponential}
\textbf{Official.} With $r=\lambda_q/\lambda_p$: $\KL=-\log r+r-1$.

\textbf{Error.} $r-\log r\approx1+\KL$ is rounded near 1 with absolute error $\approx u$ while $\KL\approx x^2/2$, $x=r-1$: relative error $\approx2u/x^2$, spurious zero for $\KL<u/2$. Autograd's $1-1/r$ cancels with relative error $\approx u/|x|$. FP64 evaluation leaves 33.5\% significant errors for FP32 inputs because $x$ can be as small as $2^{-24}$.

\textbf{Repair.} $d=\lambda_q-\lambda_p$ (direct), $x=d/\lambda_p$, $\KL=\varphi(x)$ as in \cref{app:blocks}; analytic gradients $\partial_{\lambda_p}=-x/\lambda_p$, $\partial_{\lambda_q}=x/\lambda_q$.

\textbf{Why it works.} See \cref{app:blocks}; KL and gradients are computed from the same rounded value $\tilde x$ of $x$, whose rounding changes KL by a relative $O(u)$. $P=Q$ gives $x=0$ and three exact zeros; a final $\max(\cdot,0)$ guarantees $\KL\ge0$.

\textbf{Frozen data} (308,000). KL: \A{} 72.57 / 85.64\%; \textsc{Off.+} 0 / 33.49\%; \textsc{Rewrite} 0 / 0\% (max.\ 6 ULP). Rate gradient: \A{} 48.78 / 75.11\%; all other methods 0. A near-exhaustive FP16 sweep of $3.6\times10^{7}$ pairs gave at most 7 ULP for \textsc{Rewrite} and 1 ULP for the FP32 evaluation.

\subsection{Poisson}
\textbf{Official.} With rates $a=\lambda_p$, $b=\lambda_q$: $\KL=a(\log a-\log b)-(a-b)$. Let $x=(b-a)/a$.

\textbf{Error.} Two cancellations: $\log a-\log b\approx-x$ with each logarithm carrying an error $\approx u|\log a|$, then $a(\cdot)-(a-b)$ leaves $a\,x^2/2$. The relative error is $\approx4c_{\log}u|\log a|/x^2$; because $a$ ranges up to $10^4$, even the control bin $\KL\in[0.1,1)$ has about 39\% outputs off by $\ge8$ ULP, and 25--28\% of all outputs are negative. Gradients vanish when both logarithms round to the same value.

\textbf{Repair.} $\KL=a\,\varphi(x)$, written so that the small $\varphi$ is never formed and then multiplied by a large $a$: with $d=b-a$, $s=d/(a+b)$, $z=s^2$,
\[\KL=d\,s-2\,(a\,s)\,z\,Q(z)\qquad(-\tfrac34\le x\le3),\]
and $a\max(x-L,0)$ or $\max(d-aL,0)$ with $L=\log(b/a)$ outside. Gradients: $\partial_{\lambda_p}=-L$, $\partial_{\lambda_q}=d/b$.

\textbf{Why it works.} The factor $a$ cancels exactly between the truncation error and the KL, so the bound of \cref{app:blocks} is inherited; $|d\,s|$ and $|a\,s|$ never exceed $|d|$, so nothing overflows before the result does.

\textbf{Frozen data} (323,498). KL: \A{} 75.64 / 85.82\%; \textsc{Off.+} 9.75 / 34.44\%; \textsc{Rewrite} 0.002 / 0\%. Gradients: \A{} 54.8--78.3\%; others 0. Negative KL disappears.

\subsection{Normal}
\textbf{Official.} With means $\mu_p,\mu_q$, standard deviations $\sigma_p,\sigma_q$, $\delta=\mu_p-\mu_q$, $r=\sigma_p/\sigma_q$ and $w=\delta/\sigma_q$: $\KL=\tfrac12\big(r^2+w^2-1-\log r^2\big)$.

\textbf{Error.} (i) $r^2-1-\log r^2$ cancels like the Exponential kernel. (ii) $w^2$ is added to $r^2\approx1$ first and is lost when $w^2<u$ (10.6\% spurious zeros in FP16); in FP16 $w^2$ can overflow before the KL does. (iii) Autograd's scale gradients pass through $\tfrac12-\tfrac1{2r^2}$. (iv) The exact $\partial_{\sigma_q}=(\sigma_q^2-\sigma_p^2-\delta^2)/\sigma_q^3$ has a genuine zero on $\sigma_q^2=\sigma_p^2+\delta^2$, where the official code returns non-zero. FP64 evaluation leaves 13.84\% for FP32 inputs.

\textbf{Repair.} \Cref{eq:normal}: $\KL=\varphi(e)+\tfrac12e^2+\tfrac12w^2$, with $\varphi(e)$ from \cref{app:blocks} ($s=(\sigma_p-\sigma_q)/(\sigma_p+\sigma_q)$) and $w^2/2=(w\cdot\tfrac12)\,w$.
Gradients: $\partial_{\mu_p}=w/\sigma_q=-\partial_{\mu_q}$, $\partial_{\sigma_p}=e(e+2)/\sigma_p$, $\partial_{\sigma_q}=N/\sigma_q^3$ with $N=\sigma_q^2-\sigma_p^2-\delta^2$.
Where cancellation in $N$ is possible ($\sigma_q\ge\max(\sigma_p,|\delta|)/2$), $N$ is computed \emph{exactly} in the working precision: $\delta$ by TwoSum, power-of-two scaling, five Dekker products, and a ten-term Grow-Expansion.

\textbf{Why it works.} Three non-negative terms; the truncation bound is inherited because $\KL\ge\varphi(e)$. With $N$ exact, $\partial_{\sigma_q}$ suffers only 4--5 roundings. If $\sigma_q^2=\sigma_p^2+\delta^2$ holds exactly, a 2-adic argument shows that all scaled operands lie on a common binary grid, every product is exact, and the computed $\hat N$ is $+0$; all 109,558 FP16 and 908,771,452 FP32 Pythagorean triples we enumerated return $+0$.
Computing $(\sigma_q-\sigma_p)(\sigma_q+\sigma_p)-\delta^2$ in working precision is not enough: it rounds, e.g.\ in FP16 for $(\mu_p,\sigma_p,\mu_q,\sigma_q)=(329,1080,0,1129)$.

\textbf{Frozen data} (339,499). KL: \A{} 42.40 / 82.18\%; \textsc{Off.+} 0 / 13.84\%; \textsc{Rewrite} 0 / 0\%. Scale gradients: \A{} 32.9--61.0\%; \textsc{Rewrite} has one new case in FP16.

\subsection{Laplace}
\textbf{Official.} With locations $\mu_p,\mu_q$, scales $a=\sigma_p$, $b=\sigma_q$, $r=a/b$ and $d=|\mu_p-\mu_q|$: $\KL=-\log r+d/b+r\,e^{-d/a}-1$.

\textbf{Error.} Four $O(1)$ terms sum to a small KL (43\% spurious zeros and 1.4\% negative values in FP16). Autograd's location gradient $(1-e^{-t})/b$, $t=d/a$, vanishes when $e^{-t}$ rounds to 1 (24.5\% in FP16); scale gradients are differences of $O(1/b)$ terms and flip sign; in FP16, $d/(a\cdot a)$ becomes $0/0$ for $a<2^{-12}$, $d=0$. FP64 evaluation leaves 31.95\% (KL) and 5.7\% (scale gradient) for FP32 inputs.

\textbf{Repair.} With $x=(a-b)/b$ and $t=d/a$,
\[\KL=\varphi(x)+r\,\xi(-t),\qquad\xi(-t)=e^{-t}-1+t\ge0 .\]
$\xi(-t)$ and $\chi(t)=1-(1+t)e^{-t}$ use alternating series for $t\le1$ (7/11/18 and 8/12/19 terms) and $\emo$ beyond. Gradients: $\partial_{\mu_p}=-\mathrm{sgn}(\mu_p-\mu_q)\,\emo(-t)/b$, $\partial_a=(a-b)/(ab)-\chi(t)/b$, $\partial_b=-(x+r\,\xi(-t))/b$. Products of the form $t^2X/b$ are evaluated as $(t\cdot X)(t/b)$ to avoid underflow.

\textbf{Why it works.} Two non-negative terms. The series alternate with decreasing terms, so the tail is bounded by the first neglected term: the relative truncation after $K$ terms is $\le3/(K+2)!$ for $\xi(-t)$ and $\le6(K+1)/(K+2)!$ for $\chi$, below $u/16$. There is no structural zero of the scale gradients for $d\ne0$: it would require $e^{-t}$ rational for rational $t\ne0$, impossible by the Lindemann--Weierstrass theorem. The genuine zero crossings of $\partial_a,\partial_b$ are resolved by the higher precision.

\textbf{Frozen data} (339,692). KL: \A{} 58.32 / 85.87\%; \textsc{Off.+} 2.24 / 31.95\%; \textsc{Rewrite} 1.46 / 0\% (FP16 residue only). Gradients: \A{} 45.4--79.7\%; \textsc{Rewrite} $\le0.15\%$, with 86 new cases at zero crossings ($<0.03\%$ of the repaired ones).

\subsection{Pareto}
\textbf{Official.} For $s_p\ge s_q$ (scales) and shapes $a_p,a_q$: $\KL=a_q\log\dfrac{s_p}{s_q}-\log\dfrac{a_q}{a_p}+\dfrac{a_q}{a_p}-1$.

\textbf{Error.} Four $O(1)$ terms for a small KL (33.6\% spurious zeros in FP16); autograd's shape gradients pass through $1-1/(a_q/a_p)$. Scale gradients are accurate. FP64 evaluation leaves 6.99\% for FP32 inputs.

\textbf{Repair.} With $x=(s_p-s_q)/s_q\ge0$ and $y=(a_q-a_p)/a_p$,
\[\KL=a_q\,\lgp(x)+\varphi(y).\]
Gradients: $\partial_{s_p}=a_q/s_p$, $\partial_{s_q}=-a_q/s_q$, $\partial_{a_p}=((a_p-a_q)/a_p)/a_p$, $\partial_{a_q}=\lgp(x)+(a_q-a_p)/(a_pa_q)$; when the second, negative, term of $\partial_{a_q}$ overflows alone while the sum is finite, the merged numerator $(a_q\lgp(x)+(a_q-a_p)/a_p)/a_q$ is used.

\textbf{Why it works.} Two non-negative terms; the series occurs only in $\varphi\le\KL$. For $s_p\ne s_q$, $\lgp(x)$ is the logarithm of a rational number $\ne1$, hence transcendental, so $\partial_{a_q}$ has no structural zero.

\textbf{Frozen data} (206,950). KL: \A{} 73.52 / 83.76\%; \textsc{Off.+} 0 / 6.99\%; \textsc{Rewrite} 0 / 0\%. Shape gradients: \A{} 47.9--81.4\%; \textsc{Rewrite} 0.40 / 0.50\% on $\partial_{a_q}$ (14 new cases at its zero crossing).

\subsection{Bernoulli (probs and logits)}
\textbf{Official.} With logits $l_p,l_q$ and probabilities $p=\sigma(l_p)$, $q=\sigma(l_q)$: $\KL=p\,(\mathrm{softplus}(-l_q)-\mathrm{softplus}(-l_p))+(1-p)\,(\mathrm{softplus}(l_q)-\mathrm{softplus}(l_p))$. The \texttt{probs} constructor first forms $l=\log c-\lgp(-c)$ from $c=\mathrm{clamp}(p,\texttt{eps},1-\texttt{eps})$.

\textbf{Error.} Each term is a probability times a difference of two nearly equal softplus values, and the two terms have opposite signs, while $\KL\approx(p-q)^2/(2p(1-p))$: 17.5--20.8\% of the outputs are negative. Gradients inherit the same chain plus the sigmoid/logit conversion and the clamp. For large $|l|$ the probability is subnormal. FP64 evaluation leaves 27.15\% (probs) and 25.48\% (logits) for FP32 inputs.

\textbf{Repair (both parameterisations).} With $x_1=q/p-1$ and $x_2=(1-q)/(1-p)-1$,
\[\KL=p\,\varphi(x_1)+(1-p)\,\varphi(x_2),\]
with series arguments $s_1=(q-p)/(p+q)$, $s_2=(p-q)/((1-p)+(1-q))$.
\emph{Probs}: no clamp; $\partial_p=\lgp(x_2)-\lgp(x_1)$, $\partial_q=(q-p)/(q(1-q))$; closed forms at $p,q\in\{0,1\}$.
\emph{Logits} ($z=l_p$, $w=l_q$): $\partial_z=p(1-p)(z-w)$, $\partial_w=q-p$; probabilities, log-ratios and $p-q$ are written in logit space ($\sigma$ through $e^{-|z|}$; a softplus difference written by hand, because \texttt{torch.nn.functional.softplus} returns its argument $y$ for $y>20$ and thereby drops $\lgp(e^{-y})$ in FP64; $p-q=p(1-q)(1-e^{w-z})$ for $|z-w|\le2$). Where a probability is subnormal ($|\mathrm{logit}|>9.70$, $87.34$, $708.40$ for FP16, FP32, FP64), every ``probability $\times$ factor'' is evaluated as a product of two or four normal factors $e^{z/2}$, $e^{z/4}$ with one rounding. \textsc{Ours} uses the midpoint-aware final rounding of \cref{app:blocks} for $|z|,|w|\le1/16$.

\textbf{Why it works.} Two non-negative terms; KL cannot be negative; the interior identity $p=q$ gives three exact $+0$ (including $-0$ versus $+0$ logits). The midpoint rule is necessary because for $|z|,|w|\approx10^{-4}$ the true gradient lies within $10^{-7}$ ULP of an FP16 subnormal midpoint (its leading term $(w-z)/4$ \emph{is} a midpoint); without it, the FP32-evaluated variant misrounds 7 samples of the benchmark.

\textbf{Frozen data} (327,099 probs; 353,392 logits). KL, probs: \A{} 61.43 / 81.06\%; \textsc{Off.+} 0.29 / 27.15\%; \textsc{Rewrite} 0.41 / 0\%. KL, logits: \A{} 49.92 / 75.95\%; \textsc{Off.+} 1.29 / 25.48\%; \textsc{Rewrite} 0.012 / 0.012\%. $P$-gradient: \A{} 60.3--78.0\%; \textsc{Off.+} 1.5--9.5\%; \textsc{Rewrite} $\le0.002\%$. Exhaustive check of all $2.36\times10^{8}$ FP16 pairs $(p,q)$: \textsc{Rewrite} at most 6 ULP (KL), FP32 evaluation at most 1 ULP.

\subsection{Binomial (probs and logits)}
\textbf{Official.} $\KL=n\big(p\,(l_p-l_q)+\lgp(-p)-\lgp(-q)\big)$ (equal total counts $n\in\{1,10,100,1000\}$ in our data).

\textbf{Error.} Three terms of mixed sign: 13.4--23.2\% negative and 16.6--30.2\% spurious zeros. When $q$ rounds to 1, $\lgp(-q)=-\infty$: the logits configuration still returns $\infty$ for 2.2\% (FP32 evaluation) and 5.3\% (FP64 evaluation) of FP16 inputs. FP64 evaluation leaves 37.5\% (probs) and 43.8\% (logits) for FP32 inputs.

\textbf{Repair.} $\KL=n\cdot\KL_{\mathrm{Bernoulli}}$ with the Bernoulli algorithm, but $n$ must enter \emph{before} any subnormal intermediate is formed and never in a product with an intermediate that can overflow: e.g.\ inside the series segment $n\varphi=((n\,x)\,s)-((((2n)\,s)\,s)\,s)\,Q(z)$, then multiplied by $p$ or $1-p$; in logit space $n$ multiplies only normal factors; the midpoint rule includes $n$ in its exact leading term.

\textbf{Why it works.} Each expression is $n$ times an already verified Bernoulli quantity with the multiplications reordered; after $n$ has entered, every further step multiplies by a factor $\le1$ or is an exact addition. Multiplying by $n$ last would make the same-precision FP16 KL wrong by up to 316 ULP and misround 7 gradients in the FP32-evaluated variant.

\textbf{Frozen data} (329,885 probs; 345,715 logits). KL, probs: \A{} 83.86 / 85.60\%; \textsc{Off.+} 10.43 / 37.51\%; \textsc{Rewrite} 0.32 / 0\%. KL, logits: \A{} 69.69 / 86.95\%; \textsc{Off.+} 34.93 / 43.79\% (6.12\% for FP16 inputs even in FP64); \textsc{Rewrite} 0.020 / 0.031\%. $P$-gradient: \A{} 73.7--87.8\%.

\subsection{Geometric (probs and logits)}
\textbf{Official.} For the failure-count Geometric $P(k)=p(1-p)^k$: $\KL=-H(p)-\lgp(-q)/p-\mathrm{logit}(q)$, where the entropy $H(p)$ of $P$ is computed as \texttt{binary\_cross\_entropy\_with\_logits}$(\text{logits},\text{probs})/p$.

\textbf{Error.} Three terms, each a logarithm or a logarithm divided by $p$, sum to a much smaller KL: the worst family in our study (91--97\% significant KL errors, 19--27\% negative). With logits, $q$ rounding to 1 gives $-\infty$: 2.8--12.2\% infinite and 1.6--2.5\% NaN KL values, and 4.4--15.8\% NaN gradients. FP64 evaluation leaves 48.4\% (probs) and 51.5\% (logits) for FP32 inputs.

\textbf{Repair.} The Geometric KL equals the Bernoulli KL divided by $p$. With $x_1=q/p-1$, $x_2=(1-q)/(1-p)-1$, $\bar\varphi(x)=\varphi(x)/x$ and $g(x)=\lgp(x)/x$ (with $\bar\varphi(0)=0$ and $g(0)=1$), all outputs are single products:
\[\KL=x_1\big(\bar\varphi(x_1)-\bar\varphi(x_2)\big),\]
where $\bar\varphi(x_1)$ and $\bar\varphi(x_2)$ have opposite signs, so their difference does not cancel; inside the series segment $\bar\varphi=s-a(1-s)$ and $g=(1-s)(1+a)$ with $a=s^2Q(s^2)$.
\emph{Probs}: $\partial_p=((-x_1)\,g(x_2))/p/(1-p)$, $\partial_q=(x_1/q)/(1-q)$.
\emph{Logits} ($z$ and $w$ the logits of $P$ and $Q$): $\partial_w=x_1$, $\partial_z=-x_1g(x_2)$, with $x_1=\sigma(-w)\,\emo(w-z)$ evaluated in three branches by the signs of $w-z$ and $w$ (the difference $z-w$ exact by TwoSum where Sterbenz's lemma does not apply) and small or large factors ($\sigma(-w)$, $e^{-z}$) multiplied last through split exponents. Midpoint-aware rounding for $|z|,|w|\le1/16$ (leading term $(w-z)/2$).

\textbf{Why it works.} The identities used are $r\,x_2=-x_1$ with $r=(1-p)/p$, and $\lgp x=2\operatorname{atanh}s$ only; the KL is a single product, so there is no cancellation, no subnormal intermediate and no premature overflow. As a check, the truth was also evaluated from the official formula at 1,500 digits; it agrees after rounding.

\textbf{Frozen data} (304,861 probs; 352,178 logits). KL, probs: \A{} 91.17 / 94.19\%; \textsc{Off.+} 36.53 / 48.35\%; \textsc{Rewrite} 0 / 0\%. KL, logits: \A{} 94.35 / 96.62\%; \textsc{Off.+} 42.77 / 51.47\% (6.60\% for FP16 inputs in FP64); \textsc{Rewrite} 0.17\% / one sample. $P$-gradient: \A{} 78.5--92.5\%; \textsc{Off.+} 13.8--19.5\%.

\subsection{Gumbel}
\textbf{Official.} With $P=(m,b)$, $Q=(n,c)$ (location, scale) and Euler's constant $\gamma$:
$\KL=-\log\frac bc-\frac nc+\frac mc+\gamma\frac bc+\exp\!\big(\frac nc+\log\Gamma(1+\frac bc)-\frac mc\big)-(1+\gamma)$.

\textbf{Error.} $n/c$ and $m/c$ are rounded separately, so for large $|\text{loc}|/\text{scale}$ their absolute error exceeds the KL; four $O(1)$ terms then sum to a small value (10--14\% negative, 17--36\% spurious zeros). Gradients inherit the cancellation; 7.28\% of the FP16 $Q$-scale gradients are NaN. FP64 evaluation leaves 38.4\% for FP32 inputs.

\textbf{Repair.} With $t=(m-n)/c$, $r=b/c$, $y=\log\Gamma(1+r)-t$ and $x=\emo(y)$,
\[\KL=\varphi(x)+H(r),\qquad H(r)=\log\Gamma(r)+\gamma\,(r-1)\ge0,\]
where $H$ is convex with $H(1)=H'(1)=0$. Gradients: $\partial_m=-x/c$, $\partial_n=x/c$; $c\,\partial_b$ has two exactly equivalent forms, $x\,\psi(1+r)+(\psi(r)+\gamma)$ and $e^{y}\psi(1+r)+(\gamma-1/r)$, of which the one with the smaller terms is chosen; $c\,\partial_c=-(t\,c\,\partial_m+r\,c\,\partial_b)$ by scale invariance. $\log\Gamma$ and $\psi$ are implemented from exact-coefficient zeta series in four ranges of $r$ (\cref{app:blocks}). Eight FP16 range guards handle overflow of $m-n$, of $x$, and of individual products.

\textbf{Why it works.} Two non-negative terms; the identities used are $\log\Gamma(1+r)-\log r=\log\Gamma(r)$, $\psi(1+r)-1/r=\psi(r)$ and $\lgp(x)=y$. Residual cancellation in the gradients occurs only at genuine zeros ($y\approx0$, $\partial_c$ crossing zero).

\textbf{Frozen data} (339,249). KL: \A{} 66.18 / 90.02\%; \textsc{Off.+} 13.78 / 38.42\%; \textsc{Rewrite} 0.007 / 0\%. Gradients: \A{} 73.7--92.5\%; \textsc{Off.+} 2.4--8.7\%; \textsc{Rewrite} 0.31--0.75\% (117 new cases near gradient zeros, $<0.02\%$ of the repaired ones).

\subsection{Gamma}
\textbf{Official.} With $P=(a_p,b_p)$, $Q=(a_q,b_q)$ (concentration, rate):
$\KL=a_q\log\frac{b_p}{b_q}+\log\Gamma(a_q)-\log\Gamma(a_p)+(a_p-a_q)\,\psi(a_p)+(b_q-b_p)\frac{a_p}{b_p}$.

\textbf{Error.} Four terms of size $O(a)$ or $O(a\log a)$ for a KL between $10^{-15}$ and 1; the cancellation grows with $a$ (up to $10^4$ in R1): about 27\% negative. Autograd differentiates through \texttt{digamma}; in FP16 \texttt{lgamma} overflows for $a$ above about 8200 (about 4\% NaN and 4\% infinite concentration gradients). FP64 evaluation leaves 36.9\% for FP32 inputs.

\textbf{Repair.} With $\alpha=a_q/a_p-1$ and $z=\dfrac{a_pb_q}{a_qb_p}-1$ (ratio of the two means minus one),
\[\KL=a_q\,\varphi(z)+\tfrac12\varphi(\alpha)+B_w(a_q,a_p),\]
where $B_w$ is the Bregman divergence of the convex Binet function $w$; all three terms are non-negative.
Gradients: $\partial_{a_p}=(a_q/a_p)z-\alpha(\tfrac12+g(a_p))/a_p$ with $g(s)=s^2w''(s)$; $\partial_{b_p}=-a_qz/b_p$; $\partial_{a_q}=\Delta L-\lgp(z)$ with $L=\psi-\log$ and $\Delta L=L(a_q)-L(a_p)$; $\partial_{b_q}=a_qz/b_q$.
The product difference in $z$ is formed exactly using scaled TwoProduct and TwoSum before division, so equal means give $z=0$ exactly. For near pairs $B_w$ and $\Delta L$ are same-sign integrals evaluated by Gauss--Legendre quadrature; for far pairs $w$ is evaluated by shifts and series. Eight FP16 range guards use the mantissa/exponent form.

\textbf{Why it works.} Three non-negative terms, an exact product difference for $z$, and same-sign quadrature sums. Equivalence with the official formula was checked at 60 and 70 digits; the truth is computed from the official formula at 120--160 digits and does not depend on the rewriting.

\textbf{Frozen data} (331,067). KL: \A{} 78.47 / 88.67\%; \textsc{Off.+} 9.09 / 36.92\%; \textsc{Rewrite} 0.003 / 0.002\%. Gradients: \A{} 64.4--82.5\%; \textsc{Rewrite} $\le0.21\%$ (21 new cases near gradient zeros).

\subsection{Beta}
\textbf{Official.} With concentrations $P=(a,b)$, $Q=(c,d)$ and the Beta function $B$: $\KL=\log B(c,d)-\log B(a,b)+(a-c)\psi(a)+(b-d)\psi(b)+(c-a+d-b)\psi(a+b)$, i.e.\ six \texttt{lgamma} and three \texttt{digamma} terms.

\textbf{Error.} Nine terms of size $O(a\log a)$: about one third of the non-identity outputs are negative; in FP16 \texttt{lgamma}$(a+b)$ overflows (2.4\% NaN). The gradients are differences of first-order terms and, for small parameters, of terms $\approx-1/a$ and $\approx-1/b$. FP64 evaluation leaves 43.8\% for FP32 inputs.

\textbf{Repair.} Let $s=a+b$, $s_q=c+d$, $\alpha=c/a-1$, $\beta=d/b-1$, $z=ad/(bc)-1$, $v(x)=\log\Gamma(x)-x\log x+x$ and $\omega(x)=x\,v''(x)$ (decreasing). Then
\[\KL=c\,\varphi(r_1)+d\,\varphi(r_0)+B_V,\qquad r_1=\tfrac bs\,z,\quad r_0=-\tfrac as\,\tfrac z{1+z},\]
where $B_V=B_v(c,a)+B_v(d,b)-B_v(s_q,s)$ is the Bregman divergence of the convex function $V(a,b)=v(a)+v(b)-v(a+b)$.
Adding three $B_v$ directly cancels when one parameter is much smaller than the other (by a factor of about 100 in our samples), so for near pairs
\[B_V=\tfrac12\mathcal Q+T_3(c,a)+T_3(d,b)-T_3(s_q,s),\qquad \mathcal Q=a\big(\omega(a)-\omega(s)\big)\alpha^2+b\big(\omega(b)-\omega(s)\big)\beta^2+\omega(s)\tfrac{ab}s(\beta-\alpha)^2 ,\]
with three non-negative terms in $\mathcal Q$ and the third-order remainders $T_3(x,y)=B_v(x,y)-\tfrac12(x-y)^2v''(y)=\int_y^x\tfrac12(x-t)^2v'''(t)\,dt$, evaluated through $t^3v'''(t)=t^3w'''(t)-1$ with the Binet function $w$; for far pairs the three $B_v$ are used directly.
Gradients use identities that are exact for arbitrary shifts, e.g.\ $\partial_a=-\alpha\,(\omega(a)-\omega(s))+\tfrac bs(\beta-\alpha)(1+\omega(s))$.
The shared numerator $ad-bc$ is formed exactly from scaled mantissas before evaluating $z$, $r_1$, $r_0$ and $\beta-\alpha$.

\textbf{Why it works.} All leading terms are non-negative, with their ratio numerators computed exactly; residual cancellation is confined to the third-order remainders, to the $Q$-side gradients near their zeros and to far pairs, all inside the error estimate.

\textbf{Frozen data} (329,156). KL: \A{} 85.38 / 92.62\%; \textsc{Off.+} 40.32 / 43.77\%; \textsc{Rewrite} 2.11 / 0.65\%. Gradients: \A{} 73.9--87.6\%; \textsc{Off.+} 0.7--2.9\%; \textsc{Rewrite} 0.15--1.57\%. \textsc{Rewrite} introduces 961 new cases, mostly $Q$-side gradients at small parameters (at most 0.29\% of the repaired ones). This is the family in which the same-precision variant is weakest.

\subsection{Dirichlet ($K=2,4,16$)}
\textbf{Official.} With concentrations $a_1,\dots,a_K$ of $P$ and $c_1,\dots,c_K$ of $Q$: $\KL=\log\Gamma(s)-\sum_j\log\Gamma(a_j)-\log\Gamma(s_q)+\sum_j\log\Gamma(c_j)+\sum_j(a_j-c_j)\big(\psi(a_j)-\psi(s)\big)$, $s=\sum a_j$, $s_q=\sum c_j$.

\textbf{Error.} Large \texttt{lgamma}/\texttt{digamma} terms cancel almost completely: the KL is not correctly rounded in 98.76\% (FP16) and 99.92\% (FP32) of the samples, negative in 37.6\% / 32.8\%, NaN in 3.49\% of FP16 samples, and the gradient direction is reversed in 6.29\% / 1.45\%. In FP64, 56.3\% of FP32-input KL values are still not correctly rounded and 2.1\% are negative.

\textbf{Repair.} The Beta construction generalises. With $\alpha_j=c_j/a_j-1$, $\gamma=(s_q-s)/s$, $\delta_j=\alpha_j-\gamma$, $r_j=a_js_q/(c_js)-1$,
\[\KL=\sum_jc_j\,\varphi(r_j)+\tfrac12\mathcal Q+\sum_jT_3(c_j,a_j)-T_3(s_q,s),\qquad \mathcal Q=\sum_ja_j\big(\omega(a_j)-\omega(s)\big)\alpha_j^2+\omega(s)\sum_ja_j\delta_j^2,\]
with every term of $\mathcal Q$ non-negative and $T_3$ as for Beta; with a far pair, $\KL=\sum_jc_j\varphi(r_j)+\sum_jB_v(c_j,a_j)-B_v(s_q,s)$.
$\delta_j$ and $r_j$ are formed from double-length ratios, never as $\alpha_j-\gamma$. When $\sum a_j$ overflows in FP16, $s$ is represented as $16\bar s$ with $\bar s$ an exact expansion of $\sum_ja_j/16$.

\textbf{Why it works.} As for Beta, to which the formula reduces for $K=2$.

\textbf{Frozen data} (314,451). KL: \A{} 84.3--89.0 / 92.1--93.8\%; \textsc{Off.+} 29.6--55.9 / 41.6--49.8\%; \textsc{Rewrite} 0.009--2.45 / 0.003--0.69\%. Gradients: \A{} 79--95\%. \textsc{Ours} is correctly rounded on every sample.

\subsection{Categorical (probs and logits; $K=2,4,16$)}
\label{app:cat}
\textbf{Official.} Over $K$ categories with probabilities $p_k$ ($P$) and $q_k$ ($Q$): $\KL=\sum_kp_k(\log p_k-\log q_k)$, term by term, with terms where $q_k=0$ set to $+\infty$. The constructor produces probabilities and log-probabilities first: \texttt{probs} uses $p=x/\sum x$ for the raw parameters $x$ and $\log(\mathrm{clamp}(p,\texttt{eps},1-\texttt{eps}))$; \texttt{logits} uses $z-\mathrm{logsumexp}(z)$ and \texttt{softmax} for the raw logits $z$.

\textbf{Error.} Five sources. (i) \emph{Cancellation}: for close distributions each term is first order ($\approx p_k\rho_k$) and the sum second order, because $\sum_kp_k\rho_k=0$ exactly; in addition $\log p_k$ and $\log q_k$ carry errors of about $u|\log p_k|$. (ii) \emph{Clamp} (probs): probabilities below \texttt{eps} are replaced by \texttt{eps} in value and gradient. (iii) \emph{Spurious $+\infty$} (logits): \texttt{softmax} underflows, $q_k$ is computed as 0 and the term is set to $\infty$ although the true probability is positive. (iv) \emph{Normalisation}: rounding of $p=x/\sum x$ makes $\sum p\ne1$, which is amplified in the gradient. (v) \emph{Argument validation} (FP16 probs): the simplex check uses a tolerance of $10^{-6}$, below FP16 resolution, and rejects about 25\% of valid inputs. FP64 evaluation leaves 32--38\% for FP32 inputs.

\textbf{Repair, probs} (raw parameters $x$ of $P$ and $y$ of $Q$; $S_x=\sum x$, $S_y=\sum y$). With $\rho_k=q_k/p_k-1$ and the linear term removed by $\sum_kp_k\rho_k=0$,
\[\KL=\sum_kp_k\,\varphi(\rho_k),\qquad\rho_k=\frac{N_k}{x_kS_y},\qquad N_k=y_kS_x-x_kS_y,\]
\[\frac{\partial\KL}{\partial y_k}=\frac{N_k}{y_kS_xS_y},\qquad\frac{\partial\KL}{\partial x_k}=-\frac{\lgp(\rho_k)+\KL}{S_x}=\frac1{S_x}\sum_{j\ne k}p_j\log\frac{y_jx_k}{x_jy_k}.\]
$S_x,S_y$ are exact expansions; $N_k$ is computed exactly from $4K$ Dekker products. Of the two equivalent forms of $\partial_{x_k}$ the one with the smaller sum of absolute terms is used.

\textbf{Repair, logits} (raw logits $z$ of $P$ and $w$ of $Q$; $d_k=w_k-z_k$). For any constant $a$ and $t_k=d_k-a$,
\[\KL=\lgp(E)-M=\sum_kp_k\,\xi(t_k)-\varphi(E),\quad E=\sum_kp_k\,\emo(t_k),\quad M=\sum_kp_kt_k,\]
\[\frac{\partial\KL}{\partial z_k}=-p_k\,(t_k-M),\qquad\frac{\partial\KL}{\partial w_k}=p_k\big(e^{t_k-L}-1\big),\quad L=\lgp(E).\]
The centre is chosen as follows: with $r=\arg\max z$ and $s_k=d_k-d_r$, $c=\fl(\sum p_ks_k)$ and $t_k=s_k-c$, so that $M$ is of rounding size. $p_k=e^{z_k-z_{\max}}/Z$, with $z_{\max}=\max_jz_j$ and $Z=\sum_je^{z_j-z_{\max}}$, is carried as (mantissa, exponent) and cannot underflow to zero; $s_k,t_k$ are exact expansions of the inputs. For $E>1$ the form $\log(\sum p_ke^{t_k})-M$ is used. Neither \texttt{softmax}, \texttt{log\_softmax} nor a clamp is called.
The vectorised form used downstream (\cref{eq:logits}) is an equivalent arrangement of this identity, with the weights taken from the second argument, and is tested against the certified implementation. It is evaluated in FP64 and rounded once. The difference $a_j-c_j$ is centred at the largest student logit, so that a common offset between $a$ and $c$ gives exactly zero value and gradient; its rounding error, if any, is recovered by TwoSum and added back; and where $\delta_j$ is large, $q_je^{\delta_j}$ is formed as $\exp(\log q_j+\delta_j)$. It is not itself certified by interval arithmetic.

\textbf{Zero probabilities.} A category with $p_k=0<q_k$ contributes $\lim_{p\downarrow0}p\,\varphi(q_k/p-1)=q_k$. Such inputs are reduced to the support $S$ of $P$: $\KL(P\|Q)=\KL(P_S\|Q_S/Q(S))-\log Q(S)$, a sum of two non-negative terms; $q_k=0<p_k$ gives $+\infty$.

\textbf{Why it works.} The cancellation is done algebraically: the \texttt{probs} KL is a sum of non-negative terms, while in the \texttt{logits} KL, $S=\sum_kp_k\xi(t_k)$ satisfies $E=M+S$ and $\KL=S-\varphi(E)$. When $|M|\ll S\ll1$, $\varphi(E)/S\approx S/2$; for $E>1$ the other form is used. $N_k$ is exact, so the relative error of $\rho_k$ does not depend on how close $p$ and $q$ are. Proportional \texttt{probs} and shifted \texttt{logits} (different parameters, same distribution) give exactly $+0$. Residual cancellation in the same-precision variant occurs where a gradient component itself crosses zero; there its ULP error can be large while its error relative to the gradient vector is of order $u$.

\textbf{Frozen data} (372,429 probs; 367,746 logits). KL, probs: \A{} 46.6--47.6 / 86.0--86.9\%; \textsc{Off.+} 3.4--4.6 / 32.2--32.5\%; \textsc{Rewrite} $\le0.01\%$. KL, logits: \A{} 59.4--62.7 / 88.3--90.5\%; \textsc{Off.+} 12.4--18.3 / 35.8--38.4\%; \textsc{Rewrite} $<0.01\%$ / 0. Gradients: \A{} 47.5--65.3 / 75.0--88.9\%; \textsc{Rewrite} at most 0.81\% ($Q$-side, logits $K=16$, FP32). \textsc{Rewrite} returns $\infty$ for a batch of \texttt{probs} values at the overflow threshold, eight of which ($K=2$) have a finite true KL and are genuine errors. \textsc{Ours} is correctly rounded on every sample.

\subsection{ContinuousBernoulli (probs and logits)}
\textbf{Official.} With probability parameters $\lambda_p,\lambda_q$, natural parameters $\eta=\mathrm{logit}(\lambda)$, mean $\mu_p$ of $P$ and log-normaliser $C(\lambda)$: $\KL=\mu_p(\eta_p-\eta_q)+[C(\lambda_p)+\lgp(-\lambda_p)]-[C(\lambda_q)+\lgp(-\lambda_q)]$. The constructor clamps $\lambda$ to $[\texttt{eps},1-\texttt{eps}]$ (for \texttt{logits} after a sigmoid, while $\eta$ keeps the raw value); $C$ and the mean use a truncated Taylor polynomial for $\lambda\in(0.499,0.501]$ and closed forms elsewhere.

\textbf{Error.} (i) \emph{Clamp}: outside $[\texttt{eps},1-\texttt{eps}]$ the normaliser and mean belong to a different distribution than the input; with \texttt{logits} $z$ this happens for $|z|>\log(1/\texttt{eps})$ (about 6.9, 15.9, 36 for FP16, FP32, FP64), where one term uses raw logits and the other two clamped probabilities, so the result is wrong in any precision. (ii) Near pairs: three first-order terms with a second-order sum. (iii) Near $\lambda=\tfrac12$ outside the window: the mean is a difference of two numbers $\approx1/(2|2\lambda-1|)$, cancelling by a factor of about 500 at the window edge. (iv) Inside the window the polynomial is an approximation, not an identity. (v) NaN gradients in FP16 from the unselected branch of \texttt{torch.where}.

\textbf{Repair.} Write the log-partition function as $A(\eta)=\eta/2+g(\eta)$ with
\[g(x)=\log\frac{\sinh(x/2)}{x/2}=\sum_{k\ge1}c_kx^{2k},\qquad c_k=\frac{B_{2k}}{2k\,(2k)!},\]
where $B_{2k}$ are the Bernoulli numbers,
so that the linear part cancels exactly and, with $p=\eta_p$, $q=\eta_q$, $d=q-p$,
\[\KL=g(q)-g(p)-d\,g'(p),\qquad\partial_p\KL=-d\,g''(p),\qquad\partial_q\KL=g'(q)-g'(p).\]
Three regions in $(|p|,|q|)$, each with an exact identity: (A) both $\le4$: $\KL=d^2\sum_kc_kH_{2k-2}(p,q)$ with the polynomials $H_{n-2}$ defined by $q^n-p^n-np^{n-1}d=d^2H_{n-2}(p,q)$, leading term $d^2/24$, with 17/27/50 terms; (B) same sign, both away from 0: $\KL=\varphi(\rho)-\varphi(\tau)-h_p\,\xi(-D)$ with $D=|q|-|p|$, $\rho=D/|p|$, $E_p=e^{-|p|}$, $E_q=e^{-|q|}$, $\tau=(E_p-E_q)/(1-E_p)$ and $h_p=E_p/(1-E_p)$; (C) far pairs: $g(|q|)+[|p|g'(|p|)-g(|p|)]-\mathrm{sgn}(pq)\,|q|\,g'(|p|)$, whose three terms are non-negative for opposite signs.
For \texttt{probs}, $d=\log(\lambda_q/\lambda_p)+\log(\bar\lambda_p/\bar\lambda_q)$ with $\bar\lambda=1-\lambda$ by TwoSum; the two logarithms have the same sign, and the factor $1/(\lambda(1-\lambda))$ of the gradient is merged into one product with one rounding.
Midpoint-aware final rounding on the \texttt{logits} path.

\textbf{Why it works.} No clamp: logits up to $|z|=1005$ and subnormal $\lambda$ are evaluated for the distribution that was passed in. The linear term is removed before any arithmetic; the series converges for $|\eta|\le4<2\pi$, with the remainders of all six series asserted $\le u/16$ at import time; there is no window. The remaining cancellation is bounded by a factor of 5--9 per region and enters the error estimate.

\textbf{Frozen data} (342,070 probs; 376,691 logits). KL, probs: \A{} 78.37 / 89.75\%; \textsc{Off.+} 6.27 / 38.92\%; \textsc{Rewrite} 0.09 / 0.04\%. KL, logits: \A{} 86.17 / 90.89\%; \textsc{Off.+} 27.81 / 42.56\% (5.07\% for FP16 inputs in FP64); \textsc{Rewrite} $<0.01\%$. Gradients: \A{} 78.6--90.8\%. For 938 \texttt{probs} gradients \textsc{Rewrite} returns $\infty$ where the true value exceeds the format; these are not counted as errors but carry no finite error guarantee.

\subsection{MultivariateNormal ($d=2,4$; \texttt{scale\_tril})}
\textbf{Official.} With dimension $d$, means $\mu_p,\mu_q$, lower-triangular Cholesky factors $L_p,L_q$ of the covariances (\texttt{scale\_tril}), $M=L_q^{-1}L_p$ and $v=L_q^{-1}(\mu_p-\mu_q)$:
$\KL=\big[\sum_i\log L_{q,ii}-\sum_i\log L_{p,ii}\big]+\tfrac12\big(\|M\|_F^2+\|v\|^2-d\big)$.

\textbf{Error.} For close pairs $\|M\|_F^2-d$ and the log-determinant difference are both first order and their sum is second order; the log-determinant difference itself cancels. The dominant cause is cancellation, not ill-conditioning: the KL error rate is similar for condition numbers $\le100$, up to $10^4$, and for large means. 16.4\% ($d=2$) and 11.1\% ($d=4$) of non-identity outputs are negative. FP64 evaluation fixes most gradients but leaves about 36\% of the KL values.

\textbf{Repair.} $M$ is lower triangular with positive diagonal. With $N=M-I$ and, for each diagonal entry $n=N_{ii}$, $(1+n)^2-1-2\log(1+n)=n^2+2\varphi(n)$,
\[\KL=\sum_i\varphi(N_{ii})+\tfrac12\|N\|_F^2+\tfrac12\|v\|^2,\qquad N=L_q^{-1}(L_p-L_q),\]
\[\partial_{\mu_p}=-\partial_{\mu_q}=L_q^{-\top}v,\quad\partial_{L_p}=\mathrm{tril}\big(L_q^{-\top}N\big)+\mathrm{diag}\big(N_{ii}/L_{p,ii}\big),\quad\partial_{L_q}=-\mathrm{tril}\big(L_q^{-\top}(N+N^\top+NN^\top+vv^\top)\big).\]
$N$ is obtained by forward substitution from the exact difference $L_p-L_q$ (TwoSum), never as $M-I$. All matrix operations are element-wise in double-length arithmetic ($d\le4$; no call to \texttt{torch.linalg}).

\textbf{Why it works.} Three non-negative terms; $N$ is small because it is solved from an exact small right-hand side; mean-only perturbations give $N=0$ and covariance-only perturbations give $v=0$ exactly. Double-length arithmetic absorbs the condition number of the triangular solves: solving in plain FP32 would be worse than the official code on 5--15\% of the low-rank outputs.

\textbf{Frozen data} (202,000; FP32 only, because the official FP16 implementation is unsupported). KL: \A{} 88.56 / 88.86\% ($d=2/4$); \textsc{Off.+} 35.81 / 35.78\%; \textsc{Rewrite} 0 (max.\ 2 ULP). Gradients: \A{} 27.1--43.0\%; \textsc{Off.+} 4.2--7.5\%; \textsc{Rewrite} all correctly rounded. \textsc{Ours} is correctly rounded on every sample. Gradient components in the strict upper triangle of \texttt{scale\_tril} are not scored; the official code leaks non-zero gradient there.

\subsection{LowRankMultivariateNormal ($d=2,4$; $r=1,2$)}
\textbf{Official.} With covariances $\Sigma=WW^\top+\mathrm{diag}(D)$, where $W\in\mathbb R^{d\times r}$ is the factor of rank $r$ and $D$ the positive diagonal, the Woodbury identity and capacitance matrices give
$\KL=\tfrac12\big[\log\det\Sigma_q-\log\det\Sigma_p+\mathrm{tr}(\Sigma_q^{-1}\Sigma_p)+(\mu_q-\mu_p)^\top\Sigma_q^{-1}(\mu_q-\mu_p)-d\big]$, each part computed separately.

\textbf{Error.} The same second-order cancellation as for the full-covariance case; in addition $D$ is swamped by the rounding of $WW^\top$ when the covariance is assembled; and the parameterisation is not unique (column permutations and sign flips of $W$, transfers between $WW^\top$ and $D$), so $P=Q$ can hold with different parameters. 19.8--21.3\% of non-identity outputs are negative; in FP64 about 40\% of the KL values are still wrong and 0.54\% negative.

\textbf{Repair.} We return to the form of the previous subsection without forming the Cholesky factor of $\Sigma_p$. With $L_q=\mathrm{chol}(\Sigma_q)$, $\delta=\mu_p-\mu_q$, $v=L_q^{-1}\delta$, $\Delta W=W_p-W_q$ and $\Delta D=D_p-D_q$,
\[E=\Sigma_p-\Sigma_q=\Delta W\,W_p^\top+W_q\,\Delta W^\top+\mathrm{diag}(\Delta D),\qquad X=L_q^{-1}EL_q^{-\top},\qquad I+X=GG^\top,\]
and $N=G-I$ is computed by a \emph{differential} Cholesky recursion,
\[s_j=X_{jj}-\sum_{t<j}G_{jt}^2,\qquad N_{jj}=\frac{s_j}{1+\sqrt{1+s_j}},\qquad G_{ij}=\frac{X_{ij}-\sum_{t<j}G_{it}G_{jt}}{G_{jj}},\]
after which $\KL=\sum_i\varphi(N_{ii})+\tfrac12\|N\|_F^2+\tfrac12\|v\|^2$.
Gradients: $H_p=L_q^{-\top}X(I+X)^{-1}L_q^{-1}$, $H_q=L_q^{-\top}\big[L_q^{-1}(E+\delta\delta^\top)L_q^{-\top}\big]L_q^{-1}$, and $\partial_{W_p}=H_pW_p$, $\partial_{D_p}=\tfrac12\mathrm{diag}(H_p)$, $\partial_{W_q}=-H_qW_q$, $\partial_{D_q}=-\tfrac12\mathrm{diag}(H_q)$, $\partial_{\mu_p}=L_q^{-\top}v=-\partial_{\mu_q}$.
$E$ is built from the exact parameter differences. Three refinements are necessary: the double-length square root is corrected to be correctly rounded (the vectorised platform square root differed by 1 ULP from the correctly rounded one on 17.2\% of FP32 and 1.3\% of FP64 results); $H_q$ adds $E+\delta\delta^\top$ \emph{before} solving, so that an exactly vanishing sum gives an exact zero; and semantic identity is decided exactly: if $\mu_p=\mu_q$ and $W_pW_p^\top+\mathrm{diag}(D_p)=W_qW_q^\top+\mathrm{diag}(D_q)$ as rational numbers (tested with error-free expansions, no tolerance), all outputs are $+0$.

\textbf{Why it works.} $L_qG$ is the Cholesky factor of $\Sigma_p$, so the non-negative formula of the full-covariance case applies with $M=G$; $E$ is an exact small quantity; $\sqrt{1+s}-1$ is written without cancellation.

\textbf{Frozen data} (202,000; FP32 only). KL: \A{} 91.1--93.4\%; \textsc{Off.+} 39.5--41.9\%; \textsc{Rewrite} 0 (max.\ 3 ULP). Gradients: \A{} 59.3--71.0\%; \textsc{Off.+} 10.8--24.4\%; of about 4.24 million gradient values \textsc{Rewrite} misrounds 72 and has one significant error, shared with \A. \textsc{Ours} is correctly rounded on every sample.
\section{Interface B: \texttt{torch.nn.functional.kl\_div}}
\label{app:ifaceB}

\subsection{Official implementation and its errors}

Element-wise, the official code computes $\ell=\mathrm{xlogy}(t,t)-t\,x$, where $\mathrm{xlogy}(t,t)=t\log t$ and $0$ at $t=0$ (\texttt{log\_target=False}), or $\ell=e^{t}(t-x)$ (\texttt{log\_target=True}), then sums and divides by a factor \emph{den} that depends on the reduction (1 for \texttt{sum}, the batch size for \texttt{batchmean}, the number of elements for \texttt{mean}). Gradients are produced by autograd. We found six classes of behaviour; the first five are operator errors that a better implementation removes, and the first four of them also disappear in FP64.

\begin{enumerate}
\item \textbf{Element-wise cancellation (\texttt{log\_target=False}).} When $t\approx e^{x}$, $\log t$ and $x$ are nearly equal, and the rounding error $u|\log t|$ of the logarithm is amplified by the subtraction. In FP32, 65--99\% of \texttt{sum} outputs are off by $\ge8$ ULP and 35--70\% by $\ge1\%$, the more so the smaller the KL. With \texttt{log\_target=True} the element-wise error is small ($\le0.3\%$ at $\ge8$ ULP).
\item \textbf{Gradient with respect to the target under scaling.} Autograd writes the derivative of \texttt{xlogy} in its second argument as $(g\,t)/t$, where $g$ is the upstream factor ($1/(BK)$ for \texttt{mean}, with batch size $B$ and $K$ classes). The exact value of this term is $g$, but the product is formed first: when $g\,t$ is subnormal it loses precision and when it underflows the term vanishes. With FP16 and $K=256$, 99.6\% of gradient rows have relative $L_2$ error $\ge1\%$ and 95.4\% contain a component of the wrong sign; with $K=1024$ every row has cosine below 0.9. The same term makes the gradient $0/0=$ NaN at $t=0$. Evaluating $(g\log t+(g\,t)/t)+(-g)\,x$ in this order reproduces the official gradient bit for bit on 39,513,600 components (including 57,706 NaNs).
\item \textbf{Components rounded to zero under \texttt{mean} (\texttt{log\_target=True}).} $e^{t}$ is rounded once and its product with $g$ then lands on a midpoint of the subnormal grid and is rounded to 0 (20.1\% of target-gradient rows and 14.6\% of input-gradient rows for FP16, $K=256$).
\item \textbf{Overflow before scaling.} The exact \texttt{mean}/\texttt{batchmean} is finite but the sum overflows first (extreme strata only).
\item \textbf{NaN at boundary inputs.} \texttt{log\_target=True} with a target of $-\infty$ (probability zero) returns NaN on all 20,400 such batches; so do $x=-\infty$ with $t=0$ and the target gradient at $t=0$. These are not removed by higher precision.
\item \textbf{Input-representation error.} $T_{\mathrm{op}}$ differs from $T_{\mathrm{e2e}}$ by $\ge1\%$ on 40--99\% of the batches; evaluating the operator in FP64 changes the total error by less than 3 percentage points; with \texttt{log\_target=False} two identical distributions give a non-zero $T_{\mathrm{op}}$. This cannot be repaired inside the function.
\end{enumerate}

\subsection{Repair}

The repair keeps the signature (plus a \texttt{method} argument) and computes in FP64 with merged formulas and a single final rounding.

\paragraph{Forward, \texttt{log\_target=False}.} Write $t=m\,2^{k}$ with an integer $k$ and $m\in[\sqrt{1/2},\sqrt2)$. Then $\log t-x=(k\,\ln2_{\mathrm{hi}}-x)+\log m+k\,\ln2_{\mathrm{lo}}$, where $\ln2=\ln2_{\mathrm{hi}}+\ln2_{\mathrm{lo}}$ is split so that $\ln2_{\mathrm{hi}}$ has few significant bits; $k\,\ln2_{\mathrm{hi}}$ is then exact and cancels against $x$ without error; the remaining error comes from $|\log m|<0.35$. The four terms are accumulated with TwoSum.

\paragraph{Forward, \texttt{log\_target=True}.} $e^{t}$ and $t-x$ are multiplied in FP64; if $t-x=0$ the result is 0 regardless of $e^{t}$.

\paragraph{Reduction.} Compensated pairwise summation of the unrounded element values, division by \emph{den}, one rounding.

\paragraph{Backward.} Merged analytic gradients replace autograd's term-by-term evaluation ($G$ is the incoming gradient of the reduced loss, so that $g=G/\mathit{den}$ above):
\begin{center}\small
\begin{tabular}{lll}
\toprule
\texttt{log\_target} & $\partial/\partial x$ & $\partial/\partial t$\\
\midrule
False & $-G\,t/\mathit{den}$ & $G\,(\log t+1-x)/\mathit{den}$\\
True & $-G\,e^{t}/\mathit{den}$ & $G\,e^{t}\,(t-x+1)/\mathit{den}$\\
\bottomrule
\end{tabular}
\end{center}
The bracket $\log t+1-x$ uses the forward decomposition; $t-x+1$ is an error-free three-term sum, which also decides exactly whether it is zero. Each component is resolved in a fixed order: outside the domain $\to$ non-finite $G$ $\to$ $G=0$ $\to$ undefined cell (NaN) $\to$ exact zero factor $\to$ formula.

\paragraph{Certified rounding.} When an intermediate of the fast path is not finite (e.g.\ $e^{t}$ overflows although the result is finite), or when the FP64 value is closer to a rounding boundary of the target format than the a-priori error bound of the fast path, the sum is rewritten as $R+\sum_\tau c_\tau f(\tau)$, where $\tau$ runs over the distinct target values, $f$ is $\exp$ (\texttt{log\_target=True}) or $\log$ (\texttt{log\_target=False}) and $R,c_\tau$ are exact rationals, and evaluated by interval arithmetic on a precision ladder from 128 to 16,384 bits until both end points round to the same value. If they still differ at 16,384 bits, the implementation raises an error and returns no value; this did not occur on the benchmark. The guard is necessary, e.g.\ with \texttt{log\_target=True} for $t=(0,0,-1000)$, $x=(-1,-2^{-24},-1001)$, whose exact value $1+2^{-24}+e^{-1000}$ sits next to a rounding midpoint that the dropped $e^{-1000}$ decides.
FP64$\to$FP16 rounding goes through a round-to-odd FP32 value, which is equivalent to a single rounding.

\paragraph{Why it works.} The large term $k\ln2_{\mathrm{hi}}$ is exact, so the residual error is independent of $|\log t|$; the backward pass never forms $(g\,t)/t$; the unit roundoff of FP64 is $2^{29}$ (FP32) or $2^{42}$ (FP16) times smaller than that of the target, and the few results too close to a rounding boundary are certified by interval arithmetic. The guard uses an a-priori error bound of the fast path that assumes FP64 \texttt{log} and \texttt{exp} accurate to 1 ULP; on the benchmark, every scored component equals the certified rounding. Zeros ($t=x$, $t-x+1=0$, $G=0$) are decided exactly.

\subsection{Results on the frozen data}

\begin{center}\small
\begin{tabular}{lrr}
\toprule
 & forward & backward\\
\midrule
scored components & 83,899,496 & 483,279,156\\
official: correctly rounded & 52.5\% & 74.9\%\\
ours (FP64 method): correctly rounded & 100\% & 100\%\\
components repaired & 39,822,967 & 121,446,622\\
regressions & 0 & 0\\
semantic violations (NaN, sign, zero) & 0 & 0\\
\bottomrule
\end{tabular}
\end{center}

\paragraph{Subnormal gradients.} For 233 FP16 vectors with subnormal reference gradients, componentwise correct rounding improves relative $L_2$ error but can reduce cosine similarity. All corresponding components are correctly rounded.

\paragraph{Comparison methods.} Evaluating the repaired formulas in FP32 for FP16 inputs fully repairs 7 of 10 forward cells; evaluating them in the input precision repairs 4 of 20 forward and 0 of 20 backward cells completely, because correct rounding is lost and FP16 subnormal results are rounded twice. This is why the strict method computes in FP64.

\paragraph{Cost and scope.} The reference implementation is 28--48$\times$ slower on CPU ($B=64$, $K=1024$); \cref{tab:cost_b} reports CPU and GPU costs, reaching 222$\times$ for larger inputs. The full benchmark was repeated on an A100 with identical correctly rounded component counts (\cref{app:gpu}).

\paragraph{What the operator repair does not do.} In the distillation experiment (\cref{sec:exp2}), replacing \fkl{} alone gives bit-identical training trajectories; the spurious gradient arises in \texttt{log\_softmax} backward. In the language-model experiment (\cref{sec:exp3}), the dominant error comes from rounded probability inputs. The logits path of \cref{eq:logits} addresses both sources.

\subsection{Cost of the reference implementations}
\label{app:cost}

\Cref{tab:cost,tab:cost_b,tab:cost_logits} give running times relative to the official call (forward plus backward; median of 7 repetitions, each the mean of several calls) on one A100 GPU and on four CPU threads of the same node (PyTorch 2.4.0).
The node was shared with other jobs, so the CPU columns are less reliable than the GPU columns.
None of the repairs has been optimised.

\paragraph{Closed forms.} The official code and ours are both compositions of tensor operations in Python.
On the GPU \textsc{Rewrite} costs 2.4--5.9$\times$ the official call for Exponential, Laplace, Poisson, Normal and Bernoulli, and the strict mode 3.4--12$\times$.
Gamma (7--14$\times$ and about 30$\times$) and Beta (54--116$\times$ and about 320$\times$) are much slower because their repairs evaluate a quadrature.
Precision is not the cost: the official formulas in FP64 run at 1.0--1.1$\times$ on the GPU (1.6$\times$ for Beta at the larger size), and the overhead of the strict mode over \textsc{Rewrite} comes from its additional rounding and conversion steps.
The cost of \textsc{Rewrite} is the additional arithmetic: series instead of one logarithm, all branches of a piecewise formula evaluated and then selected, double-length sums, and hand-written gradients.
On CPU the factors are larger (\textsc{Rewrite} 2.7--22$\times$ for the five families).

\paragraph{Interface B.} The official \fkl{} is a compiled composite; the repair is 27--59$\times$ slower on the GPU and 28--222$\times$ on CPU, growing with the number of classes.
Skipping the boundary check that certifies the final rounding saves 24--31\% on the GPU.
The remainder is error-free arithmetic and the unfused evaluation of many small tensor operations.

\paragraph{Logits path.} It evaluates a 16-term series in FP64 for every element.
On the GPU it costs 3--10$\times$ the official call for FP32 logits and 5--22$\times$ for FP16 or BF16 logits, the latter at a vocabulary of 151,936, where the official call in FP64 is itself 4--5$\times$ slower than in half precision; one call on $512\times151{,}936$ BF16 logits takes 72\,ms against 3.2\,ms.

\begin{table}[h]
\caption{Closed forms: running time relative to the official FP32 call, for 65,536 / 1,048,576 pairs of FP32 parameters. ``FP64'': the official formula evaluated in FP64.}
\label{tab:cost}
\centering\footnotesize
\begin{tabular}{lcccccc}
\toprule
 & \multicolumn{3}{c}{GPU (A100)} & \multicolumn{3}{c}{CPU (4 threads)}\\
\cmidrule(lr){2-4}\cmidrule(lr){5-7}
 & FP64 & \textsc{Rewrite} & \textsc{Ours} & FP64 & \textsc{Rewrite} & \textsc{Ours}\\
\midrule
Exponential & 1.1 / 1.0 & 2.5 / 2.9 & 3.5 / 8.0 & 1.6 / 3.1 & 3.3 / 7.2 & 6.9 / 17\\
Laplace & 1.0 / 1.1 & 2.6 / 3.1 & 3.6 / 5.4 & 1.7 / 2.4 & 4.3 / 5.4 & 8.8 / 12\\
Poisson & 1.0 / 1.1 & 2.4 / 2.7 & 3.4 / 5.0 & 1.8 / 2.8 & 2.7 / 7.3 & 5.4 / 18\\
Normal & 1.0 / 1.0 & 4.7 / 2.6 & 5.4 / 7.0 & 1.4 / 2.9 & 11 / 22 & 27 / 40\\
Bernoulli (logits) & 1.0 / 1.0 & 5.9 / 4.8 & 7.8 / 12 & 1.7 / 2.5 & 20 / 19 & 56 / 43\\
Gamma & 1.0 / 1.0 & 14 / 7.3 & 32 / 30 & 1.5 / 1.6 & 12 / 12 & 38 / 41\\
Beta & 1.1 / 1.6 & 116 / 54 & 326 / 322 & 1.6 / 1.4 & 41 / 51 & 150 / 393\\
\bottomrule\relax  
\end{tabular}
\end{table}

\begin{table}[h]
\caption{\fkl: running time relative to the official call in the input precision (\texttt{batchmean}). ``no check'': the repair with the boundary check of the final rounding skipped.}
\label{tab:cost_b}
\centering\footnotesize
\begin{tabular}{lllcccccc}
\toprule
 & & & \multicolumn{3}{c}{GPU (A100)} & \multicolumn{3}{c}{CPU (4 threads)}\\
\cmidrule(lr){4-6}\cmidrule(lr){7-9}
inputs & \texttt{log\_target} & $B\times K$ & FP64 & no check & repaired & FP64 & no check & repaired\\
\midrule
FP32 & False & $64\times1{,}024$ & 1.1 & 22 & 29 & 1.3 & 27 & 33\\
FP32 & True & $64\times1{,}024$ & 1.1 & 20 & 27 & 3.0 & 22 & 28\\
FP16 & False & $64\times1{,}024$ & 1.1 & 24 & 33 & 1.0 & 23 & 30\\
FP16 & True & $64\times1{,}024$ & 1.1 & 23 & 32 & 3.9 & 34 & 48\\
FP32 & False & $64\times32{,}768$ & 1.4 & 39 & 52 & 1.9 & 36 & 48\\
FP32 & True & $64\times32{,}768$ & 1.3 & 32 & 44 & 3.6 & 69 & 93\\
FP16 & False & $64\times32{,}768$ & 1.3 & 43 & 59 & 1.7 & 38 & 54\\
FP16 & True & $64\times32{,}768$ & 1.3 & 35 & 51 & 8.2 & 147 & 222\\
\bottomrule\relax  
\end{tabular}
\end{table}

\begin{table}[h]
\caption{Logits path: running time relative to \fkl{} on \texttt{log\_softmax} outputs (\texttt{log\_target=True}) in the precision of the logits. ``FP64'': the same official call on logits cast to FP64.}
\label{tab:cost_logits}
\centering\footnotesize
\begin{tabular}{llccccc}
\toprule
 & & \multicolumn{3}{c}{GPU (A100)} & \multicolumn{2}{c}{CPU (4 threads)}\\
\cmidrule(lr){3-5}\cmidrule(lr){6-7}
logits & $B\times V$ & official (ms) & FP64 & logits path & FP64 & logits path\\
\midrule
FP32 & $64\times1{,}024$ & 0.19 & 1.2 & 3.1 & 2.4 & 4.4\\
FP32 & $64\times32{,}768$ & 0.21 & 2.0 & 8.0 & 3.1 & 8.5\\
FP32 & $32\times151{,}936$ & 0.49 & 2.6 & 8.8 & 2.8 & 14\\
FP32 & $512\times151{,}936$ & 6.20 & 2.3 & 10 & -- & --\\
BF16 & $64\times1{,}024$ & 0.19 & 1.2 & 4.6 & 7.2 & 13\\
BF16 & $64\times32{,}768$ & 0.20 & 2.0 & 9.5 & 4.4 & 17\\
BF16 & $32\times151{,}936$ & 0.25 & 4.9 & 20 & 4.4 & 26\\
BF16 & $512\times151{,}936$ & 3.24 & 4.3 & 22 & -- & --\\
FP16 & $64\times1{,}024$ & 0.20 & 1.2 & 4.6 & -- & --\\
FP16 & $64\times32{,}768$ & 0.20 & 1.9 & 9.4 & -- & --\\
FP16 & $32\times151{,}936$ & 0.25 & 4.9 & 20 & -- & --\\
FP16 & $512\times151{,}936$ & 3.24 & 4.3 & 22 & -- & --\\
\bottomrule\relax  
\end{tabular}
\end{table}
\section{Repetition on a GPU}
\label{app:gpu}

\paragraph{Setup.} All measurements of \cref{sec:measure} and all repairs were repeated on one NVIDIA A100 (40\,GB) with PyTorch 2.4.0+cu124, on the same frozen inputs and against the same reference values, with the scoring code of the CPU study.
Interface~A: the 33 configurations and sub-configurations (2,707,000 scored FP16 and 4,527,454 scored FP32 samples); the official implementation and the repairs were run on the GPU and, in the same PyTorch build, on the CPU for a bit-wise comparison.
Interface~B: the 20 configurations (942,200 batches, 83.9 million forward and 483.3 million backward components).
The multivariate-normal repairs were evaluated on the A100 using the same frozen inputs and references; both devices showed no significant errors.

\begin{figure}[h]
\centering
\includegraphics[width=\textwidth]{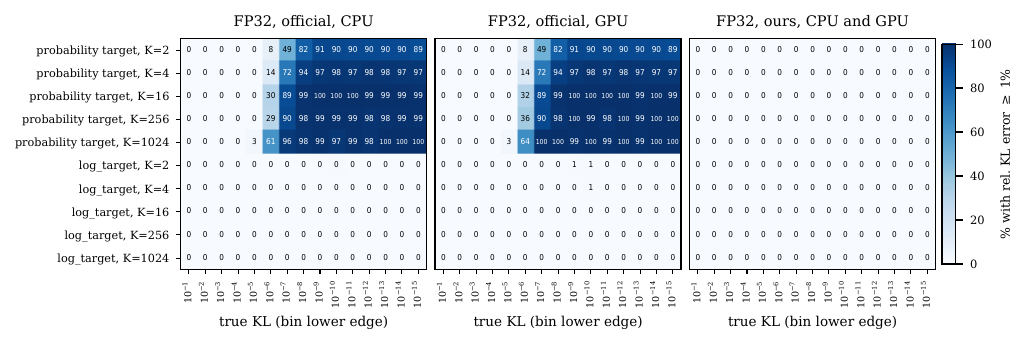}
\caption{Interface B, FP32 inputs, region R0, batches of one row, \texttt{reduction="sum"}. Share of samples whose output has relative error $\ge1\%$, per configuration and KL interval. The left and middle panels show the official \fkl{} on CPU and on GPU, and the right panel \textsc{Ours}, whose CPU and GPU results are identical (all zero).}
\label{fig:heatB32}
\end{figure}

\begin{figure}[h]
\centering
\includegraphics[width=\textwidth]{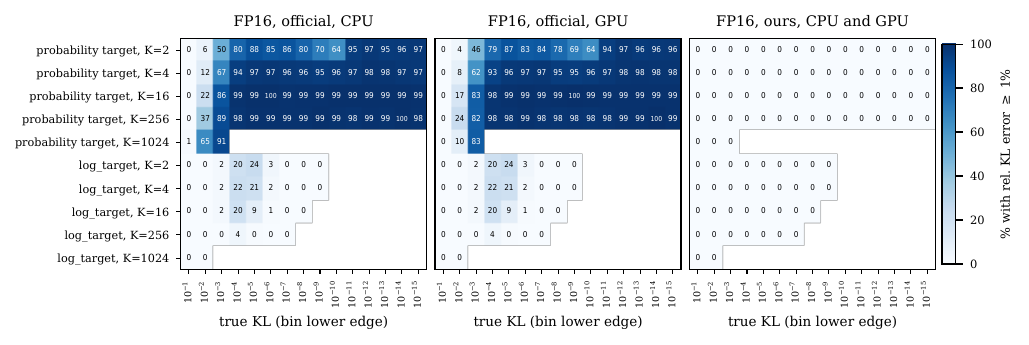}
\caption{Interface B, FP16 inputs, as \cref{fig:heatB32}. Blank cells have fewer than 100 samples.}
\label{fig:heatB16}
\end{figure}

\paragraph{Results.} \Cref{tab:gpu} summarises the two devices; \cref{fig:heat32,fig:heat16,fig:heatB32,fig:heatB16} resolve the relative error by configuration and KL interval.
The official implementations fail on the GPU as they do on the CPU.
For interface~A the rates agree within 0.6 points for all configurations with two exceptions: Gumbel with FP16 inputs (significant-error rate 93.5\% on the GPU, 66.2\% on the CPU; the share with relative error $\ge1\%$ in R0 is practically the same, 73.2\% and 73.3\%) and MultivariateNormal with $d=2$ (86.6\% against 88.6\%).
For \fkl{} with \texttt{log\_target=True} and FP32 inputs, 28--50\% of the forward and 32--34\% of the backward components are not correctly rounded on the GPU, against 17--44\% and 11--13\% on the CPU.
The opposite also occurs: with FP16 probability targets and many classes the \texttt{sum} of the official implementation is more accurate on the GPU ($K=1024$, $\KL\in[10^{-2},10^{-1})$: 10\% of the samples off by $\ge1\%$ against 65\% on the CPU).
\textsc{Ours} gives no significant error for interface~A on either device, and for \fkl{} the counts of correctly rounded components are the same on both devices for each of the 20 configurations.

\paragraph{The devices disagree bit-wise.} In the same PyTorch build, the KL value of the official interface-A implementation differs between CPU and GPU for 27.3\% of the FP32 and 5.6\% of the FP16 samples.
For \fkl{}, 12--13\% of the element-wise outputs and 33\% (FP16) to 49\% (FP32) of the \texttt{sum} outputs differ between the GPU run and the CPU run; this comparison is across the two PyTorch versions.

\begin{table}[h]
\caption{Official implementation and \textsc{Ours} on CPU (PyTorch 2.15.0) and on an A100 GPU (PyTorch 2.4.0), all KL intervals and parameter regions. Interface~A: significant-error rate of the KL value (\%). Interface~B: share of components that are not correctly rounded (\%); the backward rate of the official implementation excludes the boundary-gradient components.}
\label{tab:gpu}
\centering\small
\begin{tabular}{llrrrr}
\toprule
 & & \multicolumn{2}{c}{\A} & \multicolumn{2}{c}{\textsc{Ours}}\\
\cmidrule(lr){3-4}\cmidrule(lr){5-6}
 & inputs & CPU & GPU & CPU & GPU\\
\midrule
Interface A, KL value & FP16 & 70.08 & 71.47 & 0 & 0\\
 & FP32 & 87.67 & 87.66 & 0 & 0\\
Interface B, forward & FP16 & 38.74 & 38.79 & 0 & 0\\
 & FP32 & 56.19 & 61.15 & 0 & 0\\
Interface B, backward & FP16 & 21.95 & 21.95 & 0 & 0\\
 & FP32 & 25.16 & 35.69 & 0 & 0\\
\bottomrule
\end{tabular}
\end{table}

\section{Downstream Experiments: Details}
\label{app:downstream}

\subsection{Parameter Fitting}

\paragraph{Protocol.} Loss: \texttt{kl\_divergence(Geometric, Geometric)}, failure-count convention. Parameterisations: \texttt{probs}$=p=\sigma(\theta)$, with the logistic sigmoid $\sigma$, or \texttt{logits}$=\theta$. Directions: trainable first argument, $\KL(\mathrm{Geo}(p)\|\mathrm{Geo}(p^\star))$, and trainable second argument. The target is rounded to the computation precision first, so the rounded value is the exact optimum. Methods: official (PyTorch 2.15.0); the repaired formula in the target precision; the repaired formula in FP64 with one rounding of loss and gradient. Optimisers: Adam, learning rate $10^{-2}$, 3000 steps (main); Adam $10^{-3}$ and SGD with momentum 0.9 at 0.1, 10,000 steps (robustness). True KL by 60-digit arithmetic. CPU only.

\begin{table}[h]
\caption{Parameter fitting: 12 starts per target, shared by all settings and methods (log-uniform in $[p^\star/5,\min(5p^\star,0.95)]$; the target is rounded once to the format of the loss). Relative deviation of the fitted $p$ from the target after 3000 Adam steps (median / maximum over starts) and the largest true KL.}
\centering\small
\resizebox{\textwidth}{!}{%
\begin{tabular}{llll}
\toprule
Setting & Official: deviation & Official: max.\ true KL & Repaired: deviation\\
\midrule
FP16 probs, $\KL(\mathrm{Geo}(p)\|\mathrm{Geo}(p^\star))$, $p^\star=0.01$ & 3.7\% / 9.8\% & $4.3\times10^{-3}$ & 0.08\% / 0.31\%\\
FP16 probs, $\KL(\mathrm{Geo}(p)\|\mathrm{Geo}(p^\star))$, $p^\star=0.001$ & 95\% / 96\% & 1.7 & 0 / 0\\
FP16 logits, $\KL(\mathrm{Geo}(p)\|\mathrm{Geo}(p^\star))$, $p^\star=0.001$ & 95\% / 96\% (9 runs); 3 runs non-finite & 0.18 & 0 / 0\\
FP16 probs, $\KL(\mathrm{Geo}(p^\star)\|\mathrm{Geo}(p))$, $p^\star=0.001$ & 0 / 100\% (5 runs collapse to $p\approx0$) & 6.0 & 0 / 0\\
FP32 probs, $\KL(\mathrm{Geo}(p)\|\mathrm{Geo}(p^\star))$, $p^\star=0.001$ & 0.014\% / 0.027\% & $3.7\times10^{-8}$ & 0.00007\% / 0.00045\%\\
FP32 logits, $\KL(\mathrm{Geo}(p)\|\mathrm{Geo}(p^\star))$, $p^\star=0.01$ & 0.0012\% / 0.0032\% & $5.1\times10^{-10}$ & 0.00005\% / 0.0004\%\\
\bottomrule
\end{tabular}}
\end{table}

\begin{figure}[h]
\centering
\includegraphics[width=0.5\textwidth]{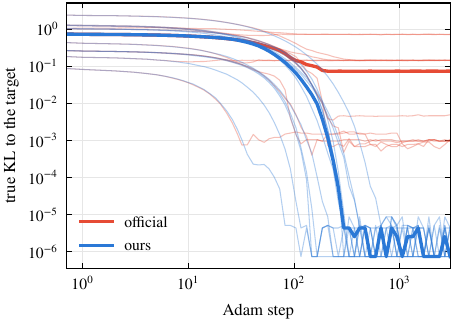}
\caption{Parameter fitting in FP16 with two Geometric targets ($p^\star=0.01$ and $0.002$) and Adam with learning rate $10^{-2}$. The curves show the exact KL to the target along the fit, from the same 12 starts for both losses (thin lines), and their median (thick). With the official loss (red) the runs stop improving after about 200 steps, at a median true loss of $0.074$, whereas with the repaired loss (blue) they reach the resolution of FP16.}
\label{fig:exp1full}
\end{figure}

\paragraph{Further observations.} With the trainable distribution in the second argument the official code usually works, but not always (fourth row). With a logit parameterisation, 3 of the 12 official FP16 runs for $p^\star=0.001$ abort with non-finite gradients. With SGD in FP16 and $p^\star=0.001$ the official loss diverges to $p=1$ from all three tested starts (true KL 6.9). In a two-parameter version (the sum of two Geometric KLs, targets $0.01$ and $0.002$ in FP16, $0.01$ and $0.001$ in FP32; same 12-start protocol) the final true loss over 12 starts is, median / maximum, $0.074$ / $0.73$ (official) against $7.5\times10^{-7}$ / $5.2\times10^{-6}$ (repaired) in FP16, and $6.9\times10^{-9}$ / $4.9\times10^{-7}$ against $4.0\times10^{-13}$ / $8.0\times10^{-12}$ in FP32. On a $15\times15$ grid around the region visited by the official FP16 trajectories, the official descent direction is within $25^\circ$ of the true one at 54\% of the points and more than $90^\circ$ away at 5\%; the official runs stall where the directions oppose.
\Cref{fig:exp1full} shows the true loss along the 12 runs.

\paragraph{The failure is not convergence to a biased point.} In the two-parameter version (\cref{fig:exp1full}), the official runs stop improving but do not stop moving: the worst one travels a path of length 0.45 in its last 500 steps for a net displacement of 0.0024, about 1.0 away from the target in logit units; after step 500, 82\% of its gradients make an angle of more than $90^\circ$ with the true gradient (median cosine $-0.999$).
Its reported loss jumps between about $+3$ and $-1.5$ while the true loss stays at 0.73, so the loss curve cannot be used for monitoring either; in three of the four combinations of precision and direction, about a third of the official training steps report a negative KL.

\paragraph{Limits.} One or two scalar parameters, no model and no data noise, one distribution family, CPU arithmetic.

\subsection{Sequential Bayesian Design}

\paragraph{Setup.} Bayesian experimental design chooses the next observation by its expected information gain, the expected KL between the posterior after the observation and the current one \citep{lindley1956,chaloner1995}.
We take $K=10$ Bernoulli arms with unknown rates and Beta posteriors; an arm with posterior $\mathrm{Beta}(a,b)$ has
\begin{align*}
\mathrm{EIG}(a,b)={}&\tfrac{a}{a+b}\,\KL\bigl(\mathrm{Beta}(a{+}1,b)\,\|\,\mathrm{Beta}(a,b)\bigr)\\
&+\tfrac{b}{a+b}\,\KL\bigl(\mathrm{Beta}(a,b{+}1)\,\|\,\mathrm{Beta}(a,b)\bigr),
\end{align*}
and in every round the arm with the largest EIG receives the next $m$ observations, for a total of 100{,}000.
The KL is computed with the official FP32 \texttt{kl\_divergence(Beta, Beta)} or with our rewrite, also in FP32; the reference policy uses the exact EIG.
We run batch sizes $m\in\{1,100,1000,10^4\}$ from a cold start (no observations) and from a warm start (each arm already has $10^2$ to $10^5$ observations), with 5 seeds and the same outcomes for all policies.

\paragraph{Why this setting.} The information gain of one more observation shrinks like the reciprocal of the number of observations already made, so the experiment itself drives the KL into the range where the closed form fails: for $\mathrm{Beta}(n{+}1,n)$ against $\mathrm{Beta}(n,n)$ the true KL is $2.5\times10^{-4}$, $2.5\times10^{-5}$ and $2.5\times10^{-6}$ for $n=10^3,10^4,10^5$, and the official FP32 code returns $-1.5\times10^{-3}$, $-5.6\times10^{-3}$ and $+0.31$.

\paragraph{Summary of the result.} With the official KL, 40--91\% of the rounds choose a wrong arm in seven of the eight settings (\cref{tab:exp2beta_full}), and 35--60\% of the EIG values it computes there are zero or negative.
With the rewrite no round is wrong in any of the 40 runs: its EIG is within a relative $4.4\times10^{-7}$ of the exact value, and its allocation of observations is that of the exact policy (identical in 39 runs, one observation apart in the other).
The consequence for the experiment depends on the feedback.
When the posterior is updated after every observation ($m=1$), a wrong choice costs one observation, and the total posterior variance at the end is only 1--10\% higher than with the exact policy.
When a choice commits a batch, the loss can be much larger: for $m=1000$ the worst arm ends with up to 2.5 times the variance, and for $m=10^4$ from a warm start the total variance is 1.2 to 5.2 times and the worst arm up to 11 times that of the exact policy, because an arm with few observations is passed over in most rounds.
From a cold start with $m=10^4$ every round goes to an arm without observations, where the official value is accurate, and nothing changes.

\paragraph{Protocol.} Arm rates are drawn uniformly from $[0.05,0.95]$; the prior is $\mathrm{Beta}(1,1)$, so an arm with $s$ successes and $f$ failures has the posterior $\mathrm{Beta}(1+s,1+f)$.
In the warm start arm $k$ has $\mathrm{round}(10^{y_k})$ earlier observations with $y_k$ drawn uniformly from $[2,5]$.
The outcome of the $j$-th observation of an arm is the same for all policies, and ties are broken by the lowest index.
The policies differ only in the two KL values of the EIG: \texttt{torch.distributions.kl\_divergence(Beta, Beta)} in FP32 (official), the rewrite of \cref{app:ifaceA} evaluated in FP32 (ours), and the identity $\KL(\mathrm{Beta}(a{+}1,b)\,\|\,\mathrm{Beta}(a,b))=\log\frac{a+b}{a}+\psi(a+1)-\psi(a+b+1)$, with the digamma function $\psi$, in FP64 (exact), which agrees with 60-digit arithmetic to $1.1\times10^{-8}$ on posteriors with up to $5\times10^{6}$ observations.
The weights $a/(a+b)$ and $b/(a+b)$ and the combination are computed in FP32 for the official code and for ours.
A round counts as wrong if the exact EIG of the chosen arm is below $(1-10^{-3})$ times the largest exact EIG.
The posterior variance of an arm rate is $ab/((a+b)^2(a+b+1))$, the expected squared error of its estimate.

\begin{table}[t]
\caption{Sequential Bayesian design, all settings; 100{,}000 observations, $K=10$, ranges over 5 seeds. ``Lost'': mean over the rounds of $1-\mathrm{EIG}_{\mathrm{chosen}}/\mathrm{EIG}_{\max}$, with exact values at the policy's own posterior. ``Zero or negative'': share of the EIG values computed by the official code. ``Worst arm'': largest posterior variance of an arm, relative to the largest of the exact policy. ``Fewest'': smallest number of observations of an arm at the end over the seeds, official code (exact policy). The rewrite has no wrong round and variance ratios of 1.0000 in all 40 runs; its allocation of the observations is identical to that of the exact policy in 39 runs and differs by one observation in one (cold start, $m=1$).}
\label{tab:exp2beta_full}
\centering\footnotesize
\setlength{\tabcolsep}{2.9pt}
\begin{tabular}{lrrccccccr}
\toprule
 & & & \multicolumn{4}{c}{official: decisions} & \multicolumn{2}{c}{official: variance ($\times$ exact)} & \\
\cmidrule(lr){4-7}\cmidrule(lr){8-9}
start & $m$ & rounds & wrong (\%) & first wrong round & lost (\%) & zero or negative (\%) & sum & worst arm & fewest\\
\midrule
cold & 1 & 100{,}000 & 91 & 380--422 & 20--22 & 41--44 & 1.01--1.05 & 1.08--1.17 & 8{,}308 (9{,}998)\\
 & 100 & 1{,}000 & 84--88 & 11--14 & 25--31 & 43--46 & 1.02--1.08 & 1.12--1.79 & 5{,}600 (10{,}000)\\
 & 1{,}000 & 100 & 63--75 & 13--15 & 28--45 & 36--48 & 1.06--1.27 & 1.20--2.50 & 4{,}000 (10{,}000)\\
 & 10{,}000 & 10 & 0 & -- & 0 & 15--30 & 1.00 & 1.00 & 10{,}000 (10{,}000)\\
\addlinespace
warm & 1 & 100{,}000 & 84--91 & 1--97 & 20--24 & 43--50 & 1.01--1.10 & 0.98--1.67 & 8{,}749 (14{,}836)\\
 & 100 & 1{,}000 & 84--90 & 1--5 & 22--31 & 44--52 & 1.03--1.11 & 1.09--1.61 & 8{,}901 (14{,}802)\\
 & 1{,}000 & 100 & 83--85 & 1--4 & 36--47 & 40--52 & 1.06--1.26 & 1.15--1.85 & 8{,}206 (14{,}271)\\
 & 10{,}000 & 10 & 40--70 & 1--3 & 31--58 & 35--60 & 1.17--5.16 & 1.12--10.96 & 576 (10{,}291)\\
\bottomrule\relax  
\end{tabular}
\end{table}

\paragraph{Results.} \Cref{tab:exp2beta_full} lists all settings.
From a cold start with $m=1$ the first wrong choice occurs after about 380 observations, that is, when each arm has about 38; from then on the share of wrong rounds grows, and the observations per arm at the end range from 8{,}308 to 16{,}176 instead of 9{,}998 to 10{,}001.
The EIG computed by the official code is off by factors of up to $3\times10^{4}$; that of the rewrite by at most $4.4\times10^{-7}$ in relative terms.
Evaluating the rewrite in FP64 and rounding once (strict mode) gives the same choices as the FP32 rewrite in the $m=1$ runs.
The outcome is sensitive to which rounds go wrong: for the warm start with $m=10^4$ the ratio of the total variance is 1.17, 1.19, 1.22, 2.55 and 5.16 in the five seeds, and in the last one an arm that starts with 576 observations receives none of the ten batches, while the exact policy gives it 10{,}000.

\paragraph{Limits.} Simulated Bernoulli arms, one acquisition rule (the arm with the largest one-step EIG), $K=10$, five seeds.
For $m=1$ the effect on the final variance is small, and for the cold start with $m=10^4$ the official code makes no wrong choice.
Only the Beta closed form is exercised.

\subsection{Identical-Start Distillation}

\paragraph{Protocol.} Teacher and student: MLP $32\to256\to V$ with identical weights (seeded); inputs: fixed random batches; loss: forward KL of the output distributions, mean over the batch; 300 steps; metrics at the last step: top-1 disagreement with the teacher, exact KL (FP64), relative weight change. Grid: $V\in\{1024,32768,151936\}\times\{$FP32, FP16, BF16$\}\times\{$Adam $10^{-4}$, $3{\times}10^{-4}$, $10^{-3}$, $3{\times}10^{-3}$, $10^{-2}$; SGD $0.01$, $0.1$, $1\}\times4$ losses $\times10$ seeds $=2{,}880$ runs on one A100 GPU (PyTorch 2.4.0). Adam uses the default $\beta=(0.9,0.999)$ and \texttt{eps}$=10^{-8}$. Weights, optimiser state and the forward pass of both networks are FP32; the logits are cast to the stated format, in which the loss and the start of the backward pass are computed, without autocast or gradient scaling. All runs completed; none was non-finite; 740 runs repeated in an independent execution are bit-identical.

\begin{figure}[h]
\centering
\includegraphics[width=0.9\textwidth]{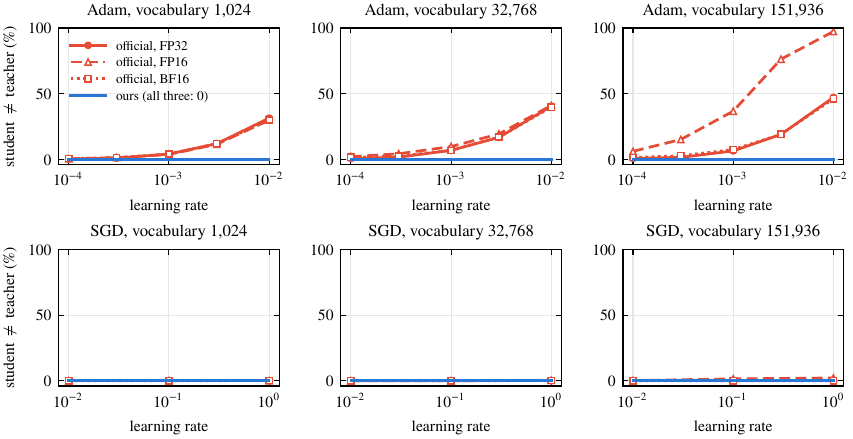}
\caption{Identical-start distillation. Top-1 disagreement between student and teacher after 300 steps as a function of the learning rate (median over 10 seeds). The official \fkl{} is shown in red, with line style and marker indicating the precision, and ours in blue (identically zero in every cell).}
\label{fig:exp2}
\end{figure}

\paragraph{Complete results.} \Cref{fig:exp2} plots \cref{tab:exp2}. Under Adam all 27 curves (3 vocabularies $\times$ 3 precisions $\times$ 3 official variants) increase strictly with the learning rate. In the 45 Adam comparisons between $V=1024$ and $V=151{,}936$ the larger vocabulary is always worse; between $V=32{,}768$ and $151{,}936$ three comparisons are reversed (two FP32, one BF16), within the seed range. Under SGD the largest median is 2.1\% ($V=151{,}936$, FP16, learning rate 1). 24.4\% of the official runs report a negative loss at step 0 (e.g.\ median $-1.7\times10^{-8}$ for $V=151{,}936$, FP32). The median initial gradient norm of the official \fkl{} loss is $4$--$5\times10^{-8}$ in FP32, $2\times10^{-4}$ to $1.6\times10^{-2}$ in FP16 and about $1.5\times10^{-3}$ in BF16.

\paragraph{Limits.} Synthetic MLP, identical initialisation, 300 steps. The logits path in BF16 is exploratory (BF16 is not covered by the certified evaluation).

\subsection{KL in Language Models}

\paragraph{Models.} Qwen2.5-1.5B-Instruct, FP16 forward pass; perplexity of the original model 11.66 on the 16,384 selection tokens. Quantized versions and their perplexity: RTN 8-bit per channel 11.68; GPTQ 8-bit 11.67; AWQ 4-bit 12.74; GPTQ 4-bit 13.32; RTN 4-bit (group 128) 13.23; RTN 3-bit (group 128) 31.14. Published GPTQ/AWQ weights were dequantized to FP16 and run through ordinary linear layers. Only the KL computation changes precision.

\begin{table}[h]
\caption{Language models, Part A. Reported mean KL (\texttt{batchmean}) and its deviation from the FP64 reference. ``inf'': the reported value is infinite.}
\centering\small
\setlength{\tabcolsep}{3.4pt}
\begin{tabular}{lcccccc}
\toprule
Model & reference & \fkl{} FP16 & \texttt{Categorical} FP16 & interface-B fix FP16 & \textbf{logits path FP16} & \fkl{} FP32\\
\midrule
GPTQ W8 & $4.4929\times10^{-4}$ & $-19.0\%$ & inf & $+16.5\%$ & \textbf{0.0\%} & 0.0\%\\
RTN W8 & $1.7765\times10^{-3}$ & $-4.4\%$ & inf & $+4.4\%$ & \textbf{0.0\%} & 0.0\%\\
AWQ W4 & 0.10139 & $0.1\%$ & inf & $0.1\%$ & \textbf{0.0\%} & 0.0\%\\
GPTQ W4 & 0.13633 & $0.0\%$ & inf & $0.1\%$ & \textbf{0.0\%} & 0.0\%\\
RTN W4 & 0.15677 & $0.0\%$ & inf & $0.0\%$ & \textbf{0.0\%} & 0.0\%\\
RTN W3 & 0.99191 & $0.0\%$ & inf & $0.0\%$ & \textbf{0.0\%} & 0.0\%\\
\bottomrule
\end{tabular}
\end{table}

\begin{table}[h]
\caption{Language models, Part A. Per-token errors on 16,384 tokens for the two 8-bit models (GPTQ / RTN). ``Total'' error is relative to the exact KL of the FP16 model logits; ``operator'' error is relative to the exact value on the inputs the call received. For the logits path the last column is agreement with the rounded FP64 reference.}
\centering\small
\begin{tabular}{lccccc}
\toprule
Method & negative & infinite & total $\ge1\%$ & total $\ge100\%$ & operator correctly rounded\\
\midrule
\fkl{} FP16 & 30.2 / 12.3\% & 0 & 99.0 / 97.0\% & 49.2 / 21.6\% & 0.1 / 0.2\%\\
\texttt{Categorical} FP16 & 1.3 / 1.8\% & 98.0 / 97.1\% & 100 / 100\% & 100 / 99.9\% & --\\
interface-B fix FP16 & 16.9 / 7.5\% & 0 & 98.8 / 96.8\% & 45.6 / 20.0\% & 100 / 100\%\\
\textbf{logits path FP16} & \textbf{0} & \textbf{0} & 2.0 / 1.1\%$^{\ast}$ & 0.1 / 0\% & \textbf{100 / 100\%}\\
\fkl{} FP32 & 16 / 4 tokens & 0 & 7.5 / 4.3\% & 0.2 / 0\% & --\\
\texttt{Categorical} FP32 & 87 / 40 tokens & 0 & 20.1 / 10.9\% & 1.0 / 0.5\% & --\\
\textbf{logits path FP32} & \textbf{0} & \textbf{0} & \textbf{0} & \textbf{0} & \textbf{100 / 100\%}\\
\bottomrule
\end{tabular}

\smallskip
{\footnotesize $^{\ast}$All of these tokens have exact KL $\le2.8\times10^{-6}$, below the smallest normal FP16 number; they are correctly rounded.}
\end{table}

\paragraph{Part A, further observations.} For the 4- and 3-bit models, whose KL is 0.1--1 nat, all finite methods agree with the reference within 0.1\%. The six models are far apart compared with these errors (the closest pair, GPTQ and RTN at 4 bits, differs by 15\%). The repaired \fkl{} reduces the share of tokens with operator error $\ge1\%$ from 97.6 / 95.5\% to 0.3 / 0.1\%, while 98.8 / 96.8\% of the tokens have input-representation error $\ge1\%$; the logits path has no input-representation error. Casting the FP16 logits to BF16 before computing the KL changes the divergence itself (reported values off by 31--124\% at 8 bits for every method, including the logits path, which is then correctly rounded for the BF16 logits): the KL should be computed from the logits in the precision in which the model produced them. All 15 model pairs are ordered correctly by every finite method except FP16 \texttt{reduction="mean"}, which reports both 8-bit models as 0 (a tie). The references use a cancellation-free formula in FP64, and the logits path reproduces the reference mean to four digits. The official \texttt{Categorical} KL is infinite because it treats an underflowed zero in $Q$ as a true zero.

\begin{table}[h]
\caption{Language models, Part B. Ranking of 28 single-layer quantizations: strict inversions among 378 pairs / reported ties, and Kendall's $\tau_b$ against the reference.}
\centering\small
\begin{tabular}{lcccc}
\toprule
 & \multicolumn{2}{c}{8-bit per channel} & \multicolumn{2}{c}{4-bit, group 128}\\
\cmidrule(lr){2-3}\cmidrule(lr){4-5}
Method & inversions / ties & $\tau_b$ & inversions / ties & $\tau_b$\\
\midrule
\fkl{} FP16 \texttt{batchmean} & 19 / 1 & 0.898 & 1 / 0 & 0.995\\
\fkl{} FP16 \texttt{mean} & 0 / 378 & -- & 11 / 32 & 0.896\\
\fkl{} \texttt{log\_target} FP16 & 35 / 0 & 0.815 & 2 / 0 & 0.989\\
\texttt{Categorical} FP16 & \multicolumn{2}{c}{not rankable (inf)} & \multicolumn{2}{c}{not rankable (inf)}\\
interface-B fix FP16 & 20 / 0 & 0.894 & 1 / 0 & 0.995\\
\textbf{logits path FP16} & \textbf{0 / 0} & \textbf{1.000} & \textbf{0 / 0} & \textbf{1.000}\\
\fkl{} BF16 & 41 / 0 & 0.783 & 1 / 0 & 0.995\\
\texttt{Categorical} BF16 & 158 / 0 & 0.164 & 25 / 0 & 0.868\\
logits path BF16 (exploratory) & 23 / 0 & 0.878 & 1 / 0 & 0.995\\
all methods FP32 & 0 / 0 & 1.000 & 0 / 0 & 1.000\\
\bottomrule
\end{tabular}
\end{table}

\paragraph{Part B, decisions.} At 4 bits, where single-layer divergences are two orders of magnitude larger, inversions are rare (1--2 pairs in FP16 with \texttt{batchmean}). A decision counts as harmful if the mixed model has a higher true KL on a disjoint 4,096-token evaluation set. The reference protects layers $\{0,1,2,3,5,19,26,27\}$ for 8 bits and $k=8$. Official \texttt{log\_target} FP16, the official \fkl{} forms in BF16 and the BF16 logits path select layer 25 instead of layer 5; the reference KL of layers 5, 23 and 25 is $(5.787,\,5.759,\,5.474)\times10^{-5}$, and the resulting mixed model has $0.25\%$ \emph{lower} true KL on the disjoint evaluation set ($+0.87\%$ on the selection set). \texttt{Categorical} BF16 selects $\{1,2,6,9,15,22,26,27\}$ (four wrong layers): $+9.78\%$ true KL on the evaluation set, $+15.31\%$ on the selection set, perplexity difference $-0.009$. All rankable methods agree with the reference for 8 bits with $k=4$ and for 4 bits with $k\in\{4,8\}$. The reported per-layer mean of the official FP16 \fkl{} deviates from the reference by $-233\%$ to $-32\%$ at 8 bits (24 layers negative) and by $-3.5\%$ to $-0.3\%$ at 4 bits; the logits path by $\pm0.02\%$.

\paragraph{Limits.} One model, one dataset, round-to-nearest quantizers, two values of $k$; references are FP64 (not interval-certified), spot-checked at 150 digits. The most common official configuration, FP16 \fkl{} with \texttt{batchmean}, still selects the correct layers, and perplexity differences are within $\pm0.01$ throughout.

\paragraph{Part C, protocol.} Qwen2.5-1.5B-Instruct with the default (\texttt{sdpa}) attention implementation of \texttt{transformers}; WikiText-2 training text; nothing is trained.
\emph{Executions.} For each of FP32, FP16 and BF16 the same weights are run in four ways: one full forward pass with batch size 8 (the reference execution), batch size 1, token-by-token decoding with a key--value cache (what a rollout engine does), and the \texttt{eager} attention implementation; $8\times256=2{,}048$ positions.
Each alternative is compared with the reference execution of the same precision, and the three reference executions with each other.
The official calls are evaluated in the precision of the model (in FP32 for pairs of different precisions), 128 rows per call, \texttt{batchmean}; the sampled estimators use tokens drawn from the first execution and FP32 log-probabilities, as the RL fine-tuning frameworks we inspected do (\texttt{k3\_kl} of verl, the KL estimators of OpenRLHF).
\emph{Reproducibility.} The logits of the model and of five round-to-nearest copies (8, 7, 6 and 5 bits per output channel, 4 bits with group size 128) are computed once on $8\times512=4{,}096$ positions in FP16 and in BF16.
The mean KL is then computed under settings that cannot change its value: 1, 16, 128, 512 or all rows per call; five reductions (\texttt{batchmean}, \texttt{sum} divided by the rows, \texttt{mean} times $V$, \texttt{none} summed in the same precision or in FP32); two random row orders; CPU instead of GPU.

\begin{table}[h]
\caption{Language models, Part~C: relative error (\%) of the mean KL between two executions of the same weights. Rows with one precision: the named execution against the full forward pass with batch size 8, KL computed in that precision; rows with two precisions: KL computed in FP32. Last column: sampled estimator $e^{x}-x-1$ / $\emo(x)-x$, with $x$ the log-probability ratio of the two executions at a sampled token, in FP32, relative to the same estimator in FP64.}
\label{tab:exp3c_pairs}
\centering\footnotesize
\setlength{\tabcolsep}{3pt}
\resizebox{\textwidth}{!}{%
\begin{tabular}{llccccccc}
\toprule
precision & execution & true KL & top-1 differs (\%) & \fkl & \texttt{log\_target} & \texttt{Categorical} & logits path & sampled: official / cancellation-free\\
\midrule
FP32 & cached decoding & $8.1\times10^{-11}$ & 0.00 & $-2304$ & $+1967$ & $-1413$ & $0.0$ & $-1761$ / $0.0$\\
FP16 & cached decoding & $1.8\times10^{-5}$ & 0.34 & $-684$ & $-17$ & non-finite & $0.0$ & $0.0$ / $0.0$\\
BF16 & cached decoding & $1.1\times10^{-3}$ & 1.90 & $-19$ & $-2.1$ & $+79$ & $0.0$ & $0.0$ / $0.0$\\
FP32 & batch size 1 & $5.8\times10^{-11}$ & 0.00 & $+2928$ & $+4485$ & $-4000$ & $0.0$ & $-1827$ / $0.0$\\
FP16 & batch size 1 & $1.8\times10^{-5}$ & 0.29 & $-691$ & $-30$ & non-finite & $0.0$ & $0.0$ / $0.0$\\
BF16 & batch size 1 & $1.1\times10^{-3}$ & 2.00 & $-12$ & $+3.6$ & $-31$ & $0.0$ & $0.0$ / $0.0$\\
FP32 & eager attention & $6.1\times10^{-11}$ & 0.00 & $-4397$ & $-3077$ & $-292$ & $0.0$ & $-3032$ / $0.0$\\
FP16 & eager attention & \multicolumn{7}{l}{skipped: the FP16 forward pass with eager attention is non-finite at all positions}\\
BF16 & eager attention & $3.9\times10^{-2}$ & 8.01 & $-0.3$ & $-0.2$ & $-1.1$ & $0.0$ & $0.0$ / $0.0$\\
FP32 vs.\ FP16 & full forward (both) & $2.7\times10^{-5}$ & 0.44 & $0.0$ & $0.0$ & $+0.1$ & $0.0$ & $0.0$ / $0.0$\\
FP32 vs.\ BF16 & full forward (both) & $1.0\times10^{-3}$ & 2.93 & $0.0$ & $0.0$ & $0.0$ & $0.0$ & $0.0$ / $0.0$\\
FP16 vs.\ BF16 & full forward (both) & $1.1\times10^{-3}$ & 2.59 & $0.0$ & $0.0$ & $0.0$ & $0.0$ & $0.0$ / $0.0$\\
\bottomrule\relax  
\end{tabular}}
\end{table}

\begin{table}[h]
\caption{Language models, Part~C: reported KL as a percentage of the truth, FP16, official \fkl{} with a probability target. Columns 3--7: \texttt{reduction="mean"} for different numbers of rows per call; column 8: range over all 27 settings with another reduction; column 9: range of the logits path over its 12 settings.}
\label{tab:exp3c_batch}
\centering\footnotesize
\setlength{\tabcolsep}{4pt}
\begin{tabular}{lcccccccc}
\toprule
 & & \multicolumn{5}{c}{\texttt{mean}, rows per call} & other & logits\\
\cmidrule(lr){3-7}
copy & true KL & 1 & 16 & 128 & 512 & 4,096 & reductions & path\\
\midrule
8 bits & $1.9\times10^{-3}$ & 57 & 49 & 58 & 0 & 0 & 94.7--95.7 & 99.98--100.01\\
7 bits & $5.7\times10^{-3}$ & 82 & 76 & 60 & 140 & 160 & 97.7--98.2 & 99.99--100.03\\
6 bits & $3.1\times10^{-2}$ & 99 & 99 & 101 & 91 & 87 & 99.6--99.7 & 99.99--100.02\\
5 bits & $8.2\times10^{-2}$ & 100 & 100 & 99 & 99 & 99 & 99.9--100.0 & 99.99--100.01\\
4 bits (g128) & $1.5\times10^{-1}$ & 100 & 100 & 101 & 100 & 98 & 99.9--100.0 & 99.97--100.01\\
\bottomrule\relax  
\end{tabular}
\end{table}

\paragraph{Part C, results.} \Cref{tab:exp3c_pairs} lists all pairs.
The gap between two executions grows with the coarseness of the precision (about $10^{-10}$, $2\times10^{-5}$ and $10^{-3}$), and in each precision the official KL of that precision cannot measure it; computed one precision higher, the official calls agree with the reference (within 0.1\% in FP32 for the FP16 and BF16 pairs, within $10^{-6}$ in FP64 for all pairs).
In BF16 the most likely token differs at 1.9\% of the positions between cached decoding and the full forward pass.
\Cref{tab:exp3c_batch} shows the dependence on the batch size: with \texttt{mean} the result is of the order of $\KL/V$, which is subnormal or zero in FP16, so that the reported value depends on how many rows are averaged at once; the same underflow makes FP16 \texttt{mean} report zero for all layers in Part~B.
With the other reductions the official value is stable over batch sizes, row orders and devices, with a bias of up to 5\% for the 8-bit copy; \texttt{log\_target} behaves like the probability target; the FP16 \texttt{Categorical} KL is non-finite in all 17 settings.
With a BF16 forward pass the official values vary by at most 3.6\% and are off by at most 6.2\%.
The logits path varies by at most 0.41\% and is off by at most 0.33\% in every setting of both precisions.
The order of the five copies is correct in every setting that returns finite values.

\paragraph{Part C, limits.} One model, one text, a single run.
The executions are variants within one framework, not a rollout engine against a training framework.
The FP32 gap of about $10^{-10}$ has no practical consequence.
The dependence on the batch size is confined to FP16 with \texttt{reduction="mean"}.
The FP16 forward pass with \texttt{eager} attention is non-finite at every position for this model, so that pair is omitted.

\paragraph{Part D, protocol.} \texttt{auto-circuit} 1.0.1 \citep{miller2024}, unmodified, with \texttt{transformer-lens} 2.4.0; models without weight processing; the prompts shipped with the library; first batch, last token position; factorised graph with separate query, key and value inputs.
The settings are IOI \citep{wang2023ioi} on GPT-2 small (batch of 64, 32{,}491 edges) and on GPT-2 medium (batch of 64; 29{,}999 of 231{,}877 edges, sampled at random with the same fraction from each of the 97 destination modules), Greater-than \citep{hanna2023} on GPT-2 small (batch of 64, 32{,}491 edges), and Docstring \citep{heimersheim2023} on the 4-layer attention-only model of that study (batch of 128, 1{,}281 edges, $V=48{,}262$).
The loop is the library's ACDC loop \citep{conmy2023} (same ablation tensors, patch mode, forward call and output slice) with one change: every edge is restored after it has been scored, so that the scores are independent of each other.
For each edge we record, on the same output logits, the library's score (\texttt{F.kl\_div} with \texttt{log\_target}, \texttt{reduction="sum"}, divided by the number of rows, in the precision of the model), the logits path, and as the reference the KL of the two logit tensors evaluated in FP64.
A score counts as wrong if it is negative or deviates from the reference by more than 10\%.
BF16 was run for the first setting only. The edge counts in \cref{tab:exp3d_main,tab:exp3d} are the edges with a positive reference score (32{,}477, 32{,}477, 1{,}272 and 29{,}999 of the 32{,}491, 32{,}491, 1{,}281 and 29{,}999 edges; 32{,}476 for Greater-than in FP16); the others have a reference of exactly zero.

\begin{table}[t]
\caption{Language models, Part~D, all settings. ``Wrong'': negative or more than 10\% from the FP64 KL of the same logits, among the edges with a positive reference. ``Selected differently'': edges whose decision $\text{score}\ge\tau$ differs from the reference's. Spearman and top-$k$ overlap are with the reference ranking of the same model; the Docstring model has fewer than 5{,}000 edges.}
\label{tab:exp3d}
\centering\footnotesize
\setlength{\tabcolsep}{3.5pt}
\resizebox{\textwidth}{!}{%
\begin{tabular}{lllrrrrrrrr}
\toprule
 & & & & & \multicolumn{3}{c}{selected differently} & & & \\
\cmidrule(lr){6-8}
task, model & precision & score & wrong & negative & $\tau=10^{-6}$ & $\tau=10^{-4}$ & $\tau=10^{-3}$ & Spearman & top 1{,}000 & top 5{,}000\\
\midrule
IOI, GPT-2 small & FP32 & library & 19.3\% & 1{,}449 & 84 & 0 & 0 & 0.9992 & 1.000 & 1.000\\
 &  & \textbf{logits path} & \textbf{0} & \textbf{0} & \textbf{0} & \textbf{0} & \textbf{0} & \textbf{1.0000} & \textbf{1.000} & \textbf{1.000}\\
 & FP16 & library & 81.4\% & 3 & 3 & 587 & 29 & 0.7447 & 0.963 & 0.637\\
 &  & \textbf{logits path} & \textbf{0} & \textbf{0} & \textbf{0} & \textbf{1} & \textbf{0} & \textbf{1.0000} & \textbf{1.000} & \textbf{0.998}\\
 & BF16 & library & 2.5\% & 3 & 5 & 9 & 20 & 0.9922 & 0.849 & 0.888\\
 &  & \textbf{logits path} & \textbf{0} & \textbf{0} & \textbf{0} & \textbf{0} & \textbf{1} & \textbf{0.9999} & \textbf{0.976} & \textbf{0.998}\\
\addlinespace
Greater-than, GPT-2 small & FP32 & library & 45.8\% & 3{,}271 & 25 & 0 & 0 & 0.9683 & 1.000 & 1.000\\
 &  & \textbf{logits path} & \textbf{0} & \textbf{0} & \textbf{0} & \textbf{0} & \textbf{0} & \textbf{1.0000} & \textbf{1.000} & \textbf{1.000}\\
 & FP16 & library & 6.8\% & 2 & 1 & 7 & 1{,}150 & 0.9496 & 0.698 & 0.733\\
 &  & \textbf{logits path} & \textbf{0.02\%} & \textbf{0} & \textbf{0} & \textbf{0} & \textbf{0} & \textbf{1.0000} & \textbf{0.999} & \textbf{0.998}\\
\addlinespace
Docstring, attn-only-4l & FP32 & library & 2.4\% & 0 & 0 & 0 & 0 & 1.0000 & 1.000 & --\\
 &  & \textbf{logits path} & \textbf{0} & \textbf{0} & \textbf{0} & \textbf{0} & \textbf{0} & \textbf{1.0000} & \textbf{1.000} & \textbf{--}\\
 & FP16 & library & 54.2\% & 10 & 12 & 107 & 4 & 0.9839 & 0.959 & --\\
 &  & \textbf{logits path} & \textbf{0} & \textbf{0} & \textbf{0} & \textbf{0} & \textbf{0} & \textbf{1.0000} & \textbf{1.000} & \textbf{--}\\
\addlinespace
IOI, GPT-2 medium & FP32 & library & 58.8\% & 4{,}401 & 23 & 0 & 0 & 0.9371 & 1.000 & 0.995\\
 &  & \textbf{logits path} & \textbf{0} & \textbf{0} & \textbf{0} & \textbf{0} & \textbf{0} & \textbf{1.0000} & \textbf{1.000} & \textbf{1.000}\\
 & FP16 & library & 81.1\% & 338 & 354 & 9{,}829 & 0 & 0.4735 & 0.375 & 0.356\\
 &  & \textbf{logits path} & \textbf{0.01\%} & \textbf{0} & \textbf{0} & \textbf{3} & \textbf{0} & \textbf{1.0000} & \textbf{0.999} & \textbf{0.999}\\
\bottomrule\relax  
\end{tabular}}
\end{table}

\paragraph{Part D, results.} \Cref{tab:exp3d} lists all settings.
\emph{FP32.} The errors are confined to small scores. On GPT-2 small with IOI the median reference score is $8.8\times10^{-7}$, and the library's score is wrong for 73\% of the edges between $10^{-8}$ and $10^{-7}$, for at least 97\% below $10^{-8}$, for 9\% between $10^{-7}$ and $10^{-6}$ and for none above; the other GPT-2 settings show the same profile with smaller medians ($8.8\times10^{-8}$ for Greater-than, $3.0\times10^{-8}$ on GPT-2 medium) and correspondingly more wrong scores.
With the library's thresholds $\tau=10^{-6},10^{-5},10^{-4},10^{-3}$ the numbers of edges selected differently are 84, 7, 0, 0 (IOI, small), 25, 1, 0, 0 (Greater-than) and 23, 1, 0, 0 (IOI, medium), and the 1{,}000 highest-scoring edges are the same in every setting.
The relative error of the logits path has a median of about $2\times10^{-8}$ and a 95th percentile of at most $3.6\times10^{-7}$.
On the Docstring model the median score is $1.9\times10^{-4}$ and only 30 of 1{,}272 scores are wrong.
\emph{FP16.} The median relative error of the library's score is 24\% (IOI, small), 3\% (Greater-than), 12\% (Docstring) and 28\% (IOI, medium).
On GPT-2 medium the threshold $\tau=10^{-4}$ selects 17{,}192 edges by the reference and 9{,}829 edges are decided differently by the library's score, 3 by the logits path.
On GPT-2 small with IOI the library's score is wrong for 341 of the 739 edges of the published circuit \citep{wang2023ioi}.
The logits path is wrong for 5 and 2 edges in two settings; these have a reference below $10^{-7}$, where FP16 has at most one significant bit (its smallest positive number is $6\times10^{-8}$), and the same value rounded to FP32 instead is wrong for none.
\emph{BF16} (IOI, small). The library's score is wrong for 2.5\% of the edges.

\paragraph{Part D, limits.} One batch and a single run per setting; GPT-2 medium is a 13\% sample of its edges.
In FP32, threshold decisions change, but the top 1,000 edges are unchanged.
That the larger model is affected more rests on one pair of models on one task.
Half-precision scores on the GPT-2 models are mostly noise of the forward pass (median $2.1\times10^{-4}$ in FP16 against $8.8\times10^{-7}$ in FP32 for IOI on GPT-2 small): the logits path recovers the score of the half-precision model, not that of the FP32 model, whose ranking the FP16 reference itself reproduces only with a Spearman correlation of 0.63 (IOI, small), 0.63 (Greater-than) and 0.39 (IOI, medium); for the Docstring model it is 0.999.
With the library's sequential ACDC (three KL computations, GPT-2 small, IOI) the logits path did not bring the half-precision circuits measurably closer to the FP32 circuit (at most 3.3\% more edges in agreement), because each decision changes all later ones, so that two accurate KL computations differ from each other by as many edges as either differs from the library's, and because in BF16 the forward pass, not the KL, decides the circuit (over 99.9\% of the edges are kept for $\tau\le10^{-4}$ with every KL computation).
\section{Use of the Interfaces, User Reports, and Settings Without a Measurable Effect}
\label{app:usage}

This appendix collects material that bounds the practical reach of the errors from both sides: where the two interfaces are called (\cref{app:callsites}), earlier reports of the same failure by users (\cref{app:reports}), and the settings we tested in which replacing the official KL changed nothing measurable (\cref{app:null}).
The experiments of \cref{app:null} are exploratory.

\subsection{Call sites in PyTorch-based libraries}
\label{app:callsites}

We downloaded the current release of 116 PyTorch-based packages from PyPI on 14 November 2026 and searched their Python sources, excluding tests, documentation and examples, for calls of \texttt{torch.distributions.kl\_divergence} (interface~A) and of \texttt{F.kl\_div} or \texttt{nn.KLDivLoss} (interface~B).
A call was counted only if the name resolves to PyTorch in that file; functions of the same name that a package defines for itself were not counted.
Interface~A is referenced in 20 packages and interface~B in 22; 39 packages use at least one (\cref{tab:callsites}).
The packages were chosen by us from reinforcement learning, probabilistic modelling, distillation, speech, compression, interpretability, robustness and uncertainty estimation.
Whether a call is affected depends on how close its two distributions are and on the precision.
The call sites whose two arguments are close by construction are those that compare a policy before and after an update (\texttt{cleanrl}, \texttt{garage}, \texttt{pfrl}, \texttt{rsl-rl-lib}, \texttt{skrl}, \texttt{stable-baselines3} with \texttt{sb3-contrib}, \texttt{tianshou}, \texttt{torchrl}), a student with its teacher, or a model with an ablated or perturbed copy of itself (\texttt{auto-circuit}, \texttt{sae-lens}, \texttt{inseq}, \texttt{easyeditor}, the adversarial-training code of \texttt{adversarial-robustness-toolbox} and \texttt{torchattacks}).
The metric module of \texttt{auto-circuit} clamps its KL at zero with the comment that numerical errors can cause tiny negative values.

\begin{table}[t]
\caption{Packages (of 116 inspected) whose non-test code calls interface~A (\texttt{kl\_divergence}) or interface~B (\texttt{F.kl\_div}, \texttt{KLDivLoss}): number of source files with at least one call. The use is our reading of the calling files. \texttt{sb3-contrib} 2.9.2 (TRPO) calls interface~A through the helper of \texttt{stable-baselines3} and is not counted separately.}
\label{tab:callsites}
\centering\footnotesize
\setlength{\tabcolsep}{4pt}
\resizebox{\textwidth}{!}{%
\begin{tabular}{llccl@{\hspace{14pt}}llccl}
\toprule
package & version & A & B & use & package & version & A & B & use\\
\midrule
\texttt{adv.-robustness-toolbox} & 1.20.1 & -- & 4 & TRADES-type training & \texttt{pfrl} & 0.4.0 & 2 & -- & TRPO, ACER\\
\texttt{allennlp} & 2.10.1 & 1 & -- & fairness metrics & \texttt{pyro-ppl} & 1.9.2 & 3 & -- & analytic ELBO terms\\
\texttt{audiocraft} & 1.3.0 & -- & 1 & audio KLD metric & \texttt{rsl-rl-lib} & 5.5.2 & 1 & -- & PPO adaptive learning rate\\
\texttt{auto-circuit} & 1.0.1 & -- & 4 & circuit discovery & \texttt{sae-lens} & 6.55.5 & -- & 1 & SAE evaluation\\
\texttt{botorch} & 0.18.1 & 2 & -- & information gain, neural process & \texttt{scvi-tools} & 1.5.2 & 21 & 1 & VAE terms\\
\texttt{cleanrl} & 0.4.8 & 4 & -- & PPG: old vs.\ new policy & \texttt{sentence-transformers} & 6.3.0 & -- & 5 & distillation losses\\
\texttt{coqui-tts} & 0.28.0 & 1 & -- & VAE loss (speech) & \texttt{skrl} & 2.1.1 & 1 & -- & KL-adaptive learning rate\\
\texttt{d3rlpy} & 2.8.1 & 1 & -- & VAE imitator & \texttt{speechbrain} & 1.1.2 & 1 & 1 & SEGAN; distillation loss\\
\texttt{disent} & 0.9.3 & 3 & -- & VAE regulariser & \texttt{stable-baselines3} & 2.9.2 & 1 & -- & policy KL helper\\
\texttt{easyeditor} & 0.0.1.dev0 & -- & 2 & model editing & \texttt{tianshou} & 2.0.1 & 2 & -- & TRPO, NPG\\
\texttt{espnet} & 202611 & 1 & 3 & VITS; label smoothing & \texttt{timm} & 1.0.31 & -- & 3 & distillation, JSD loss\\
\texttt{fairseq} & 0.12.2 & -- & 1 & non-autoregressive loss & \texttt{torch-geometric} & 2.8.1 & -- & 2 & pooling; faithfulness\\
\texttt{garage} & 2021.3.0 & 2 & -- & policy KL constraint & \texttt{torch-uncertainty} & 0.13.0 & -- & 2 & classification routines\\
\texttt{gpytorch} & 1.15.2 & 3 & -- & variational GP & \texttt{torchattacks} & 3.5.1 & -- & 1 & TPGD attack\\
\texttt{inseq} & 0.7.1 & -- & 1 & attribution & \texttt{torchdistill} & 1.1.5 & -- & 1 & distillation\\
\texttt{kornia} & 0.9.1 & -- & 1 & divergence losses & \texttt{torchrl} & 0.15.0 & 1 & -- & KL-penalty PPO\\
\texttt{lightly} & 1.6.1 & -- & 2 & self-supervised losses & \texttt{transformers} & 5.23.0 & -- & 2 & VisualBERT, D-FINE\\
\texttt{lightning-bolts} & 0.7.0 & 1 & -- & VAE & \texttt{trl} & 1.18.0 & -- & 6 & distillation trainers\\
\texttt{lightning-uq-box} & 0.4.0 & 1 & -- & probabilistic U-Net & \texttt{vec2text} & 0.0.13 & -- & 1 & inversion from logits\\
\texttt{nemo-toolkit} & 3.0.0 & -- & 1 & TTS distillation &  & & & & \\
\bottomrule\relax  
\end{tabular}}
\end{table}

\subsection{Earlier user reports}
\label{app:reports}

The failure for nearly identical distributions has been reported repeatedly by users.
We found five reports between 2019 and 2023 by searching the PyTorch issue tracker and forum.
\begin{enumerate}[leftmargin=*,itemsep=1pt,topsep=2pt]
\item Forum thread 48470 (2019): the \texttt{Categorical} KL of two close distributions is negative ($-1.65\times10^{-8}$), and the \texttt{probs} and \texttt{logits} parameterisations disagree. \url{https://discuss.pytorch.org/t/48470}
\item Issue 32520 (2020): \texttt{F.kl\_div} is negative, or non-zero where it should be zero. It was closed when the \texttt{log\_target} flag was added in version 1.6. \url{https://github.com/pytorch/pytorch/issues/32520}
\item Issue 80488 (2022): \texttt{F.kl\_div} is still negative after that change. The answer was that this is numerical error, that removing it in the core would cost performance, and that the result can be clamped. \url{https://github.com/pytorch/pytorch/issues/80488}
\item Forum thread 149312 (2022): under automatic mixed precision the \texttt{Categorical} KL is negative in more than 40\% of the cases (under 1\% without); in the example the true value is $4.1\times10^{-6}$, the FP16 \texttt{Categorical} KL $-8.79\times10^{-4}$ and the FP16 \texttt{F.kl\_div} $-1.52\times10^{-4}$. \url{https://discuss.pytorch.org/t/149312}
\item Forum thread 166383 (2023): \texttt{F.kl\_div} is negative for nearly equal outputs; the answer was that this is expected floating-point error and that the result can be clamped. \url{https://discuss.pytorch.org/t/166383}
\end{enumerate}
Clamping removes negative outputs but cannot recover the true divergence or its gradient. It also leaves inaccurate positive values unchanged.
Reports of a different kind (overflow for extreme parameters, masked classes, logic errors) are not listed.

\subsection{Settings in which we measured no effect}
\label{app:null}

The experiments of \cref{sec:exp} show where the errors change an outcome.
In none of the settings below did the choice of KL implementation change the outcome beyond the run-to-run spread (\cref{tab:null}).

\begin{table}[t]
\caption{Settings without a measurable effect of the KL implementation. ``Official'' is the KL as the cited library computes it; ``ours'' is the logits path. LLM runs use Qwen2.5-1.5B-Instruct with a BF16 forward pass on WikiText-2; ``eval KL'' is the exact (FP64) KL between teacher and student on held-out text.}
\label{tab:null}
\centering\footnotesize
\setlength{\tabcolsep}{4pt}
\begin{tabular}{p{0.19\textwidth}p{0.27\textwidth}p{0.22\textwidth}p{0.24\textwidth}}
\toprule
setting & what was run & expected mechanism & what we measured\\
\midrule
LLM distillation, nearby start & TRL GKD loss; student = 8-bit or 4-bit copy; 320 steps, 10 seeds & the BF16 loss mis-states small KLs by tens of percent and has a spurious gradient at identity & final eval KL: ours vs.\ official $+1.4\%$ and $-0.3\%$ (seed spread 2.8\%, 1.9\%)\\
\addlinespace
LoRA, nearby start or KL anchor & rank 16, two learning rates, 3 seeds; quantized starts; task loss + KL to the base model & same loss in a low-dimensional update & differences below the seed spread, no consistent sign\\
\addlinespace
LoRA, strong constraint & KL only from a perturbed adapter; task loss + $\lambda\,$KL, $\lambda=10,100,1000$; 63 runs & the KL term dominates the objective, and KL $=0$ is reachable & median final eval KL 0.018--0.041 with every path; ours does not return to the start either\\
\addlinespace
PPO adaptive learning rate & \texttt{rsl-rl-lib} 5.5.2, Beta policy, real training, $2\times8{,}000$ decisions & the Beta closed form loses accuracy for large concentrations; a simulation predicted changed decisions & 0 decisions differ from FP64\\
\addlinespace
PPO KL thresholds & synthetic categorical policies, 6--50{,}000 actions, FP32 & the KL of consecutive policies is small & 0 of 960 decisions differ; error $\le2.5\times10^{-5}$\\
\addlinespace
TRPO Fisher-vector product & double backward of the official FP32 KL at the old policy & evaluated exactly at identical distributions & relative error $\le3.3\times10^{-7}$\\
\addlinespace
Sampled KL estimators in RL fine-tuning & $k_1=-x$, $k_3=e^{x}-x-1$ ($x$: log-probability ratio at a sampled token) and full KL on FP32 log-probabilities of an LLM & $k_3$ cancels for small $x$ & batch mean within $7.8\times10^{-6}$ (relative) for KL $10^{-3}$--$1.4\times10^{-2}$\\
\addlinespace
Entropy calibration & threshold search on 4 synthetic histograms, FP32, 24 trials & argmin over many small KLs & same threshold as FP64 in 24 of 24\\
\addlinespace
Sequential ACDC & \texttt{auto-circuit} 1.0.1, GPT-2 small, 3 precisions, 4 thresholds & thresholds of $10^{-6}$--$10^{-3}$ on KL differences & $\le3.3\%$ more edges in agreement with FP32\\
\bottomrule\relax  
\end{tabular}
\end{table}

\paragraph{Distillation of an LLM from a nearby start.}
We copied the loss of the generalised knowledge distillation trainer of TRL \citep{trl} (\texttt{log\_softmax} followed by \texttt{F.kl\_div} with \texttt{log\_target}) and trained a Qwen2.5-1.5B student towards the original model for 320 steps (AdamW with $\beta=(0.9,0.999)$, \texttt{eps}$=10^{-8}$ and no weight decay, learning rate $2\times10^{-5}$, 10 seeds that fix the data order), with BF16 logits as a DeepSpeed launch would produce them, with logits cast to FP32, and with the logits path.
From an \emph{identical} start the result of \cref{sec:exp2} carries over to the LLM: with either official variant the student ends at an eval KL of about 0.10 (0.097--0.103 over the 10 seeds) with 13\% top-1 disagreement, while with the logits path no weight changes in any seed.
From the two nearby starts (round-to-nearest copies at 8 and 4 bits, eval KL 0.025 and 0.22) the three losses are indistinguishable: the median paired difference of the final eval KL between the logits path and the BF16 loss is $+1.4\%$ (95\% bootstrap interval $-1.7\%$ to $+3.0\%$; 4 wins in 10) and $-0.3\%$ ($-1.4\%$ to $+1.4\%$; 6 wins in 10), against a median seed-to-seed difference of 2.8\% and 1.9\% for a fixed loss.
The loss reported by the BF16 variant is off by 0.11\% in the median, and its parameter gradient has a cosine of 0.998 with the exact one.
The 8-bit student also ends \emph{further} from the teacher than it started (0.10 against 0.025) with every loss.

\paragraph{LoRA.}
With rank-16 adapters \citep{hu2022} at learning rates $10^{-4}$ and $3\times10^{-4}$ (3 seeds) the picture is the same: from an identical start the official losses move the student to a median eval KL of 0.025--0.037 and the logits path leaves the adapter exactly at zero, while from the two quantized starts and for a task loss with a KL anchor to the base model the paths differ by less than the seeds do, without a consistent sign.
Two stronger constraints were also tested: a KL-only objective from a slightly perturbed adapter (start KL $1.5\times10^{-3}$ to $1.2\times10^{-2}$, so that a KL of zero is reachable), and a task loss with the KL weighted by 10, 100 and 1000.
In all 21 cells (7 tasks, 3 KL paths) the median final eval KL lies between 0.018 and 0.041 (single runs: 0.006 to 0.045), above the start, whatever the KL path; in particular the logits path does not return to the start either.

\paragraph{Policy optimisation.}
Libraries for PPO and TRPO compare the policy before and after an update with interface~A, in FP32.
\emph{(a)} For the rule of \texttt{rsl-rl-lib} that raises or lowers the learning rate when the KL leaves $[d/2,2d]$ with $d=0.01$, a simulation with the Beta policy of that library predicted changed decisions for large concentration parameters and a KL far below the target (up to 90 of 200 decisions).
In real training runs with the unmodified library (a reaching task with 12 actions, two configurations, 8{,}000 decisions each) no decision differs from the one taken with an FP64 KL.
\emph{(b)} For synthetic categorical policies with 6 to 50{,}000 actions and a true KL between $2\times10^{-3}$ and $5\times10^{-2}$, the official FP32 KL is within a relative $2.5\times10^{-5}$ of the truth and changes none of 960 threshold decisions.
\emph{(c)} The Fisher-vector product that TRPO obtains by differentiating the official KL twice at the old policy agrees with the exact Fisher matrix to $3.3\times10^{-7}$ (categorical) and $7.9\times10^{-8}$ (diagonal normal).
\emph{(d)} On LLM logits, the sampled estimators that RL fine-tuning frameworks log and penalise \citep{schulman2020,shao2024}, evaluated on FP32 log-probabilities as these frameworks do, have a batch mean within a relative $7.8\times10^{-6}$ of the exact value for a true KL of $10^{-3}$ to $1.4\times10^{-2}$.

\paragraph{Entropy calibration.}
Post-training quantization tools choose a clipping threshold by minimising a KL between histograms.
On four synthetic activation distributions the threshold chosen with the official FP32 KL equals the one chosen in FP64 in all 24 trials (minimum KL $10^{-4}$ to $2\times10^{-3}$).

\paragraph{Sequential circuit discovery.}
See the limits of Part~D in \cref{app:downstream}.

\paragraph{Why the intuition fails.}
Three mechanisms explain these outcomes, and they also describe where an effect should \emph{not} be expected.
\emph{First}, the spurious part of the gradient is absolute, not relative: once student and teacher differ, the true gradient is orders of magnitude larger and the two cannot be told apart in the result. The identical start is the only training setting we found in which the spurious part is all there is.
\emph{Second}, two BF16 executions that are not bitwise identical are apart by a KL of about $10^{-3}$ or more (\cref{tab:exp3c}), and trained students remain at an eval KL of 0.02--0.1 with every loss. The error region of the official KL lies below that floor.
\emph{Third}, algorithms that control a KL in a closed loop hold it near their target (about $10^{-2}$ for PPO-type rules), again above the region in which FP32 closed forms fail; the failures of the official code start where the KL is several orders of magnitude below the working point. The sequential design of \cref{sec:exp2beta} is a closed loop of the opposite kind: there the information gain falls as the data accumulate, so the loop itself moves the KL into the failing range.
The errors therefore matter where the KL \emph{value} is the product (evaluation of near-lossless changes, monitoring, rankings and thresholds on small divergences; \cref{sec:exp2beta,sec:exp3}) or where the distributions are identical or nearly so by construction (\cref{sec:exp1,sec:exp2}), and we found no case in which they change the result of ordinary training from a start that is already some distance away.
\section{Extended Related Work}
\label{app:related}

\paragraph{Softmax, cross-entropy and soft targets.}
The work closest to ours is that of \citet{barron2026}, who identify two bugs in standard \texttt{softmax}/cross-entropy code: \emph{Softmax Collapse}, first diagnosed by \citet{prieto2025} in the context of grokking, and \emph{Softmax Cancellation}, a non-zero gradient for identical student and teacher logits; they give fused implementations in JAX \citep{bradbury2018} and show that the fix stops distillation drift.
Our distillation experiment (\cref{sec:exp2}) observes the same drift independently and our logits path is closely related to their soft-target loss.
The two papers are otherwise complementary: they treat the categorical soft-target loss and its training dynamics; we measure and repair the KL interfaces themselves, across 18 families, with certified ground truth, separate the operator error of \fkl{} from the input-representation error, and study evaluation and decision making as well as training.
More broadly, a line of work traces an occasional training failure to one floating-point mechanism: absorption in the \texttt{softmax} \citep{prieto2025}, the representation limits of hyperbolic embeddings \citep{mishne2023}, and instabilities of low-precision training of large models \citep{wortsman2023}.
We follow the same approach for the KL divergence, with the difference that the failure we study is mostly silent.
\citet{blanchard2021} analyse the rounding error of the log-sum-exp and softmax \emph{function values}; our results concern second-order quantities built from two such evaluations, which those bounds do not cover.

\paragraph{Numerical analysis.}
Our tools are standard: the rounding-error model and running error bounds of \citet{higham2002}, Sterbenz's lemma \citep{sterbenz1974}, error-free transformations and double-length arithmetic \citep{knuth1997,dekker1971,shewchuk1997,ogita2005,muller2018}, compensated summation \citep{kahan1965}, and Ziv's strategy for correct rounding \citep{ziv1991}, also used by correctly rounded mathematical libraries \citep{sibidanov2022}; ground truth relies on ball arithmetic \citep{johansson2017}.
Herbie \citep{panchekha2015} searches for accurate rewrites automatically; the rewrites needed here go beyond local algebraic rewriting and must also deliver gradients.

\paragraph{Numerical defects in deep-learning software.}
Empirical studies catalogue numerical bugs in TensorFlow and PyTorch programs \citep{zhang2018,humbatova2020,wang2022} and in the libraries themselves \citep{kloberdanz2022}, and tools detect NaN/Inf-producing operations statically or by gradient-guided search \citep{zhang2020,yan2021}.
These works focus on exceptional values. Most of the errors we measure are silent: finite, plausible, and wrong.
Other probabilistic programming systems have similar KL registries \citep{dillon2017,bingham2019,carpenter2017}.

\paragraph{Low-precision training and inference.}
Mixed-precision training \citep{micikevicius2018} and BF16 \citep{kalamkar2019,zamirai2020} are now defaults for large models.
\citet{qi2025} trace the training--inference mismatch in reinforcement-learning fine-tuning of language models to BF16 rounding and recommend FP16; \citet{yuan2025} show that limited precision makes model outputs depend on batch size and hardware. Part~C of \cref{sec:exp3} measures such gaps exactly.

\paragraph{Where the KL is used.}
The KL between close distributions is the object of interest in trust-region and proximal policy optimisation \citep{schulman2015,schulman2017}, where sample-based estimators are also common \citep{schulman2020,shao2024}; in variational autoencoders and their variants \citep{kingma2014,higgins2017,loaiza2019}; in knowledge distillation \citep{hinton2015,kim2021}, including distillation of language models \citep{agarwal2024,gu2024}; in circuit discovery \citep{conmy2023}; and in model compression, both as a recovery loss for quantized models \citep{liu2024,xin2026} and as a sensitivity metric for mixed-precision assignment \citep{dong2019,cai2020,nagel2021}.

\section{Additional Details for \cref{sec:intro,sec:measure,sec:method}}
\label{app:details}

This section gives the full versions of passages that are stated briefly in \cref{sec:intro,sec:measure,sec:method}.

\paragraph{Data.} Interface~A covers 18 distribution-specific same-family KL implementations. We exclude five generic or delegated registrations: ExponentialFamily, HalfNormal, Independent, OneHotCategorical, and TransformedDistribution. The 23 configurations include both parameterisations where applicable, categorical dimensions $K\in\{2,4,16\}$, and multivariate dimensions $d\in\{2,4\}$ with ranks $r\in\{1,2\}$.
For each configuration we draw $P$ from an ordinary parameter region R0 and from stress regions (very large or small parameters, probabilities near 0 or 1), and construct $Q$ so that the \emph{true} $\KL(P\|Q)$ falls in a prescribed decade: one control bin $[10^{-1},1)$ and 14 bins down to $[10^{-15},10^{-14})$.
Two further groups contain pairs whose parameters differ by 1--8 ULP, and identical pairs $P=Q$.
Parameters are stored in the target format first, so that the truth refers to the bit patterns the library actually receives.
Cells that the floating-point grid cannot populate (mostly FP16 pairs in the smallest KL bins, where rounding $Q$ makes it identical to $P$) stay empty and are reported as such.
This gives 7,542,469 samples (3.0M FP16, 4.5M FP32).
For interface~B we generate 942,200 batches over \texttt{log\_target}$\,\in\{$False, True$\}$, FP16/FP32, $K\in\{2,4,16,256,1024\}$, batch sizes 1 and 8, and all four reductions, from logits stratified in the same way.

\paragraph{Ground truth.} For every sample, the exact KL and every gradient component are enclosed with Arb ball arithmetic \citep{johansson2017} tightly enough that the enclosure determines a unique correctly rounded value in the \emph{target} format; samples that cannot be certified are excluded and counted.
Truth and output are always compared in the same format.
For interface~B, an independent cross-check with mpmath \citep{mpmath} on 26 million values found no disagreement.

\paragraph{Methods.} \A{} is PyTorch 2.15.0 on CPU; every measurement and every repair was repeated on an NVIDIA A100 GPU with PyTorch 2.4.0 (\cref{app:gpu}).
\textsc{Official+} evaluates the unchanged official code after promoting the parameters exactly to the next format (FP32 for FP16 inputs, FP64 for FP32 inputs) and rounds the result once to the target format; it is the natural ``just use more precision'' baseline.

\paragraph{Metrics.} We report the fraction of samples with relative error $\ge1\%$ and with negative KL, the relative $L_2$ error and cosine of the gradient vector, and one summary rate: an output is a \emph{significant error} if it is off by at least 8 ULP or is qualitatively wrong (negative, NaN or infinite KL where the truth is finite and positive; spurious zero or non-zero; sign-flipped gradient component).

\paragraph{Extreme parameters.} In stress regions even FP64 does not help where the constructor clamps probabilities before the closed form is evaluated (ContinuousBernoulli) or where \texttt{log1p} of a rounded probability returns $-\infty$ (Binomial, Geometric with logits).

\paragraph{Additional interface-B results.} For probability targets (\texttt{log\_target=False}), 65--99\% of the FP32 \texttt{sum} outputs are off by at least 8 ULP and 35--70\% by at least 1\%, increasing as KL decreases; with \texttt{log\_target=True} the forward value is accurate.
Scaled reductions produce substantial target-gradient errors in FP16: with \texttt{reduction="mean"}, FP16 and $K=256$, 99.6\% of gradient rows have relative $L_2$ error $\ge1\%$ and 95.4\% contain a component of the wrong sign, because autograd evaluates the exact upstream factor $g$ as $(g\,t)/t$ and $g\,t$ is subnormal.
Overall the official implementation is correctly rounded on 52.5\% of forward and 74.9\% of backward components.
Unlike interface~A, these \emph{operator} errors essentially disappear in FP64 (NaNs at boundary inputs remain).
What does not disappear is the input-representation error: $T_{\mathrm{op}}$ differs from $T_{\mathrm{e2e}}$ by $\ge1\%$ on 40--99\% of the batches, in any precision of the operator (\cref{app:ifaceB}).

\paragraph{Additional GPU results.} On the A100 the significant-error rate of \A{} on the interface-A data is 87.7\% for FP32 and 71.5\% for FP16 inputs (CPU: 87.7\% and 70.1\%), and the per-interval error maps of the two devices agree within one percentage point in 583 of 616 cells (\cref{fig:heat32,fig:heat16}).
For \fkl{} only 50.0\% of the forward and 71.1\% of the non-boundary backward components are correctly rounded on the GPU (CPU: 52.5\% and 76.4\% of the same components); with \texttt{log\_target=True} and FP32 inputs the share of incorrectly rounded gradient components rises from 11--13\% on CPU to 32--34\%.
The two devices do not return the same bits: with the same PyTorch build, 27\% of the FP32 KL values of \A{} differ between CPU and GPU.
\textsc{Ours} is unaffected: no significant error for interface~A and the same counts of correctly rounded components for \fkl{} on both devices.
The detailed GPU results are in \cref{app:gpu}: the summary by interface and precision (\cref{tab:gpu}), the per-interval maps for both interfaces, the bit-wise comparison of the two devices, and the cases in which the GPU is the more accurate one.

\paragraph{Error-free transformations.} Where a difference of products is needed (e.g.\ $N_k=y_kS_x-x_kS_y$ for Categorical parameters (\cref{app:cat}), $\sigma_q^2-\sigma_p^2-\delta^2$ in the Normal scale gradient, $L_q^{-1}(L_p-L_q)$ for the multivariate normal), we use error-free transformations: TwoSum \citep{knuth1997}, Dekker's TwoProduct \citep{dekker1971}, expansion sums \citep{shewchuk1997} and double-length arithmetic \citep{ogita2005}.

\paragraph{Autodiff errors.} Besides inheriting the cancellation of the formula it differentiates, automatic differentiation adds failure modes of its own, e.g.\ $(g\,t)/t$ in \fkl, or $\infty-\infty$ through \texttt{torch.where}.

\paragraph{Double rounding.} FP64$\to$FP32$\to$FP16 is a double rounding; results whose leading term is itself a rounding midpoint are handled as described in \cref{app:blocks}.

\paragraph{Results of \textsc{Rewrite}.} The residue of \textsc{Rewrite} is largest for Beta and Dirichlet; its few new errors are documented in \cref{app:ifaceA}. Its cost is moderate for five of the seven timed families and higher for Gamma and Beta (\cref{app:cost}).

\paragraph{The logits path.} \Cref{eq:logits} is a row-vectorised re-implementation of the certified \texttt{Categorical(logits)} repair of \cref{tab:main}, tested against it. It is closely related to the fused soft-target cross-entropy proposed concurrently by \citet{barron2026}; we arrive at it from the Categorical closed form and also repair the \texttt{probs} parameterisation.

\end{document}